\documentclass[11pt]{article}

\usepackage[final]{acl}

\usepackage{times}
\usepackage{latexsym}
\usepackage{amsmath}
\usepackage{booktabs} 
\usepackage{subcaption}

\usepackage{array}
\usepackage{multicol}
\usepackage{ragged2e}
\usepackage{graphicx}
\usepackage{caption}
\usepackage{needspace}

\usepackage{enumitem}
\usepackage{booktabs}
\usepackage{xspace}
\usepackage{xcolor}
\usepackage{multirow}
\usepackage{amsfonts}

\usepackage[T1]{fontenc}

\usepackage[utf8]{inputenc}

\usepackage{microtype}

\usepackage{inconsolata}
\usepackage{booktabs}
\usepackage{multirow}
\usepackage[table]{xcolor}
\usepackage{pifont}   

\usepackage{graphicx}

\title{MIS-Bench: Benchmarking Multimodal LLMs for Psychotherapeutic Interpersonal Skills Assessment}

\author{
  \textbf{Yuhan Lu},
  \textbf{Yi Yao},
  \textbf{Hua Shen}\footnotemark[1],
  \textbf{Katie Aafjes-van Doorn}\footnotemark[1],
  \textbf{Zhaonan Wang}\footnotemark[1]
  \\
  NYU Shanghai, New York University
  \\
  \texttt{\{yl12006, yy5074, huashen, kav9239, zhaonan.wang\}@nyu.edu}
}

\begin{document}
\maketitle
\renewcommand{\thefootnote}{\fnsymbol{footnote}}
\footnotetext[1]{Corresponding authors.}
\renewcommand{\thefootnote}{\arabic{footnote}}
\setcounter{footnote}{0}
\begin{abstract}

Multimodal large language models (MLLMs) are increasingly used as evaluators, yet their reliability in professional assessment tasks that require expert judgment remains unclear. We investigate this challenge in the context of assessing psychotherapeutic interpersonal skills and introduce \textbf{MIS-Bench}, a Multimodal Interpersonal Skills (MIS) benchmark comprising 996 psychotherapy response videos annotated across 8 dimensions of Facilitative Interpersonal Skills. Across 9 MLLMs with multiple modality and prompting settings, we find that current models show only modest agreement with human experts, inconsistent gains from multimodal input, and limited benefits from reasoning-based prompting. To mitigate this gap, we propose \textbf{MIS-RAFT}, a regression-aware fine-tuning method inspired by RAFT and tailored to fine-grained interpersonal skill scoring at one-decimal precision. MIS-RAFT addresses the mismatch between autoregressive token prediction and scalar-valued expert assessment, significantly improving agreement with human ratings. Overall, MIS-Bench reveals a clear gap between general multimodal capability and expert-level interpersonal judgment, while MIS-RAFT offers a promising path toward more reliable model-based assessment.\footnotemark

\footnotetext{Data and code are released on GitHub: \url{https://github.com/calistalu/MIS-Bench}.}

\end{abstract}

\begin{figure*}[!t]
    \centering
    \includegraphics[width=\textwidth]{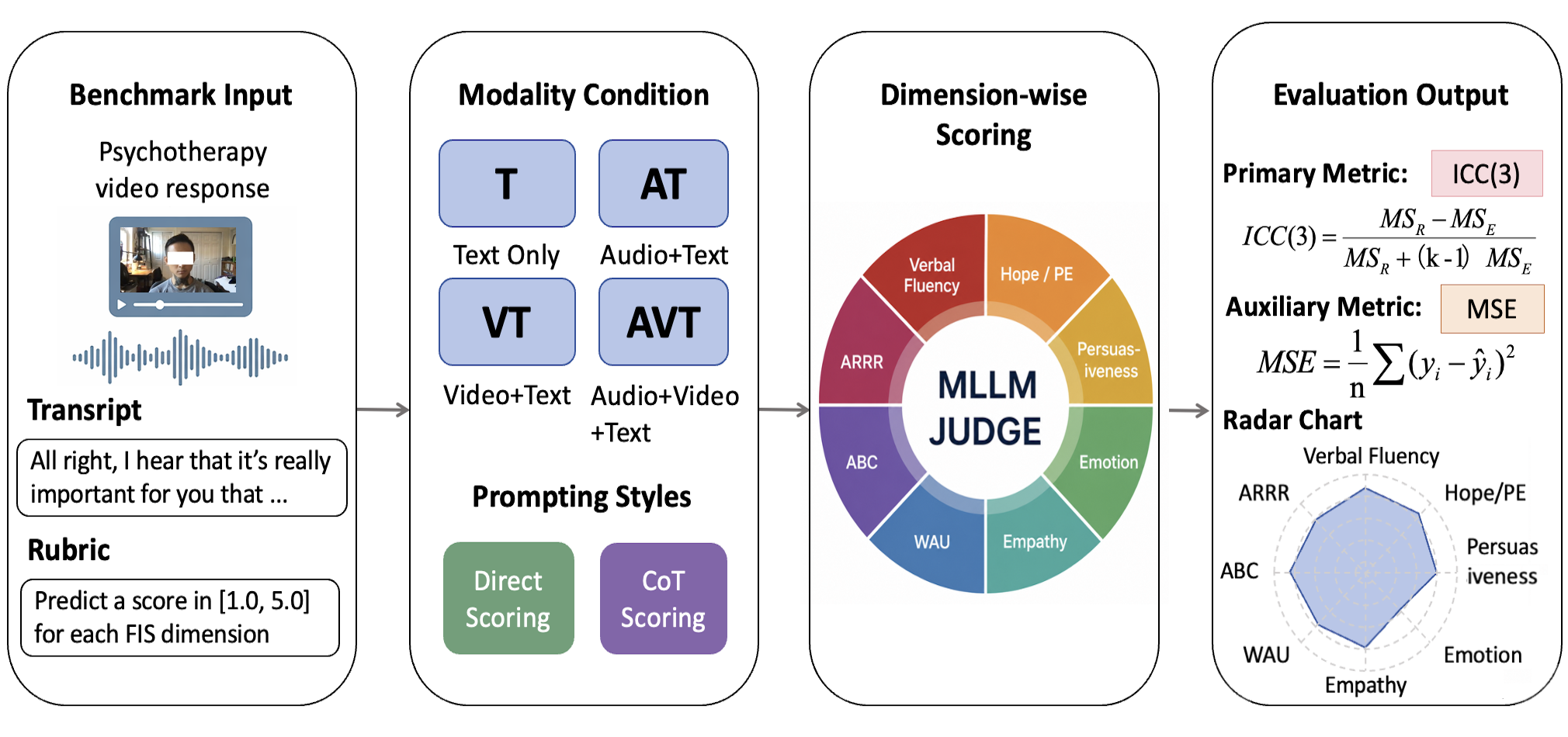}
    \caption{\textbf{Overview of MIS-Bench.} Each benchmark instance is built from a psychotherapy response video and associated transcript and rubric information. Models are evaluated with 4 modality configurations and 2 prompting styles, and predict one scalar score for each FIS dimension. Performance is measured by ICC(3) and MSE.}
    \label{fig:fisbench-overview}
\end{figure*}

\section{Introduction}
Multimodal large language models (MLLMs) are increasingly used as evaluators~\citep{zhu2023judgelm,thakur2024judging,li2024llmsasjudges}, but their reliability in \textbf{expert-based professional assessment} remains underexplored. Unlike general multimodal tasks, professional assessment often requires calibrated judgments over subtle human behaviors, domain-specific rubrics, and expert scores. Psychotherapeutic interpersonal skill assessment provides a representative and challenging setting, as accurate evaluation requires integrating what a therapist says with how the response is delivered through paralinguistic and nonverbal cues.

We study this problem through the \textbf{Facilitative Interpersonal Skills (FIS)} framework, a standardized performance-based measure of therapist interpersonal competence. In FIS, clinicians respond to challenging patient scenarios, and trained human raters score their responses across eight dimensions, including verbal fluency, emotional expression, warmth, empathy, alliance bond capacity, and alliance rupture-repair responsiveness~\citep{anderson2009therapist,anderson2016prospective}. These dimensions were carefully developed and clinically grounded, making FIS a rigorous testbed for evaluating how well MLLMs can approximate expert psychotherapist judgment.

In this work, we introduce \textbf{MIS-Bench}, a Multimodal Interpersonal Skills benchmark for evaluating MLLMs on psychotherapeutic interpersonal skill assessment, as illustrated in Figure~\ref{fig:fisbench-overview}. MIS-Bench contains 996 psychotherapy response videos annotated across the eight FIS dimensions, with transcript information, rubric descriptions, and expert ratings. Compared with existing multimodal benchmarks that emphasize general perception, question answering, or affect recognition~\citep{chen2024mllmasajudge,hu2025emobenchm,zhang2025mme}, 
%
%
MIS-Bench introduces several key advances: \ding{182} \textbf{Inherently Multimodal Evidence}, as human raters are also instructed to attend not only to verbal content but also to delivery, nonverbal cues;  \ding{183} \textbf{Fine-grained Multidimensional Scoring}, organizing evaluation around eight FIS dimensions to enable more nuanced and rigorous evaluation; \ding{184} \textbf{Expert-Based Professional Assessment}, focusing on interpersonal skills evaluation in clinical psychotherapy, MIS-Bench provides a useful probe of whether MLLMs trained primarily on large-scale web data can generalize to specialized professional evaluation.

In MIS-Bench, nine open-source and proprietary MLLMs with 4 modality configurations and 2 prompting strategies are being evaluated. Our results show that current models have only modest agreement with human experts, and performance gains vary substantially across dimensions and inconsistently across modality configurations. We also find that reasoning-based prompting offers limited benefits compared with direct scoring, suggesting that eliciting rationales does not necessarily lead to better-calibrated professional judgments.

To mitigate this gap, we further propose \textbf{MIS-RAFT}, a regression-aware fine-tuning method for multimodal interpersonal skill scoring. Inspired by Regression-Aware Fine-Tuning~\citep{lukasik2025raft}, MIS-RAFT adapts the objective to fine-grained FIS score prediction at one-decimal precision. Rather than treating scores as autoregressive tokens, MIS-RAFT directly optimizes scalar expert-score prediction, thereby bridging the mismatch between token generation and numerical professional assessment. Experiments show that MIS-RAFT substantially improves agreement with human ratings. In summary, our contributions are threefold:
\begin{itemize}[topsep=0pt, itemsep=2pt, parsep=0pt, labelwidth=*, leftmargin=1em, align=left, labelsep=5pt]
\item We introduce \textbf{MIS-Bench}, a benchmark comprising 996 psychotherapy response videos for multimodal, expert-based interpersonal skill assessment across eight FIS dimensions.

\item We systematically evaluate nine MLLMs across various modality and prompting settings, revealing modest agreement with human experts, inconsistent gains from multimodal input, and limited benefits from reasoning-based prompting.

\item We propose \textbf{MIS-RAFT}, a regression-aware fine-tuning method that mitigates the objective mismatch between autoregressive token prediction and scalar-valued expert scoring, significantly improving agreement with human ratings.
\end{itemize}

\section{Related Work}

\subsection{MLLMs as Evaluators}

Large language models have increasingly been used as automated evaluators for open-ended generation, giving rise to the LLM-as-a-Judge paradigm \citep{gu2026survey,zhu2023judgelm}. More recent work extends this idea to multimodal settings. In particular, \citet{chen2024mllmasajudge} provide one of the first systematic studies of MLLMs as judges in vision-language evaluation, while \citet{ko2025flex} show that text-only reasoning supervision can improve zero-shot multimodal evaluator performance. Most existing work, however, centers on general-purpose tasks such as pairwise comparison, ranking, and discrete scoring. MIS-Bench focuses on fine-grained assessment in a specialized professional domain, where the goal is to align with expert human ratings of psychotherapeutic interpersonal skills rather than to evaluate generic multimodal outputs.

\subsection{Existing Multimodal Benchmarks for Human-Centered Assessment}

A growing line of work studies multimodal benchmarks for human-centered capabilities such as emotion understanding and interpersonal skill assessment. Recent benchmarks including EmoBench-M \citep{hu2025emobenchm} and MME-Emotion \citep{zhang2025mme} evaluate emotional intelligence in MLLMs across diverse multimodal scenarios. Broader reviews have also examined automatic multimodal interpersonal skill assessment in social and educational settings \citep{rasipuram2020automatic,guerrerososa2025multimodal}. MIS-Bench addresses a different target: rather than general affect understanding or broad interpersonal behavior, it targets professional psychotherapy skill assessment grounded in a clinically established framework and evaluated against expert ratings.

\subsection{Computational Assessment in Psychotherapy}

Within psychotherapy research, the FIS framework has been studied as a clinically meaningful, performance-based measure of therapists' interpersonal skill, with prior reviews summarizing its theoretical basis, psychometric properties, and potential for wider use \citep{allen2024review,anderson2009therapist,anderson2016prospective,anderson2016trial,anderson2020difficult}. Attempts to automate FIS scoring have so far relied on conventional supervised models rather than MLLMs. Early efforts used conventional machine learning to predict FIS scores from linguistic or multimodal features \citep{goldberg2021computer,zech2022automatic}, and more recent work has proposed a multimodal proof of concept for AI-based FIS measurement \citep{aafjesvandoorn2025ai}, alongside discussions of its future potential for scalable assessment \citep{aafjesvandoorn2024potential}. Our work builds on this foundation, but shifts the focus from building a single automated scorer to systematically benchmarking modern open-source and closed-source MLLMs on this expert-grounded task.
\begin{table}[!t]
    \centering
    \small
    \setlength{\tabcolsep}{3pt}
    \renewcommand{\arraystretch}{1.3}
    \caption{Released feature structure of MIS-Bench. The public release contains de-identified feature representations rather than raw audiovisual recordings.}
    \label{tab:released-feature-structure}
    \begin{tabular}{@{}p{0.14\linewidth}p{0.46\linewidth}p{0.30\linewidth}@{}}
        \toprule
        Modality & Feature & Shape / Type \\
        \midrule
        Face & Action units & $T \times 8$ \\
        Face & Facial landmarks (98 points, x/y) & $T \times 196$ \\
        Face & Gaze direction & $T \times 2$ \\
        Face & Detection confidence & $T \times 2$ \\
        Audio & Wav2Vec 2.0 frame embeddings & $T \times 768$ \\
        Audio & Prosody & $T \times 15$ \\
        Audio & Log-Mel spectrogram & $T \times 80$ \\
        Audio & MFCCs & $T \times 13$ \\
        Text & Transcript words & $N_{\mathrm{word}}$ \\
        Label & FIS subscale scores & 8 scores in $[1.0, 5.0]$ \\
        \bottomrule
    \end{tabular}
\end{table}

\begin{table*}[!t]
\centering
\small
\setlength{\tabcolsep}{3.2pt}
\renewcommand{\arraystretch}{1.25}
\caption{Human single-rater agreement ceiling on the 625 fully rated videos. Values are single-measure ICC(3,1) scores reported in percentage points; brackets denote 95\% confidence intervals. Higher is better.}
\label{tab:human-icc}
\begin{tabular*}{\textwidth}{@{\extracolsep{\fill}}lcccccccc@{}}
\toprule
 & VF & H/PE & Per. & Emo. & WAU & Emp. & ABC & ARRR \\
\midrule
Human ICC(3,1) & 64.0 & 53.0 & 54.0 & 51.0 & 50.0 & 43.0 & 51.0 & 53.0 \\
95\% CI & [61.0, 68.0] & [49.0, 58.0] & [49.0, 58.0] & [46.0, 55.0] & [45.0, 54.0] & [38.0, 48.0] & [46.0, 55.0] & [48.0, 57.0] \\
\bottomrule
\end{tabular*}
\end{table*}

\section{MIS-Bench}
\subsection{Task Definition}
Grounded in the theory of \textit{Facilitative Interpersonal Skills} (FIS), MIS-Bench evaluates whether an MLLM can approximate the domain-expert assessment of psychotherapeutic interpersonal skills on a calibrated scale. Let $\mathcal{D}=\{d_1,\ldots,d_8\}$ denote the eight FIS dimensions: \textit{Verbal Fluency (VF)}, \textit{Hope and Positive Expectations (H/PE)}, \textit{Persuasiveness (Per.)}, \textit{Emotional Expression (Emo.)}, \textit{Warmth, Acceptance, and Understanding (WAU)}, \textit{Empathy (Emp.)}, \textit{Alliance Bond Capacity (ABC)}, and \textit{Alliance Rupture--Repair Responsiveness (ARRR)} \citep{anderson2009therapist,allen2024review}. For each clinician response instance $i$, we denote its video, audio, and transcript as $V_i$, $A_i$, and $T_i$, respectively. Each dimension $d \in \mathcal{D}$ is then associated with a rubric description $R_d$ and an expert-aggregated human score $y_{i,d}\in[1.0,5.0]$ obtained from trained human observer ratings.

We formulate MIS-Bench as a dimension-wise scoring task. Specifically, each clinician-response and dimension pair $(i,d)$ is treated as an independent evaluation instance. Given a modality configuration $m \in \{\text{text-only}, \text{audio+text}, \text{video+text}, \text{audio+video+text}\}$, we define the corresponding model input as $X_i^m$, which contains the available modalities, transcript text, and rubric information under that setting. Under prompting strategy $s \in \{\text{direct}, \text{CoT}\}$, an MLLM $f_\theta$ produces a response
\begin{equation}
A_{i,d}^{m,s} = f_\theta(P_s, X_i^m, R_d),
\end{equation}
where $P_s$ denotes the prompt template, and $A_{i,d}^{m,s}$ is either a direct numeric answer or a reasoning trace followed by a final numeric score. We then extract the predicted scalar score using a deterministic parser $g(\cdot)$:
\begin{equation}
\hat{y}_{i,d}^{m,s} = g(A_{i,d}^{m,s}), \quad \hat{y}_{i,d}^{m,s} \in [1.0,5.0].
\end{equation}
The benchmark evaluates how closely $\hat{y}_{i,d}^{m,s}$ approximates the expert-aggregated human score $y_{i,d}$ across dimensions, models, modalities, and prompting strategies.

\subsection{Benchmark Construction}
The current release of MIS-Bench contains 996 clinician response videos annotated under the FIS framework. Each benchmark instance is derived from a recorded clinician response video and includes associated metadata, response text fields, score annotations, and rubric descriptions, and human ratings for the eight FIS dimensions. To protect participant privacy, the public release does not include raw audiovisual recordings. Instead, it provides de-identified multimodal representations, transcripts, and FIS scores. Table~\ref{tab:released-feature-structure} summarizes the released feature structure.

MIS-Bench is constructed by repurposing clinician responses collected under the original FIS performance task into a benchmark for multimodal model evaluation. In the source protocol, clinicians first view standardized patient stimulus clips portraying a challenging therapeutic
encounter (e.g., the patient wants to stop therapy), and then provide an immediate spoken response as if they were in session with the patient \citep{anderson2009therapist,anderson2020difficult}. The ground truth labels in MIS-Bench are derived from trained human observer-raters following the original FIS rating manual. Raters are instructed to consider both the \emph{content} of the response (verbal information) and its \emph{delivery} (nonverbal and paraverbal information) when assigning scores. Each response is evaluated on the eight FIS subdimensions using a 5-point rating scale from 1 (\emph{skill deficit}) to 5 (\emph{optimal demonstration of skill}), where 3 corresponds to the ordinary level of FIS expected of a trained clinician \citep{anderson2009therapist}. 

Each video was independently rated by trained expert raters following the FIS manual. For benchmark use, we aggregate available ratings for each video--dimension pair by taking their arithmetic mean and retaining one decimal place, which serves as the ground-truth target in all experiments. Of the 996 videos, 625 have complete ratings from all three assigned raters across all eight dimensions. Table~\ref{tab:human-icc} reports single-measure ICC(3,1) of human agreement ceiling and 95\% confidence intervals among the human raters.

\subsection{Evaluation Metrics}
We use \textbf{ICC(3)} as the primary evaluation metric. ICC(3) directly measures agreement between the model and human raters while treating the model as a fixed additional rater. Higher ICC(3) values indicate stronger agreement with human expert judgment.

For each FIS dimension, let $y_{ij}$ denote the score assigned to response instance $i$ by rater $j$, where the raters correspond to the expert-aggregated human score and the model prediction. Let $n$ be the number of response instances and $k$ be the number of raters. We compute ICC(3,1) using the two-way mixed-effects formulation:
\begin{equation}
\mathrm{ICC}(3,1)=
\frac{MS_R-MS_{\mathrm{err}}}
{MS_R+(k-1)MS_{\mathrm{err}}},
\end{equation}
where $MS_R$ is the mean square for response instances and $MS_{\mathrm{err}}$ is the residual mean square:
\begin{equation}
MS_R = \frac{k}{n-1}\sum_{i=1}^{n}(\bar{y}_{i\cdot}-\bar{y}_{\cdot\cdot})^2,
\end{equation}
\begin{equation}
MS_{\mathrm{err}} =
\frac{1}{(n-1)(k-1)}
\sum_{i=1}^{n}\sum_{j=1}^{k}
(y_{ij}-\bar{y}_{i\cdot}-\bar{y}_{\cdot j}+\bar{y}_{\cdot\cdot})^2.
\end{equation}
Here, $\bar{y}_{i\cdot}$ is the mean score for response instance $i$ across raters, $\bar{y}_{\cdot j}$ is the mean score assigned by rater $j$ across response instances, and $\bar{y}_{\cdot\cdot}$ is the grand mean.

As an auxiliary error metric, we report MSE in Appendix~\ref{sec:appendix-additional-heatmaps} to measure the absolute numerical deviation between the model-predicted score and the expert-aggregated human score. Lower MSE indicates better calibration to the human expert rating.

\definecolor{bestred}{RGB}{252,220,220}
\definecolor{secondblue}{RGB}{220,235,252}
\definecolor{thirdyellow}{RGB}{252,248,196}

\newcommand{\best}[1]{\cellcolor{bestred}{#1}}
\newcommand{\second}[1]{\cellcolor{secondblue}{#1}}
\newcommand{\third}[1]{\cellcolor{thirdyellow}{#1}}
\newcommand{\cmark}{\ding{51}}

\begin{table*}[!t]
\centering
\small
\setlength{\tabcolsep}{4.5pt}
\renewcommand{\arraystretch}{1.4}
\caption{
\textbf{Overall performance comparison on MIS-Bench using dimension-level ICC (\%).}
For each model, we report the dimension-level ICC values from the configuration that achieves the highest overall score. This overall score is computed by averaging the eight predicted dimension scores for each video and comparing the resulting average against the corresponding average of the human ratings. If a model output is unparseable for any of the eight dimensions, that video is excluded when determining the overall score, with a mean exclusion rate of 1.5\%.
The top three performing results in each dimension are highlighted in \textcolor{black}{\colorbox{bestred}{red}} (1st), \textcolor{black}{\colorbox{secondblue}{blue}} (2nd), and \textcolor{black}{\colorbox{thirdyellow}{yellow}} (3rd) backgrounds, respectively.
}
\label{tab:overall-performance}
\resizebox{\textwidth}{!}{
\begin{tabular}{l c c c c c c c c c c c c}
\toprule
Model & LLM Size & A & V & T & VF & H/PE & Per. & Emo. & WAU & Emp. & ABC & ARRR \\
\midrule
\multicolumn{13}{c}{\textit{Open-source MLLMs}} \\
\midrule
Qwen2-Audio-7B-Instruct~\citep{chu2024qwen2audio} & 7B & \cmark &  & \cmark & -6.0 & 17.1 & 8.4 & 0.9 & 8.4 & 2.3 & 9.1 & 0.1 \\
Qwen2.5-VL-7B-Instruct~\citep{bai2025qwen25vl} & 7B &  & \cmark & \cmark & 2.8 & \second{39.9} & \third{29.9} & 16.5 & \second{36.4} & \third{28.8} & \third{33.0} & \third{27.4} \\
Qwen2.5-Omni-7B~\citep{xu2025qwen25omni} & 7B & \cmark & \cmark & \cmark & -3.1 & 19.4 & 15.3 & 11.0 & 18.6 & 23.0 & 26.1 & 20.1 \\
Qwen3-Omni-30B-A3B-Instruct~\citep{xu2025qwen3omni} & 30B & \cmark & \cmark & \cmark & \third{21.2} & 24.6 & 25.1 & 14.8 & 22.8 & 24.6 & 18.6 & 22.7 \\
InternVL3.5-8B-HF~\citep{wang2025internvl35} & 8B &  & \cmark & \cmark & 5.9 & 32.4 & 16.6 & 14.7 & 28.9 & 22.9 & 26.8 & 17.6 \\
MiniCPM-V-4.5~\citep{yu2025minicpmv45} & 9B &  & \cmark & \cmark & 8.8 & 26.9 & 17.2 & 14.3 & 25.1 & 22.6 & 32.8 & 18.5 \\
MiniCPM-o-4.5~\citep{openbmb2026minicpmo45} & 9B & \cmark & \cmark & \cmark & \second{30.5} & 32.6 & 26.7 & \second{21.9} & 32.4 & \second{29.2} & 31.0 & 22.1 \\
\midrule
\multicolumn{13}{c}{\textit{Closed-source MLLMs}} \\
\midrule
GPT-5.4~\citep{openai2026gpt54} & --- &  & \cmark & \cmark & \best{40.1} & \best{46.4} & \second{31.2} & \third{21.8} & \best{46.7} & \best{42.4} & \second{36.5} & \second{43.3} \\
Gemini-3.1-Pro-Preview~\citep{google2026gemini31pro} & --- & \cmark & \cmark & \cmark & 17.5 & \third{37.8} & \best{35.9} & \best{32.0} & \third{33.9} & 25.3 & \best{38.7} & \best{46.6} \\
\bottomrule
\end{tabular}
}
\end{table*}

\section{Experiments}
\label{sec:benchmarking-mllms}

\subsection{Experimental Setup}

\paragraph{Model Selection.}
We evaluate the performance on a total of nine cutting-edge MLLMs, including Qwen2-Audio-7B-Instruct~\citep{chu2024qwen2audio}, Qwen2.5-VL-7B-Instruct~\citep{bai2025qwen25vl}, Qwen2.5-Omni-7B~\citep{xu2025qwen25omni}, Qwen3-Omni-30B-A3B-Instruct~\citep{xu2025qwen3omni}, InternVL3.5-8B-HF~\citep{wang2025internvl35}, MiniCPM-V-4.5~\citep{yu2025minicpmv45}, MiniCPM-o-4.5~\citep{openbmb2026minicpmo45}, GPT-5.4~\citep{openai2026gpt54}, and Gemini-3.1-Pro-Preview~\citep{google2026gemini31pro}. For most open-sourced MLLMs, we use the vLLM framework~\citep{kwon2023vllm} to perform inference on our benchmark. For models that are not supported by vLLM or require model-specific loading and preprocessing, we instead use a unified evaluation pipeline built on the Hugging Face Transformers library~\citep{wolf2020transformers}, following the official open-source code and model cards provided by the authors. For closed-sourced MLLMs, inference is conducted directly via their official APIs. The main benchmark is conducted under zero-shot settings. To ensure a fair comparison across models, we maintain consistent prompting wherever possible, with only minor modifications when necessary. For example, for audio-language models, the word ``video'' in prompts is replaced with ``audio'' when the input modality is audio only. Beyond the main zero-shot evaluation, we further conduct a targeted few-shot CoT analysis on MiniCPM-o-4.5.

\paragraph{Evaluation Strategy.}
We evaluate each model under four modality configurations: text-only, audio+text, video+text, and audio+video+text. The text-only setting provides only the transcript; audio+text provides the audio signal together with the transcript; video+text provides the video signal together with the transcript; and audio+video+text provides visual, auditory, and textual inputs. Since not all models support all input modalities, each model is evaluated only under the modalities supported by its interface. For each modality configuration, we compare two prompting strategies: direct scoring and chain-of-thought (CoT) prompting. In the direct setting, the model is instructed to output only a numeric FIS score. In the CoT setting, the model is asked to first provide an explicit analysis of the observed therapist behaviors and then produce a final numeric score. The full prompt templates for all eight benchmark settings are provided in Appendix~\ref{sec:appendix-prompts}.

\begin{figure*}[!t]
    \centering
    \includegraphics[width=\linewidth]{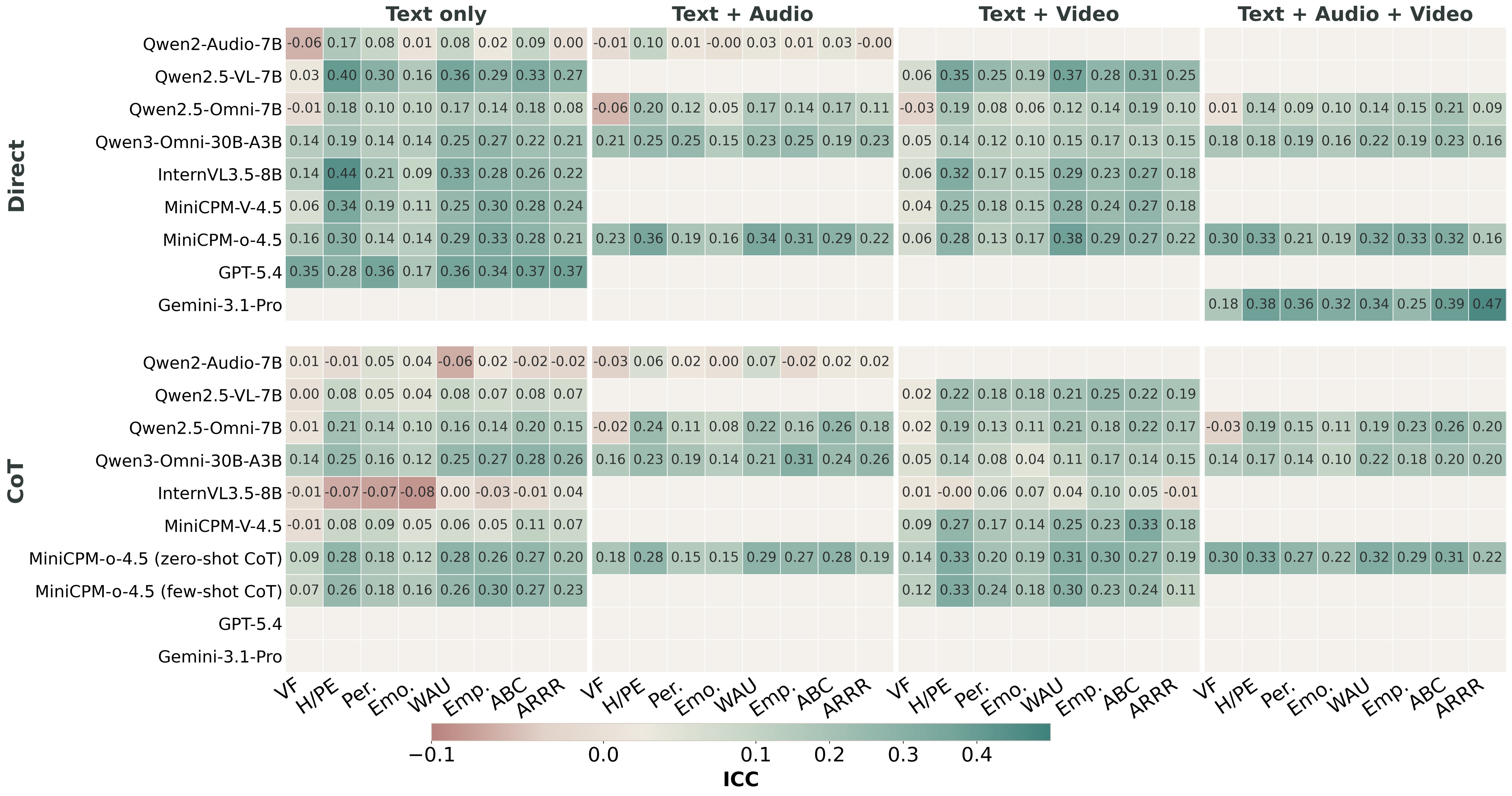}
    \caption{\textbf{Dimension-level ICC heatmaps on MIS-Bench.}
We report every supported model $\times$ modality $\times$ prompting combination across four modality configurations and two prompting strategies. In each panel, rows correspond to models and columns correspond to the eight FIS dimensions. Higher ICC indicates better agreement with human ratings, and blank cells indicate configurations unsupported by a given model. The complete grid enables controlled comparisons among models within a common configuration.
}
    \label{fig:fisbench-dimension-icc}
\end{figure*}

\subsection{Main Results}
We present the overall benchmark comparison in Table~\ref{tab:overall-performance}, and the fine-grained dimension-level results in Figure~\ref{fig:fisbench-dimension-icc}. Averaging the eight dimension-level ICC values, the top-performing models are GPT-5.4~\citep{openai2026gpt54}, Gemini-3.1-Pro-Preview~\citep{google2026gemini31pro}, and MiniCPM-o-4.5~\citep{openbmb2026minicpmo45}, with average ICC scores of 38.6\%, 33.5\%, and 28.3\%, respectively. Human single-rater agreement provides an empirical context for interpreting these model results: human ICC ranges from 43.0\% to 64.0\% across 8 dimensions (Table~\ref{tab:human-icc}). In comparison, all evaluated MLLMs report average dimension-level ICC scores below 40\%, and all open-source MLLMs remain below 30\%, well below the human agreement ceiling. This performance landscape highlights the challenging nature of the MIS-Bench benchmark. It also suggests that even state-of-the-art MLLMs are still in the early stages of developing reliable expert-level judgment for psychotherapy skill assessment.

Among open-source models, no single model family dominates across all settings. Qwen2.5-VL-7B-Instruct performs strongly under text-only direct prompting, Qwen3-Omni-30B-A3B-Instruct reaches its best result under audio+text direct prompting, and MiniCPM-o-4.5 performs best under video+text direct prompting. This pattern suggests that strong performance on MIS-Bench can arise from different model design choices, but it also indicates that current systems are highly sensitive to input configuration and prompting strategy.

Figure~\ref{fig:fisbench-dimension-icc} further reveals substantial variation across FIS dimensions. Some dimensions are consistently easier than others: averaged across evaluated configurations, \textit{Hope and Positive Expectations} achieves the highest mean ICC (0.214), followed by \textit{Alliance Bond Capacity} (0.206) and \textit{Warmth, Acceptance, and Understanding} (0.206). In contrast, \textit{Emotional Expression} (0.108) and especially \textit{Verbal Fluency} (0.071) are much more difficult. This gap suggests that current MLLMs are relatively better at capturing broad relational and supportive qualities than fine-grained delivery-related skills that depend on subtle emotional or acoustic cues.

\subsection{Observations and Insights}

Delving into the experimental results, we derive the following key observations and insights.

\paragraph{Obs. 1: Current MLLMs remain limited on expert-grounded psychotherapy skill assessment.}
Although several models achieve non-trivial alignment with human raters on this expert-grounded task, even the strongest results remain a meaningful gap. And performance varies substantially across models and settings. This result confirms that MIS-Bench is a difficult test of domain-specific judgment requiring nuanced interpretation of therapist behavior. More broadly, it suggests that strong general-purpose multimodal capability does not automatically translate into reliable evaluation of interpersonal skills in psychotherapy settings.

\paragraph{Obs. 2: Direct score prediction is generally more reliable than CoT prompting.}
Across matched model--modality comparisons, direct prompting achieves an average overall ICC improvement of 0.047 over zero-shot CoT prompting across 20 matched pairs. At the dimension level, this corresponds to 160 matched comparisons across eight FIS dimensions, where direct prompting achieves higher ICC in 96/160 comparisons and lower MSE in 159/160 comparisons. We also evaluate few-shot CoT on MiniCPM-o-4.5 in two representative settings: text-only and video+text. Although few-shot CoT improves over zero-shot CoT, it still achieves lower ICC compared with direct prompting in 14/16 corresponding dimension-level comparisons. This pattern suggests that encouraging models to verbalize intermediate reasoning does not necessarily improve judgment quality. A plausible explanation is that, for subjective scoring tasks, explicit reasoning may introduce additional unsupported inferences and weaken calibration rather than strengthen it.

\paragraph{Obs. 3:  Audio provides uniquely useful information for \textit{Verbal Fluency}.}
Among the eight FIS dimensions, \textit{Verbal Fluency} shows the clearest benefit from audio-aware evaluation. This is most evident when comparing audio+video+text against video+text settings under direct prompting: for all three models that support both configurations, adding audio improves ICC on \textit{Verbal Fluency}, with a mean gain of 0.136, and also reduces MSE by an average of 0.178. This finding is consistent with the construct itself, since verbal fluency depends not only on lexical content but also on pace, hesitation, smoothness, and other prosodic cues that cannot be fully recovered from transcript or visual information alone.

\paragraph{Obs. 4: Current MLLMs have yet to fully and effectively leverage multimodal information.}
Richer multimodal input does not lead to consistent gains across models, and the best-performing configuration varies substantially from one model to another. In our experiments, the strongest setting for different models ranges from text-only direct and video+text direct to audio+text direct and audio+video+text CoT. The main bottleneck is therefore not merely access to more modalities, but the ability to integrate multimodal evidence and convert them into stable and dimension-specific judgments. 

\begin{figure}[!t]
    \centering
    \includegraphics[width=\linewidth]{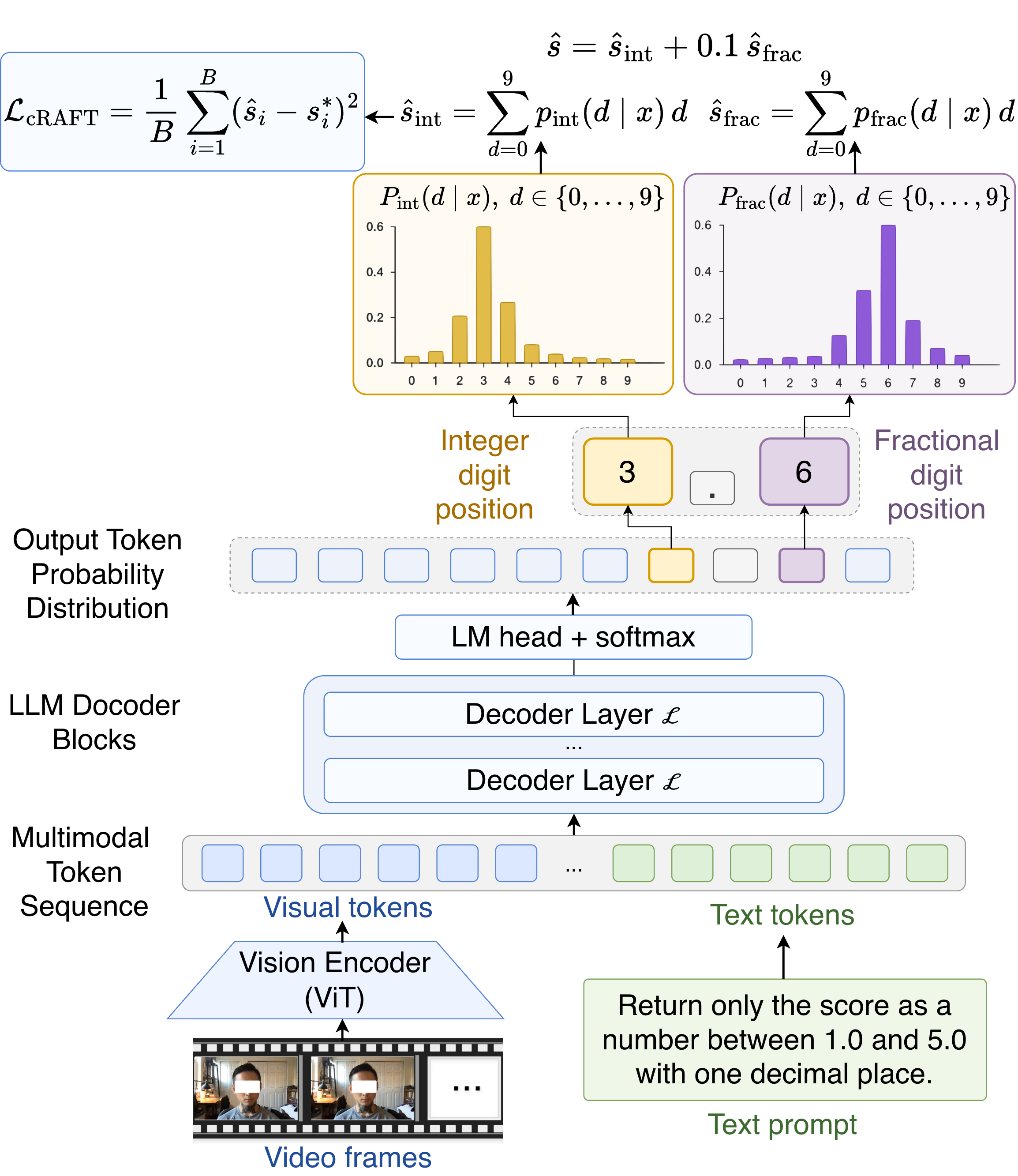}
    \caption{MIS-RAFT scoring pipeline for one-decimal score prediction.}
    \label{fig:craft}
\end{figure}

\begin{table*}[!t]
\centering
\small
\setlength{\tabcolsep}{4pt}
\renewcommand{\arraystretch}{1.4}
\begin{tabular*}{\textwidth}{@{\extracolsep{\fill}}lccc@{}}
\toprule
\textbf{FIS Dimension} & \textbf{Raw} & \textbf{Fine-tuning} & \textbf{MIS-RAFT} \\
\midrule
VF &
$-3.8$ [$-12.5$, $2.0$] &
$\mathbf{1.2}$ [$-16.3$, $18.9$] &
$-0.7$ [$-16.0$, $12.6$] \\

H/PE &
$37.6$ [$24.0$, $49.5$] &
$21.7$ [$12.3$, $31.2$] &
$\mathbf{50.7}$ [$38.0$, $62.3$] \\

Per. &
$-2.0$ [$-6.6$, $0.0$] &
$1.8$ [$-15.7$, $16.6$] &
$\mathbf{33.6}$ [$20.2$, $45.9$] \\

Emo. &
$-0.0$ [$-0.0$, $0.0$] &
$-13.5$ [$-29.4$, $4.1$] &
$\mathbf{17.0}$ [$0.6$, $31.1$] \\

WAU &
$12.9$ [$-1.6$, $27.3$] &
$-12.2$ [$-22.9$, $0.0$] &
$\mathbf{31.4}$ [$14.9$, $46.6$] \\

Emp. &
$\mathbf{43.6}$ [$28.2$, $55.9$] &
$3.7$ [$-7.6$, $15.3$] &
$35.5$ [$19.1$, $50.7$] \\

ABC &
$15.2$ [$-5.3$, $34.9$] &
$7.6$ [$-16.6$, $28.7$] &
$\mathbf{41.8}$ [$26.8$, $55.2$] \\

ARRR &
$6.7$ [$-12.8$, $25.6$] &
$11.1$ [$-1.9$, $24.2$] &
$\mathbf{39.2}$ [$24.4$, $52.4$] \\

\midrule
Mean Average &
$13.8$ &
$2.7$ &
$\mathbf{31.1}$ \\
\bottomrule
\end{tabular*}
\caption{ICC(3,1) by FIS dimension under raw inference, standard fine-tuning, and MIS-RAFT fine-tuning for the Qwen2.5-VL backbone. Values are reported in percentage points; brackets report 95\% percentile bootstrap confidence intervals computed over videos using 10,000 bootstrap resamples. Higher is better, and the best ICC in each row is shown in bold.}
\label{tab:raft-icc}
\end{table*}

\section{MIS-RAFT}

\subsection{Motivation}
In addition to the zero-shot and few-shot evaluations, we study whether task-specific adaptation can mitigate the misalignment between the autoregressive MLLM training objectives and the regression task here. In MIS-Bench, the target for each dimension is an averaged human score retained to one decimal place, which makes the problem fundamentally different from standard language generation. Under the common approach of token-level fine-tuning with cross-entropy loss, the decoder is rewarded for exact token matching and therefore treats score prediction as a task of string generation rather than numeric prediction. As a result, predictions such as ``3.1'' and ``5.0'' are both treated as equally incorrect relative to ``3.0'', even though they differ substantially in both clinical meaning and evaluation error. These considerations motivate a mitigation method that can directly optimize scalar score prediction instead of exact token matching. MIS-RAFT retains the standard autoregressive architecture and modifies only its training objective; it does not introduce an additional classification or regression head.

\subsection{Method}
To address the objective mismatch above, we propose \textbf{MIS-RAFT}, a regression-aware fine-tuning method for multimodal interpersonal skill scoring. MIS-RAFT is inspired by Regression-Aware Fine-Tuning (RAFT) proposed by \citet{lukasik2025raft}, which reformulates autoregressive numeric prediction as expected-value regression over the model's predictive distribution. Let $x$ denote the input and $s^\ast$ the gold score. Under standard token-level fine-tuning, the objective can be written as
\begin{equation}
\mathcal{L}_{\mathrm{CE}} = -\log p(\mathrm{str}(s^\ast)\mid x),
\end{equation}
where $\mathrm{str}(s^\ast)$ is the string form of the target score. This objective encourages exact string matching, but does not encode the numerical distance between nearby predictions.

In the original RAFT formulation, a numeric prediction is constructed from the model distribution over a predefined candidate set $\mathcal{S}_{\mathrm{grid}}$:

\begin{equation}
\hat{s}_{\mathrm{RAFT}}(x)
=
\sum_{s \in \mathcal{S}_{\mathrm{grid}}}
p(\mathrm{str}(s)\mid x)\, s,
\end{equation}
and then minimizing the squared loss
\begin{equation}
\mathcal{L}_{\mathrm{RAFT}}
=
\bigl(s^\ast - \hat{s}_{\mathrm{RAFT}}(x)\bigr)^2.
\end{equation}
Compared to cross-entropy, this objective is better aligned with scalar score prediction because it directly rewards numerically closer outputs.

However, this grid-based formulation is most natural when predictions are restricted to a finite, predefined set of target values. In MIS-Bench, scores are instead predicted at one-decimal precision, calling for a formulation that can directly accommodate more fine-grained values. We therefore extend RAFT from probability-weighted prediction over a single-token value grid to a compositional decimal representation, where integer and fractional digits jointly define the predicted score.

Specifically, as shown in Figure~\ref{fig:craft},  MIS-RAFT decomposes each target score into an integer digit and a fractional digit. At inference, it computes probability-weighted expectations over the digit vocabulary $\{0,1,\dots,9\}$:
\begin{align}
\hat{s}_{\text{int}} &= \sum_{d=0}^{9} p_{\text{int}}(d \mid x)\, d, \\
\hat{s}_{\text{frac}} &= \sum_{d=0}^{9} p_{\text{frac}}(d \mid x)\, d.
\end{align}
Then combined into a continuous expected score prediction
\begin{equation}
\hat{s} = \hat{s}_{\text{int}} + 0.1\,\hat{s}_{\text{frac}}.
\end{equation}
The final training objective minimizes the mean squared error between this expected score and the gold target:
\begin{equation}
\mathcal{L}_{\mathrm{cRAFT}} = \frac{1}{B}\sum_{i=1}^{B}(\hat{s}_i - s_i^\ast)^2.
\end{equation}

\subsection{Evaluation}
\paragraph{Experimental Setup.}
We evaluate MIS-RAFT as a focused ablation study on top of MIS-Bench. Following the benchmark results in the previous section, we use Qwen2.5-VL as the backbone model and apply LoRA-based parameter-efficient fine-tuning. Holding the backbone and PEFT framework fixed, we compare \textbf{raw inference} without task-specific tuning, \textbf{standard fine-tuning} with the conventional token-level cross-entropy objective, and \textbf{MIS-RAFT fine-tuning} with the regression-aware objective described above.

\begin{figure}[!t]
    \centering
    \includegraphics[width=\linewidth]{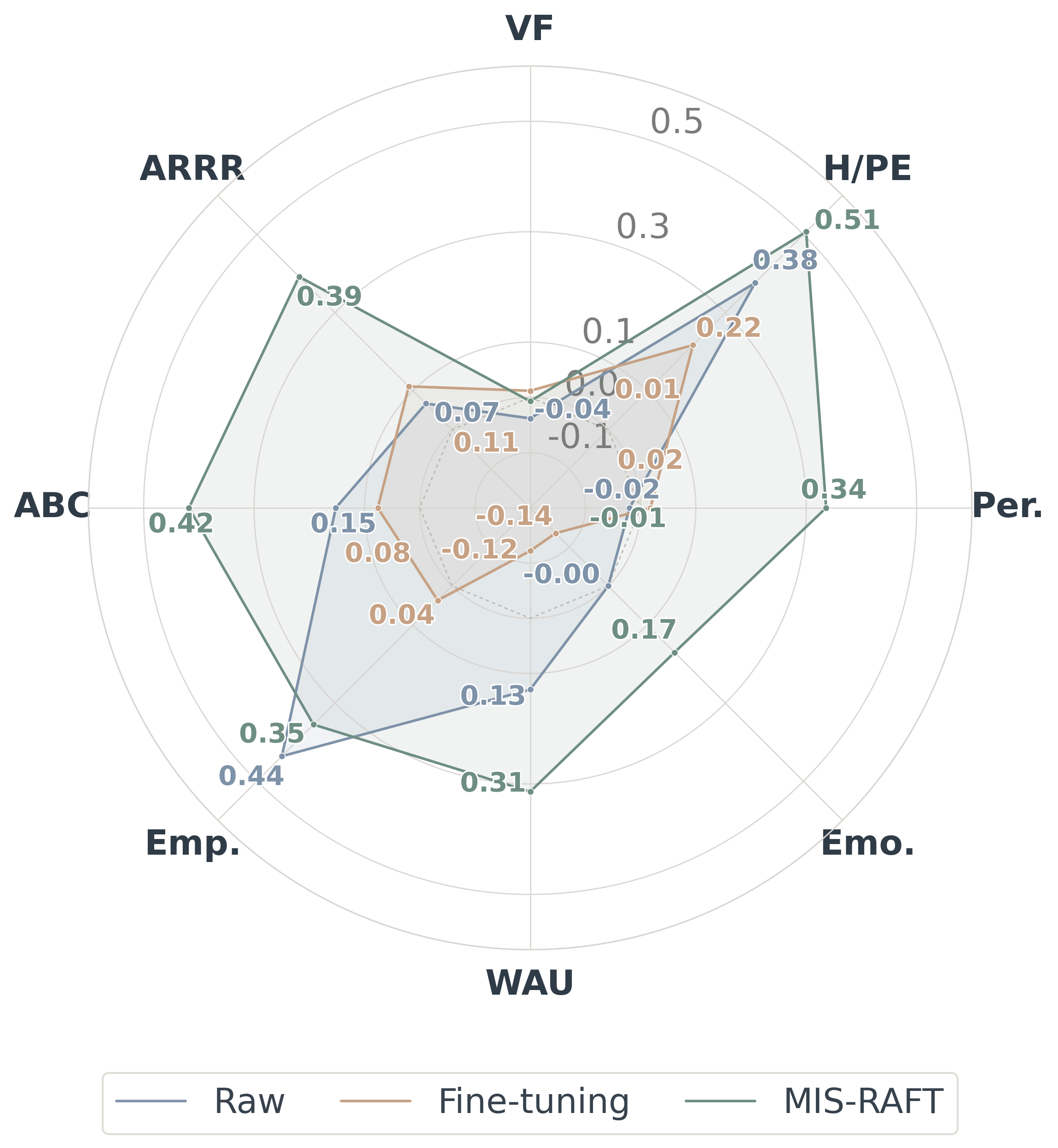}
    \caption{Radar chart of ICC scores across eight FIS dimensions.}
    \label{fig:icc-radar}
\end{figure}

\paragraph{Result Analysis.}
Table~\ref{tab:raft-icc} and Figure~\ref{fig:icc-radar} show that MIS-RAFT achieves the strongest alignment with human ratings. Averaged across the eight dimensions, raw inference attains a mean ICC of 0.138, standard fine-tuning drops to 0.027, and MIS-RAFT improves to 0.311. MIS-RAFT outperforms substantially on most dimensions, especially in \textit{Persuasiveness}, \textit{Emotional Expression}, and \textit{Alliance Bond Capacity}, indicating that the regression-aware objective is much better aligned with the nature of the task than standard token-level fine-tuning. Relative to the human single-rater ceiling in Table~\ref{tab:human-icc}, MIS-RAFT approaches human agreement on selected dimensions, reaching 0.507 on \textit{Hope and Positive Expectations} and 0.418 on \textit{Alliance Bond Capacity}.

As shown in Figure~\ref{fig:icc-radar}, raw inference is uneven, with reasonable performance on a few dimensions but weak results on  \textit{Persuasiveness, Emotional Expression, and ARRR}. Standard fine-tuning introduces instability, including negative ICCs on some dimensions and a collapse on \textit{Empathy}. In contrast, MIS-RAFT expands the profile more uniformly, which indicates broader and more balanced improvements.

\section{Conclusion}

This paper presents \textbf{MIS-Bench}, a benchmark for evaluating Multimodal Large Language Models on psychotherapy interpersonal skill assessment. Our experiments show that even the latest MLLMs, including both proprietary and open-source models, remain limited in their ability to align with expert human ratings on this expert-grounded task, with inconsistent gains from multimodal inputs, and even performance degradation under CoT prompting. We therefore introduce MIS-RAFT, a regression-aware mitigation for score prediction, and show that it improves alignment with expert ratings over standard fine-tuning. Future work could investigate a broader range of post-training strategies for this setting, including instruction tuning, preference-based alignment, and reinforcement learning across open-source MLLMs.

\section*{Limitations}
 
Our study has several limitations. First, the sample size of the current benchmark is still modest relative to the diversity and complexity of real-world psychotherapy interactions. Second, models receive only the therapist response in the selected modality configuration, whereas human raters also view the standardized patient stimulus clip, although the rubric descriptions provide partial scenario context to compensate this model-rater information asymmetry. Third, while we study a focused regression-aware adaptation strategy through RAFT, we do not systematically investigate a broader range of post-training methods or alternative scoring formulations across open-source models, which would be a valuable direction for future work.

\section*{Ethical Considerations}
This study evaluates MLLMs on psychotherapy interpersonal skill assessment in order to better understand their capabilities, limitations, and risks. All experiments are conducted in a benchmark setting using publicly available models or APIs, with no deployment in real clinical practice. Current models may still fail to align with expert judgment in systematic ways, and therefore not supposed to be used for therapist supervision, training, hiring, or patient-facing evaluation, given their potential impact on the individuals being assessed and on downstream professional outcomes. MIS-Bench is intended to expose these limitations and support research on more responsible model-based evaluation.

Before data collection, all participants were informed of the study procedures, including the recording process and the intended research use of the collected data. Written informed consent was obtained from all participants for conversation recording and for
the use of the resulting data in scientific publications, including
conference and journal papers. The data were
collected under a human subjects research protocol approved by
the Institutional Review Board (IRB) of Yeshiva University, under
Human Subject Research Protocol No. 1293523.

\section*{Acknowledgments}
This work was supported by the Dean's Undergraduate Research Fund (DURF) and Center for Data Science at NYU Shanghai. Computational resources were provided by NYU Shanghai High Performance Computing and NYU Torch.

We are grateful to Jordan Bate of Adelphi University and Katie Aafjes-van Doorn for their contributions to the collection of FIS dataset. We also thank Bonian Jia and Kaifang Mao for helpful discussions and feedback. 

\bibliography{custom}
\clearpage
\appendix

\section{Implementation Details}
\label{sec:appendix-implementation-details}

\subsection{Prompt Templates}
\label{sec:appendix-prompts}

\raggedbottom
The eight prompt templates used in the main benchmark are provided below, covering the four modality configurations and two prompting styles.

\newcommand{\promptfig}[1]{%
\noindent\includegraphics[width=\linewidth]{#1}\par\vspace{0.6em}
}

\promptfig{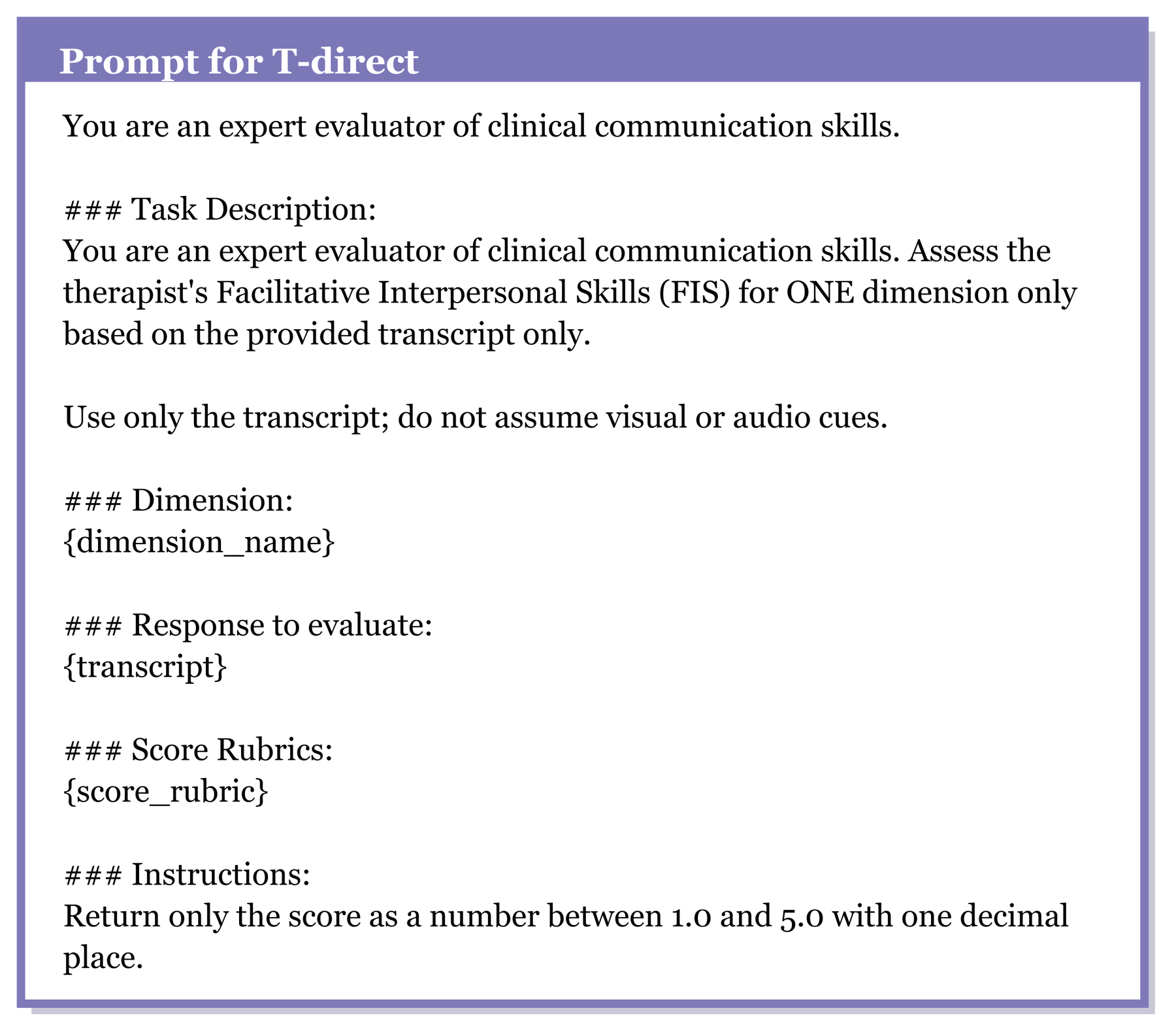}
\promptfig{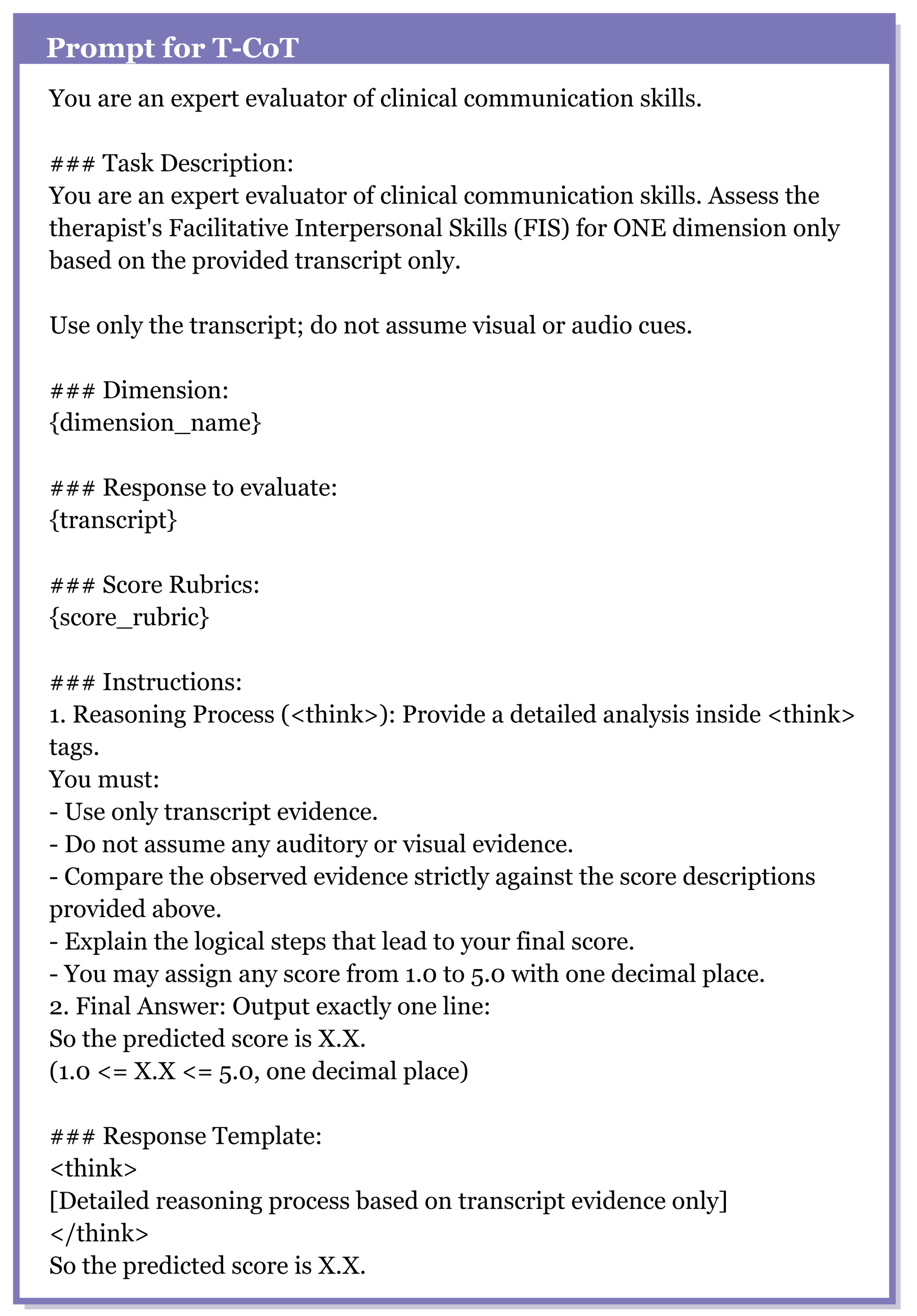}
\promptfig{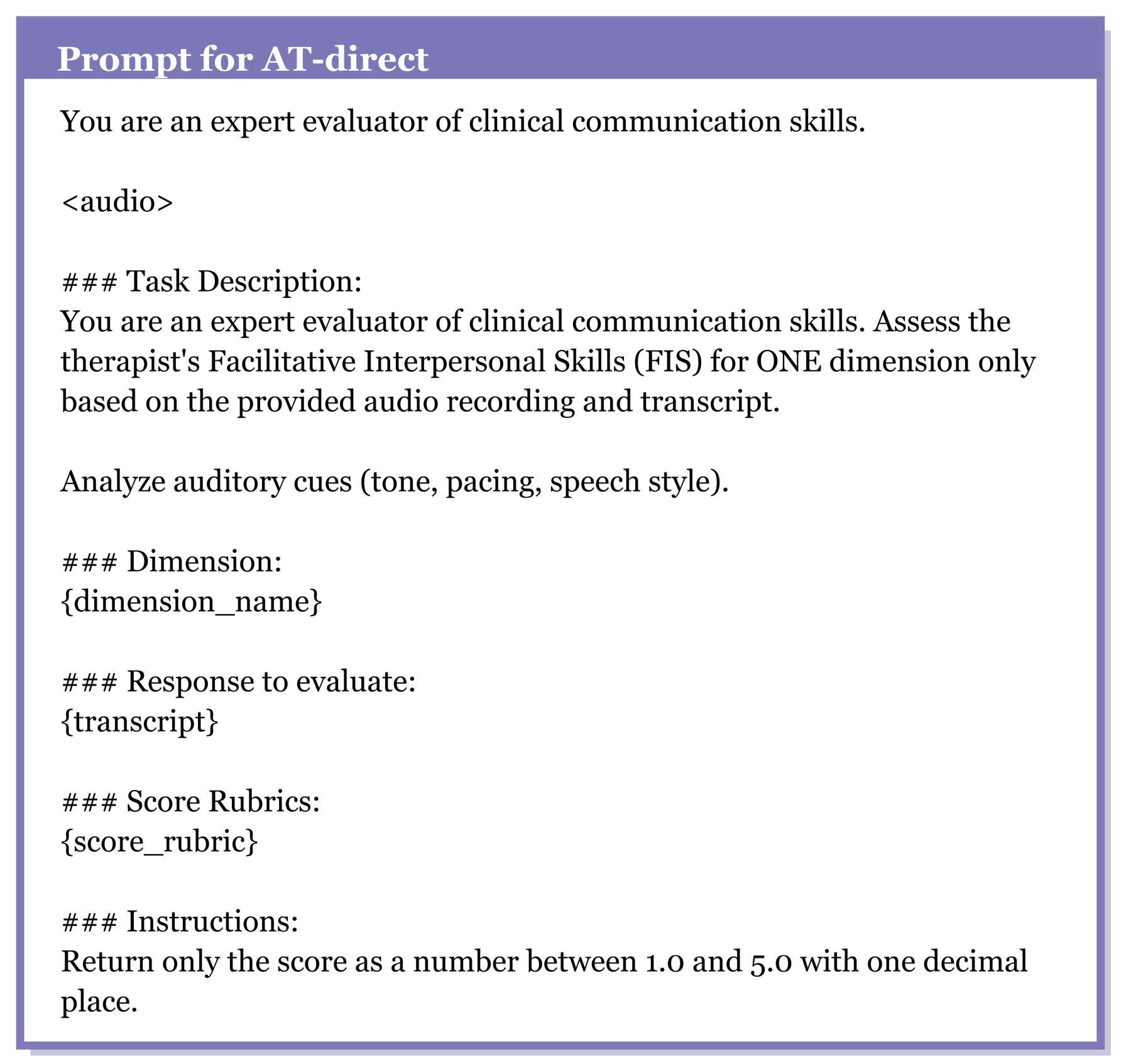}
\promptfig{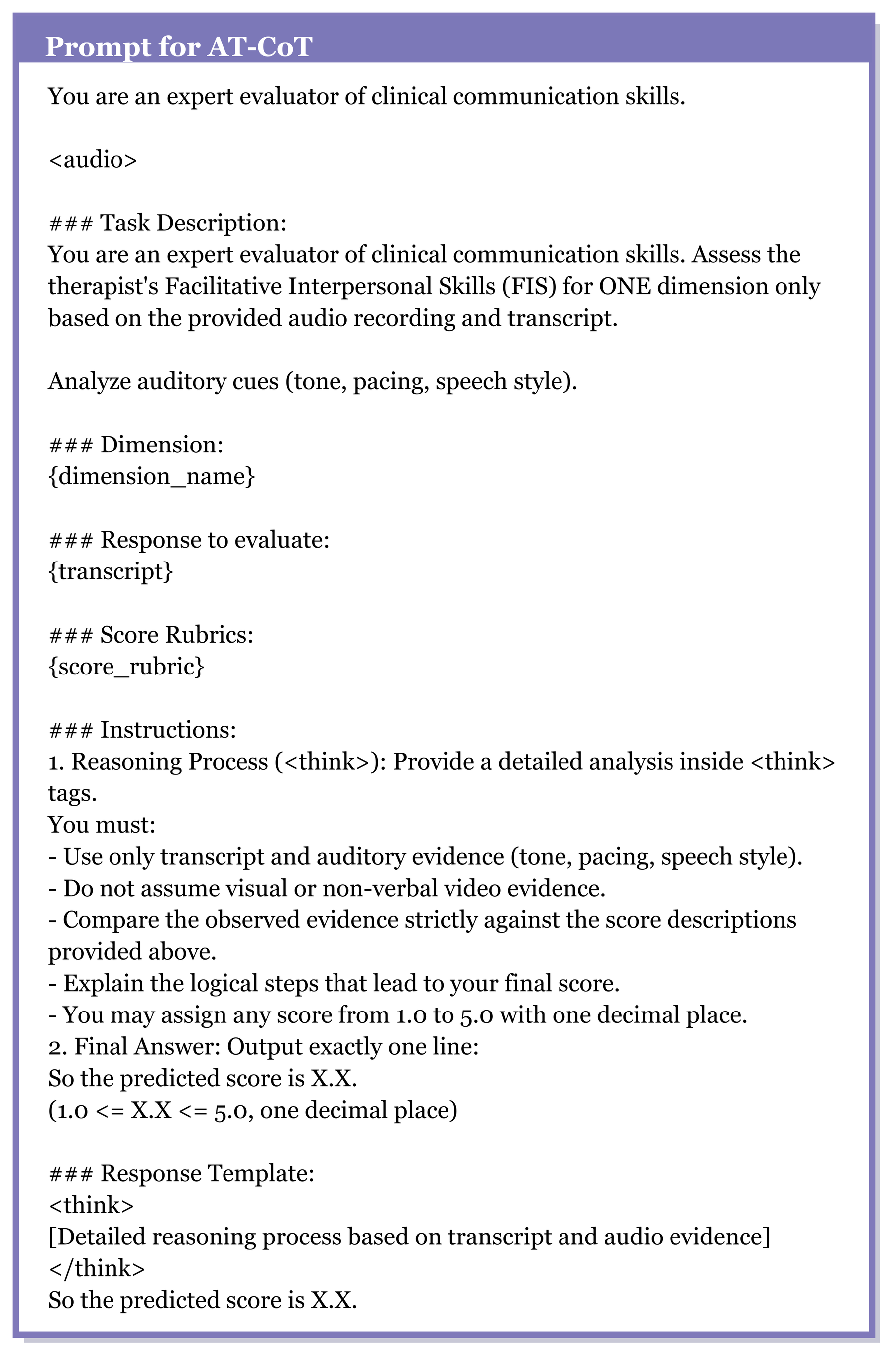}
\promptfig{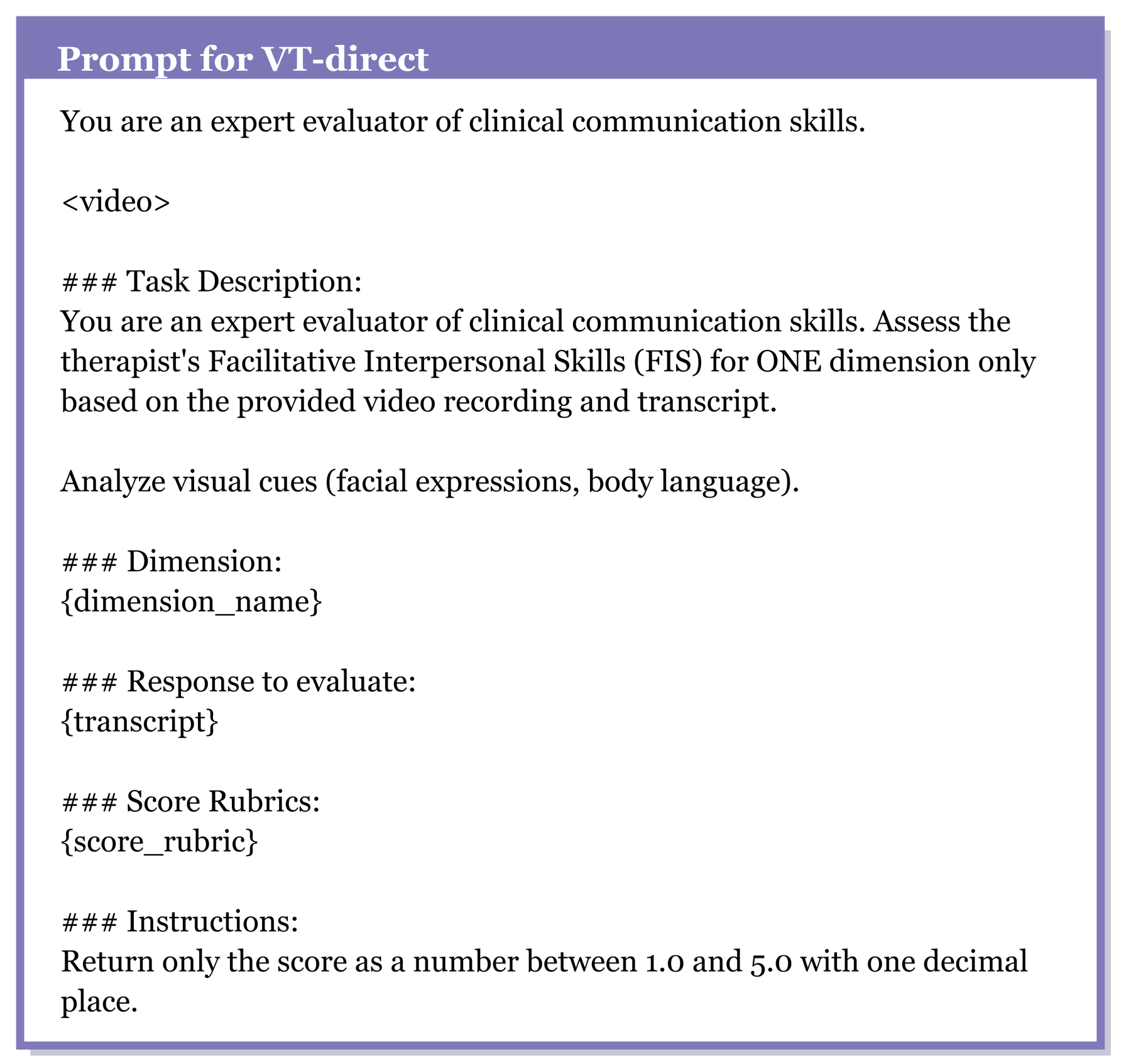}
\promptfig{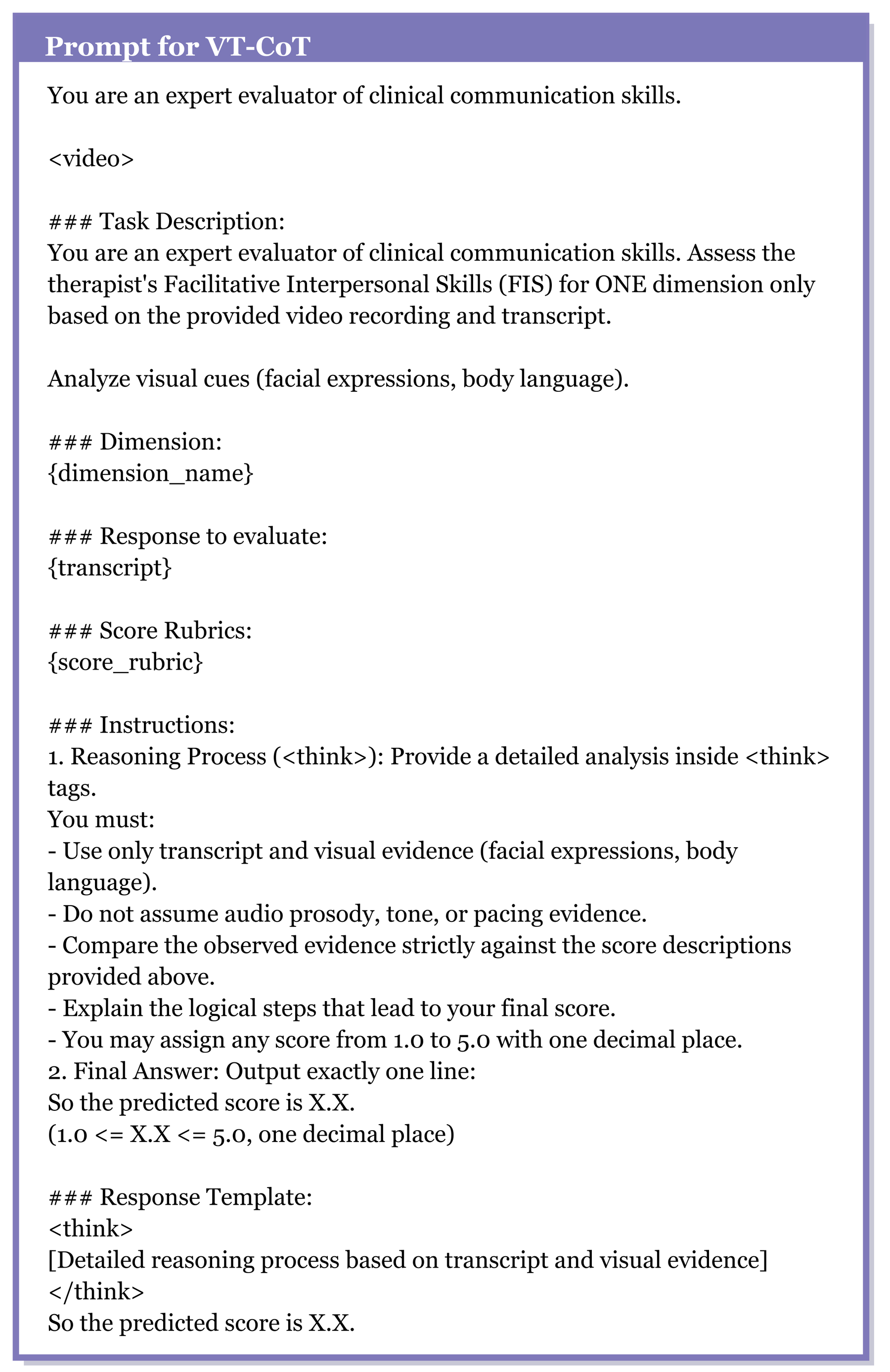}
\promptfig{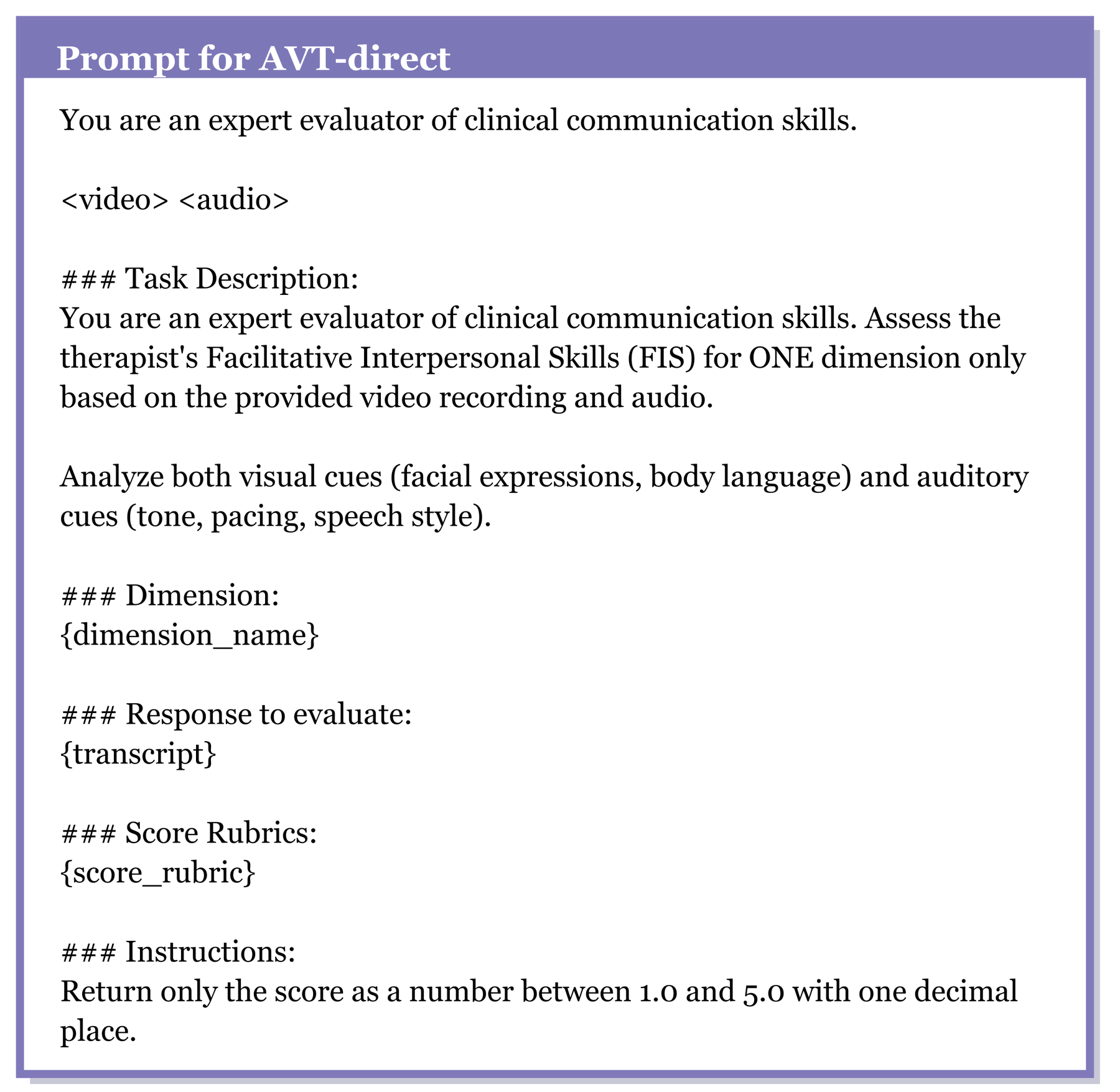}
\promptfig{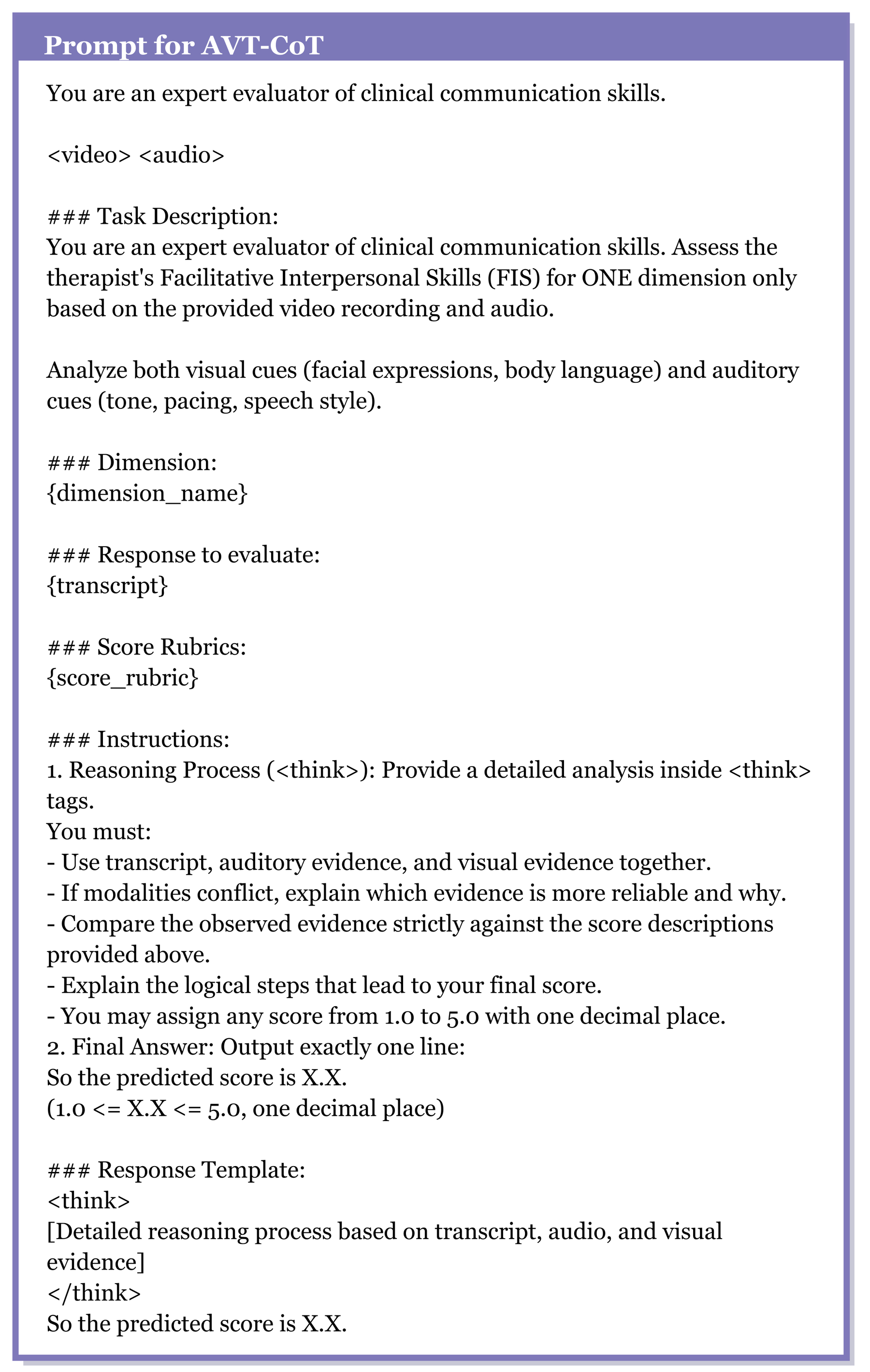}

\clearpage
\subsection{Score Rubrics} 
\label{sec:appendix-score-rubrics}
{\scriptsize
\captionof{table}{Full scoring rubrics for the eight FIS dimensions.}
\label{tab:appendix-score-rubrics}
\medskip
\textbf{Verbal Fluency} \newline  \textbf{Criterion.} Verbal Fluency. This item is a rating of the extent to which the participant is verbally capable and at-ease in communicating. The response is delivered in a relaxed manner and without significant signs of anxiety (e.g., broken speech, extended and awkward pauses, and clarity in communication). However, the content of what is said is not rated, but rather how it is spoken. \newline \textbf{Score 1.} The participant has great difficulty verbalizing a response (e.g., obviously anxious, sounds shaky or timid), reflecting a clear avoidance or anxiety. The participant may lack confidence and is clearly uncertain or even difficult to follow. \newline \textbf{Score 2.} Fluency is disrupted by the participant’s anxiety and avoidance of verbal expression. The respondent may be obviously anxious and struggling to formulate a response. At times, the communication may be choppy, even halting. [Note: In some rare instances a response could represent an avoidance of the interpersonal situation through anxious rambling. It would need to be clear that the participant’s ramblings are the result of anxiety over communicating with another]. \newline \textbf{Score 3.} A moderate level of verbal fluency indicates that the participant's response is conversational and mostly easy to follow. \newline \textbf{Score 4.} The response is fluent, and there is little that is difficult to follow. \newline \textbf{Score 5.} The participant is at great ease and communicates ideas with no anxiety, reflecting a desire to "approach" the other. The verbal quality of the response may have a "melodic," rhythmical quality and is easy to follow; the response is fluent. \\
\par\vspace{0.45em}
\textbf{Hope \& Positive Expectations} \newline  \textbf{Criterion.} Hope \& Positive Expectations. This item rates expressions of hope, optimism, and positive expectations for change. Staats (1989, 2001) defines hope as the interaction between wishes and expectations. The interpersonal skills needed for hope involve facilitating a) personal agency and b) building the pathways needed for attaining desired goals and expectations (Steed, 2002). Hope is related to persuasiveness and collaboration in the sense that hope and positive expectations are often built through offering a rationale, friendliness, and enthusiasm. As defined here, hope focuses more on building client agency for actions that will facilitate meeting the client’s goals whereas persuasion is based more on a plausible explanation (which may or may not include hope). \newline \textbf{Score 1.} The participant's response is hopeless or is even pessimistic. For example, the participant may address only issues or concerns beyond the control of the other or subtly suggests that the other cannot change or improve his/her problems. \newline \textbf{Score 2.} The participant responds with some hopelessness, including subtle expressions of feeling unable to help the client. \newline \textbf{Score 3.} The response is ordinary OR the optimism of the response is not discernible. There may be some hopefulness expressed, but with little confidence or reason for being hopeful. \newline \textbf{Score 4.} A general sense of optimism about the client’s situation is detected. Specifically, the participant's response is directed toward building the client’s agency OR facilitating the building of pathways to meet the client’s goals. \newline \textbf{Score 5.} The participant's response expresses clear hope about the client’s future and/or positive expectations about therapeutic work. In addition, for a response to be coded as a “5” there needs to be an allusion to building the client’s agency as well as how the client might participate or do something that will help move toward his/her desired goals (i.e., pathways). \\
\par\vspace{0.45em}
\textbf{Persuasiveness} \newline  \textbf{Criterion.} Persuasiveness. Persuasiveness is the capacity to induce the other to accept a view that may be different from his or her own view. It involves that ability to convey a clear, organized understanding about the meaning of the other’s source of distress. Persuasiveness implies an ability to communicate what Jerome Frank called a “believable myth.” This capacity implies that the persuasive therapist must be convincing in communicating this belief-system. Rating Notes: For rating purposes, the response does not necessarily need to convey an entire world view, but a point of view that is implied to be at least slightly different from the client in the video clip. High ratings require that the participant provide a clear belief in a point of view or rationale. It is necessary that the rationale be relevant to the other’s problems and at least somewhat novel to the other’s experience. For this item, the rater should disregard personal beliefs about the validity of the participant's rationale, but instead rate the extent to which the participant might persuade another (i.e., ability to "sell" their rationale). \newline \textbf{Score 1.} The participant’s response is unorganized, incoherent, and difficult to follow. The participant may also not know what to say. \newline \textbf{Score 2.} The participant is unpersuasive. Unpersuasive responses may be characterized by either \newline \textbf{Score 3.} The participant’s response conveys little sense of persuasiveness. \newline \textbf{Score 4.} The participant speaks persuasively. The rationale may be more implicit and it is even possible that the rationale, though present, may be unclear, superficial, or marginally relevant to the other’s problems. \newline \textbf{Score 5.} The participant is highly persuasive. Persuasive persons may speak with great confidence, certainty, and authority. Advice may or may not be given, but the participant must offer some rationale or re-framing of the other's experience. \\
\par\vspace{0.45em}
\textbf{Emotional Expression} \newline  \textbf{Criterion.} Emotional Expression. This item rates the energy and emotion in the participant's response. This item rates the extent to which the participant’s response is delivered with effective expressions of emotion. \newline \textbf{Score 1.} The participant speaks with little or no affect and may be dull or boring (e.g., speaking in a near monotone voice and without intensity). \newline \textbf{Score 2.} The participant may display some sense of interest or curiosity, but the response is not emotionally engaging. Prosody is somewhat less than typical to casual conversation. \newline \textbf{Score 3.} The participant has prosody, but it is the amount of emotion that one might find in ordinary conversation. \newline \textbf{Score 4.} The participant is emotionally expressive at a moderate level. There is more emotion than found in ordinary speech, but it is not as focused in its delivery as the maximum rating of 5. \newline \textbf{Score 5.} There is affect and prosody in the participant's voice. The response is delivered in a highly emotional and engaging manner. The primary criterion is that the vocal expression conveys emotion. There may be a more focused delivery of emotional intonations to emphasize meanings that influence other processes (e.g. persuasion). The participant may even be somewhat provocative or challenging in delivering an emotion-based response toward the client in the video clip. However, a "5" should not be rated if the affect is primarily demeaning or hostile toward the other (in which case a "3" would be the maximum rating possible). \\
\par\vspace{0.45em}
\textbf{Warmth, Acceptance, \& Understanding} \newline  \textbf{Criterion.} Warmth, Acceptance, \& Understanding. This item is a rating of the ability of the participant to care for and accept the other. Therapist behaviors/attitudes that might indicate an absence of acceptance and understanding include: a judgmental attitude, condescension, rudeness, disapproval, guilt-induction, exasperation, or annoyance. Often it will be necessary to avoid rating what the participant is doing (e.g., giving advice), but rate how it is being done. Note that accepting does not necessarily mean approval, but rather a caring attitude and determination to help the other. \newline \textbf{Score 1.} The participant has an obvious lack of respect, acceptance, or warmth for the other (e.g., clearly pejorative comments, judgmental attitude, condescension, disapproval, guilt induction, blaming the other). \newline \textbf{Score 2.} The participant conveys a subtle lack of respect, acceptance, or concern of the other (e.g., sarcasm, exasperation, annoyance). \newline \textbf{Score 3.} There is an "ordinary" level of courtesy and warmth in the response OR the participant's opinion of the other may not be clearly discernable from the response. \newline \textbf{Score 4.} The participant’s response is genuinely nonjudgmental and gently explores the other's thoughts, feelings, alternatives for dealing with future situations, etc. The participant appears concerned for and respectful of the client. \newline \textbf{Score 5.} The participant expresses clear and obvious warmth, concern and acceptance. The participant may, for example, make a compassionate attempt to relate to the other’s experience. \\
\par\vspace{0.45em}
\textbf{Empathy} \newline  \textbf{Criterion.} Empathy. The capacity to respond with an expressed understanding of the subjective experience of the client. The response must also convey an accurate understanding of the thoughts and emotions expressed in the video clip. Therefore, it is especially important that the rater have an accurate understanding of the client’s experiences in the video clips. \newline \textbf{Score 1.} Participant clearly distorts the other's experience. That is, the participant misidentifies a significant component of the other's complaints, beliefs, emotions, etc. Give a rating of 1 if the response indicates a clear disregard of the other’s experience. \newline \textbf{Score 2.} Participant does not communicate an awareness or understanding of the other’s experience, and/or there are minor distortions of the other’s experience. Some aspects of the participant's response may be irrelevant to the other's concerns (when clearly not an attempt to change the other's focus). \newline \textbf{Score 3.} Participant is generally accurate about the other's experience but only perceives the more obvious aspects of the other's experience or concerns. \newline \textbf{Score 4.} Participant comments accurately on the other's experience but not to the extent required to receive a “5" rating. The distinction between the 4 and 5 ratings are matters of intensity. \newline \textbf{Score 5.} Participant alludes to the client's experience so that it is clear that he/she has not only listened, but obtained an exceptional comprehension of what the other is experiencing. In order to receive a "5" the participant must infer something about the other's experience that is not explicitly stated by the other but is clearly identifiable in the client’s nonverbal expression. \\
\par\vspace{0.45em}
\textbf{Alliance Bond Capacity} \newline  \textbf{Criterion.} Alliance Bond Capacity. This item rates the participant's capacity to provide a collaborative environment, one in which there is recognition of the need to work with the client jointly on problems. \newline \textbf{Score 1.} The participant actively undermines a mutual collaboration. The participant may respond in a way that is over-involved or reactive (e.g., moralistic lecturing, "preaching" to the other, assuming all responsibility). The rupture may also involve withdrawal or under-involvement in the participant’s response (e.g., putting all the responsibility for change on the other). \newline \textbf{Score 2.} The participant may slightly undermine the building of a collaborative atmosphere, although it may be unintentional or superficial. \newline \textbf{Score 3.} The participant neither undermines nor attempts to enhance a collaborative effort. \newline \textbf{Score 4.} Some effort to collaborate is made but not as strong as a “5” (e.g., subtle invitations to engage in working with the client). \newline \textbf{Score 5.} Specific actions on the part of the participant help create a collaborative atmosphere. There should be a sense that the participant is attempting to work with the other to create a "we-ness" that is implied in the participant's behavior (e.g., participant checks with the other by asking questions about the "fit" of interpretations, conclusions, goals, etc.). \\
\par\vspace{0.45em}
\textbf{Alliance Rupture-Repair Responsiveness} \newline  \textbf{Criterion.} Alliance Rupture-Repair Responsiveness. Background: Each client in video clips is expressing an interpersonal issue that involves the patient-therapist relationship. Each video clip places the participant in the middle of alliance rupture episodes. Further, these rupture episodes take place at different locations within the interpersonal circle, which requires interpersonal flexibility for the therapist. The interpersonal problem with the client-therapist relationship threatens the development of the alliance. This item rates the extent to which the therapist appears responsive to the interpersonal issue. In some cases, the problem is clearly stated as when Suzie angrily berates the therapist for being ineffective. In other cases, the problem is more implicit such as when Lauren idealizes the therapist to the extent of leaving herself overly vulnerable to disappointment. \newline \textbf{Score 1.} Participant reacts to the interpersonal tension in a way that is nonproductive or in a way that likely exacerbates the rupture (e.g., responding negatively to a hostile client; responding to a controlling client with counter-control). Low scores also may be given when the participant fails to respond to the interpersonal issue involved in a way that indicates that the participant is avoiding the interpersonal issue or the relationship altogether. \newline \textbf{Score 2.} Participant addresses an issue to discuss that it is tangentially related to the interpersonal issue presented, but directs the discussion away from the present relationship situation. \newline \textbf{Score 3.} There may be more casual recognition of the interpersonal situation, but the response does not draw for further discussion of the issue or the relationship. \newline \textbf{Score 4.} Participant recognizes the other’s interpersonal issue, and may discuss this further in more general terms (or discuss some secondary element of the other's issue or the relationship). \newline \textbf{Score 5.} Participant makes attempts to repair the interpersonal issue by engaging the client in a direct discussion of the immediate moment-to-moment interaction. This may include how specific relational messages are being expressed by the client in the video clip. Optimal responses will include attempts to not only allude to the interpersonal tension, but make some attempt to repair that interpersonal issue. \\
\par\vspace{0.45em}
}

\section{Annotation Protocol and Human-Rater Reliability}
\label{sec:appendix-annotation-reliability}

Each video was independently rated by trained expert raters following the original FIS rating manual. Raters were trained by licensed clinicians who had themselves been trained by Tim Anderson, the developer of the FIS manual. Training continued until each rater achieved ICC greater than 0.8 with expert ratings on six consecutive clips. Raters also attended weekly calibration meetings to discuss ratings and prevent rater drift.

\section{Dataset Statistics}
\label{sec:appendix_dataset_statistics}

We report descriptive statistics of the FIS benchmark labels (gold scores), including inter-dimension correlation, per-dimension summary statistics, and gold-score distributions. The dataset contains 996 unique videos and 7,968 gold-score annotations (8 dimensions per video). The average video duration over all 996 videos is 62.42 seconds.

\begin{figure*}[!t]
\centering

\begin{minipage}[t]{0.47\textwidth}
\centering
\vspace{3.5em}
\small
\setlength{\tabcolsep}{4pt}
\renewcommand{\arraystretch}{1.25}
\captionof{table}{Per-dimension summary statistics of FIS gold scores.}
\label{tab:fis_dim_stats}
\begin{tabular}{lcccc}
\toprule
Dimension & Mean & Std & Min & Max \\
\midrule
VF    & 3.658 & 0.493 & 1.0 & 5.0 \\
H/PE  & 3.420 & 0.394 & 2.0 & 4.3 \\
Per.  & 3.613 & 0.454 & 1.3 & 5.0 \\
Emo.  & 3.627 & 0.426 & 2.0 & 5.0 \\
WAU   & 3.607 & 0.443 & 1.3 & 5.0 \\
Emp.  & 3.534 & 0.414 & 2.2 & 4.7 \\
ABC   & 3.564 & 0.478 & 1.3 & 5.0 \\
ARRR  & 3.433 & 0.585 & 1.0 & 5.0 \\
\bottomrule
\end{tabular}
\end{minipage}
\hfill
\begin{minipage}[t]{0.47\textwidth}
\centering
\vspace{0.8em}
\includegraphics[width=0.92\linewidth]{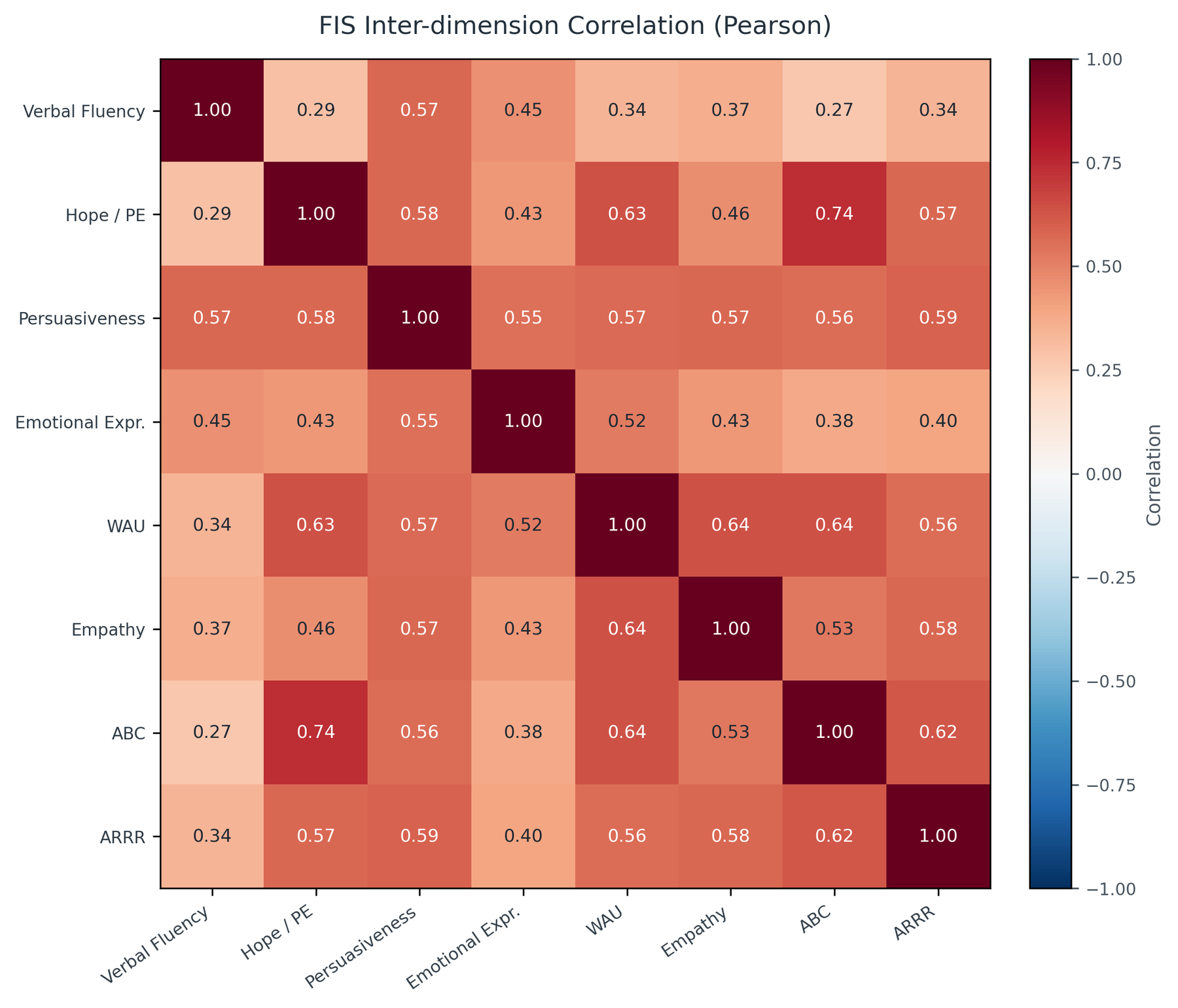}
\captionof{figure}{Inter-dimension Pearson correlation of FIS gold scores.}
\label{fig:fis_inter_dim_corr}
\end{minipage}

\vspace{3.5em}

\includegraphics[width=0.96\textwidth]{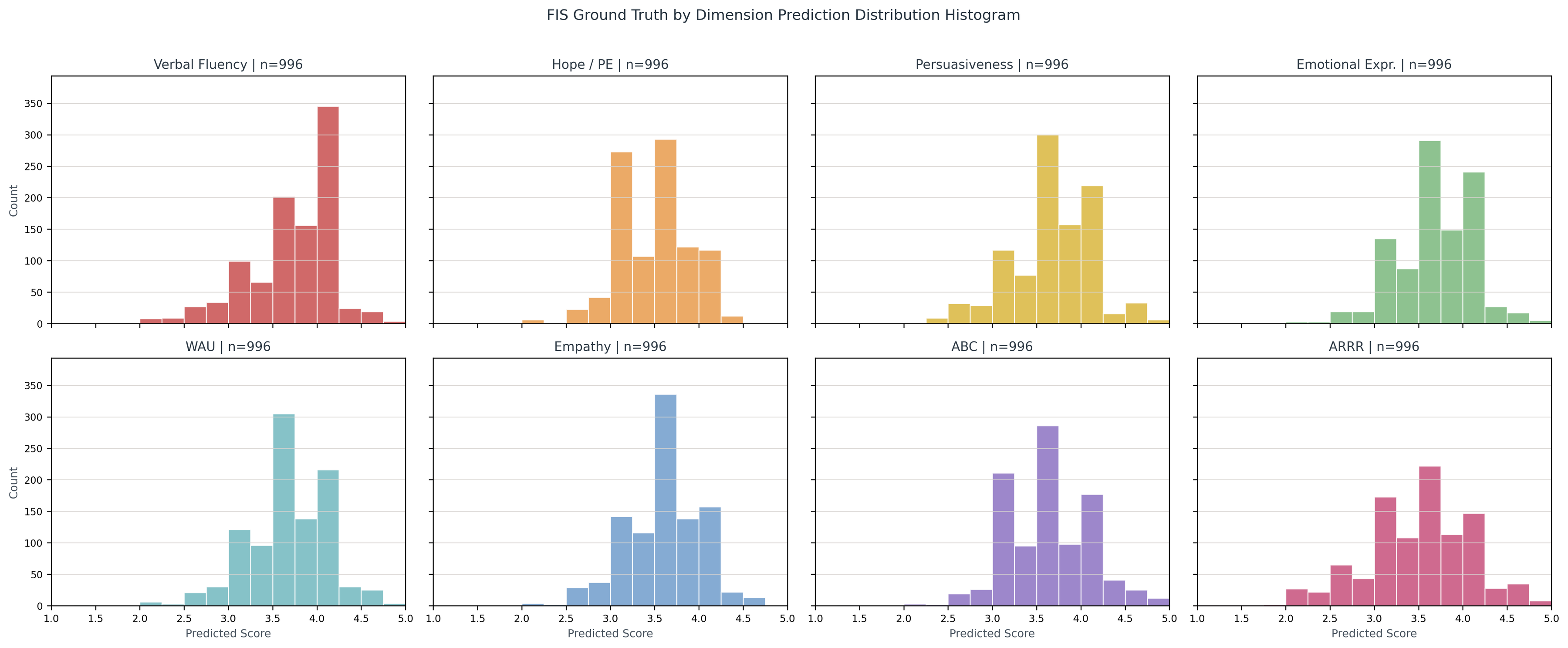}
\captionof{figure}{Gold-score histograms by dimension on the FIS dataset.}
\label{fig:fis_gold_hist_8dims}

\end{figure*}

\paragraph{Inter-dimension Correlation.}
Figure~\ref{fig:fis_inter_dim_corr} shows the Pearson correlation matrix across the eight FIS dimensions, computed from gold scores at the sample level.

\paragraph{Per-dimension Summary Statistics.}
Table~\ref{tab:fis_dim_stats} reports the mean, standard deviation, minimum, and maximum of gold scores for each FIS dimension.

\paragraph{Gold Score Distributions.}
Figure~\ref{fig:fis_gold_hist_8dims} shows the gold-score histograms for all eight FIS dimensions.

\section{Additional Experimental Results}
\label{sec:appendix-additional-results}

\subsection{MIS-RAFT Setup}
\label{sec:appendix-mis-raft-setup}

In our MIS-RAFT experiments, we fine-tune the base model Qwen/Qwen2.5-VL-7B-Instruct with 4-bit NF4 quantization and PEFT LoRA on two A100 GPUs. For both MIS-RAFT and the conventional SFT training objective, the model is trained for 2 epochs with a learning rate of $2\times10^{-5}$, per-device training batch size of 2, maximum sequence length of 512, and bfloat16 computation. We use the \texttt{paged\_adamw\_32bit} optimizer. The dataset is split at the \textbf{video level} into train, validation, and test sets with an 80/10/10 ratio, ensuring that no video appears in more than one split. For video preprocessing, we set the total pixel budget to $20480 \times 28 \times 28$ and the minimum pixel budget to $16 \times 28 \times 28$.

\clearpage
\onecolumn
\subsection{Additional Dimension-Level Heatmaps}
\label{sec:appendix-additional-heatmaps}

\includegraphics[width=\textwidth]{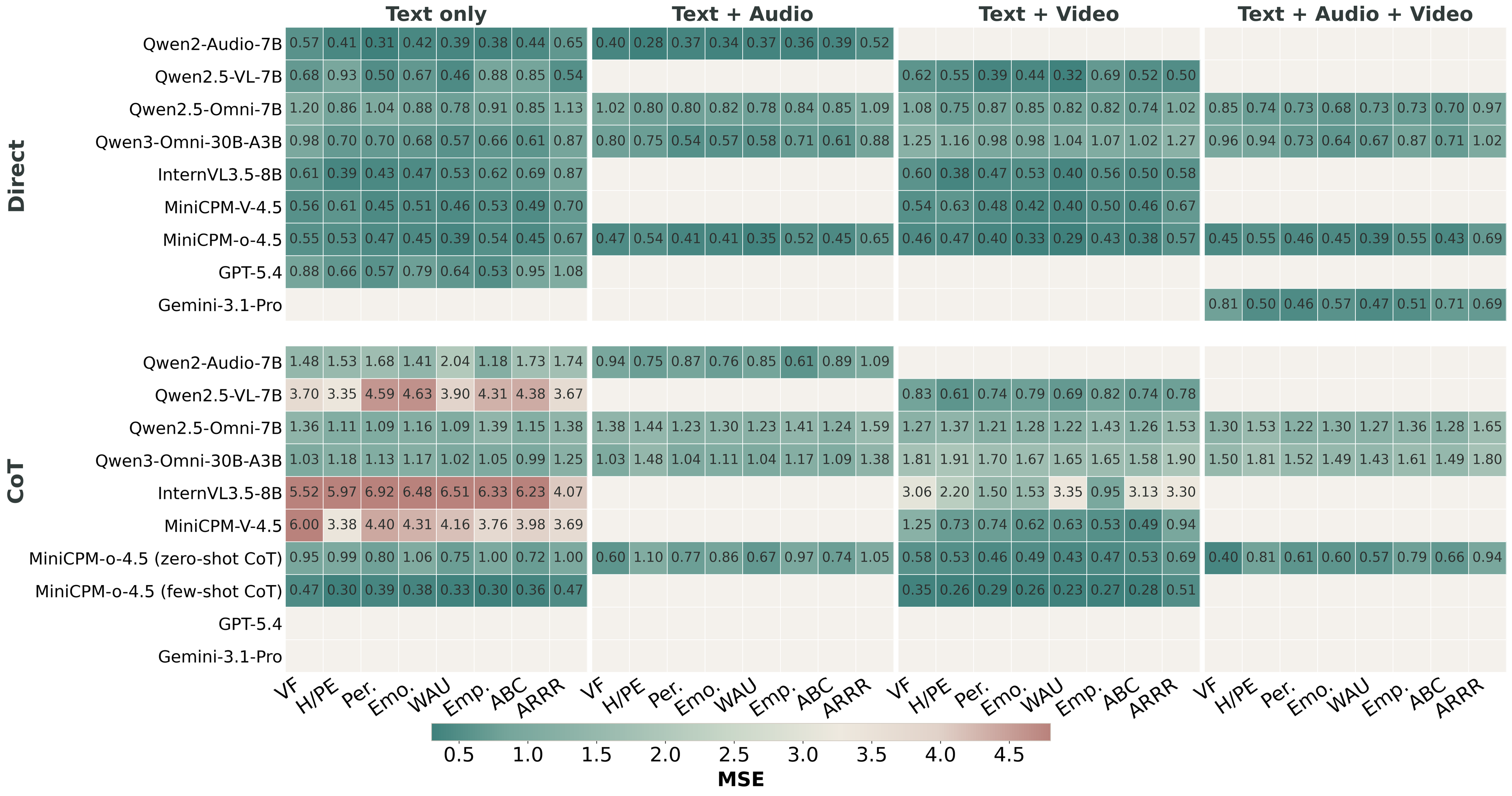}
    \captionof{figure}{\textbf{Dimension-level MSE heatmaps on MIS-Bench.} Lower MSE indicates smaller prediction error with respect to human ratings.}
    \label{fig:fisbench-dimension-mse}
\noindent
\includegraphics[width=\textwidth]{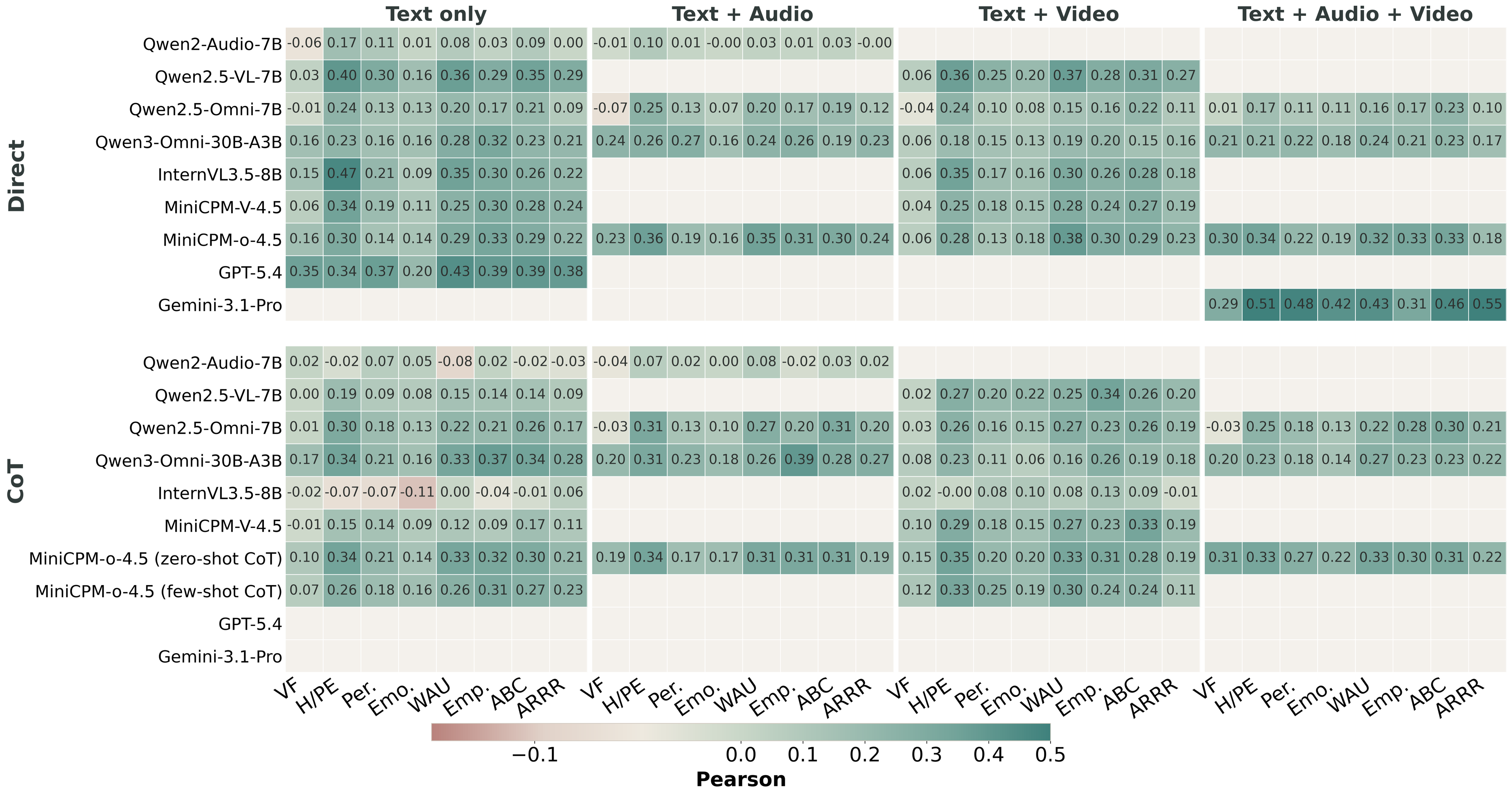}
\captionof{figure}{\textbf{Dimension-level Pearson heatmaps on MIS-Bench.}}
\label{fig:appendix-pearson-heatmap}

\medskip

\noindent
\includegraphics[width=\textwidth]{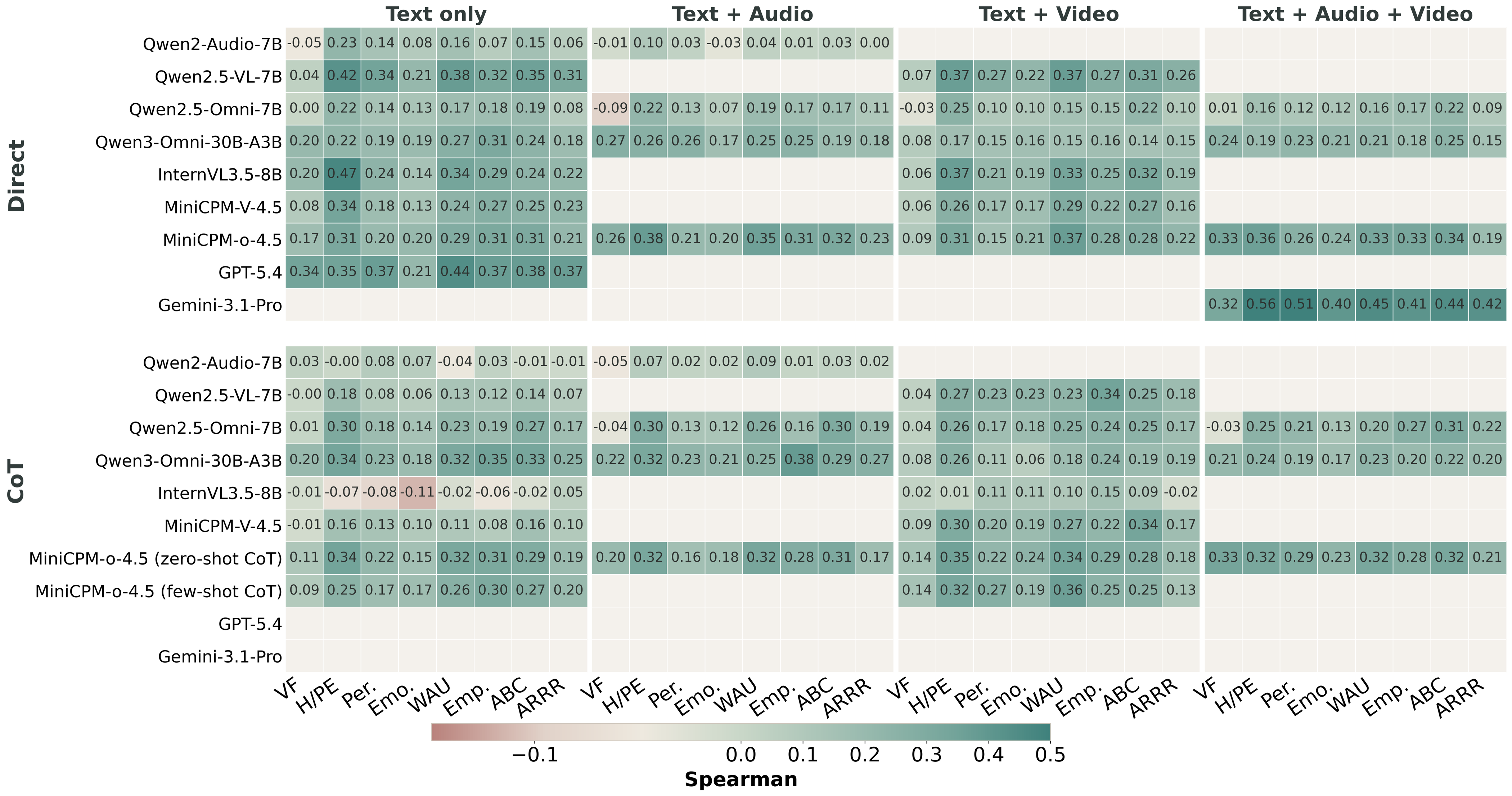}
\captionof{figure}{\textbf{Dimension-level Spearman heatmaps on MIS-Bench.}}
\label{fig:appendix-spearman-heatmap}

\medskip

\noindent
\includegraphics[width=\textwidth]{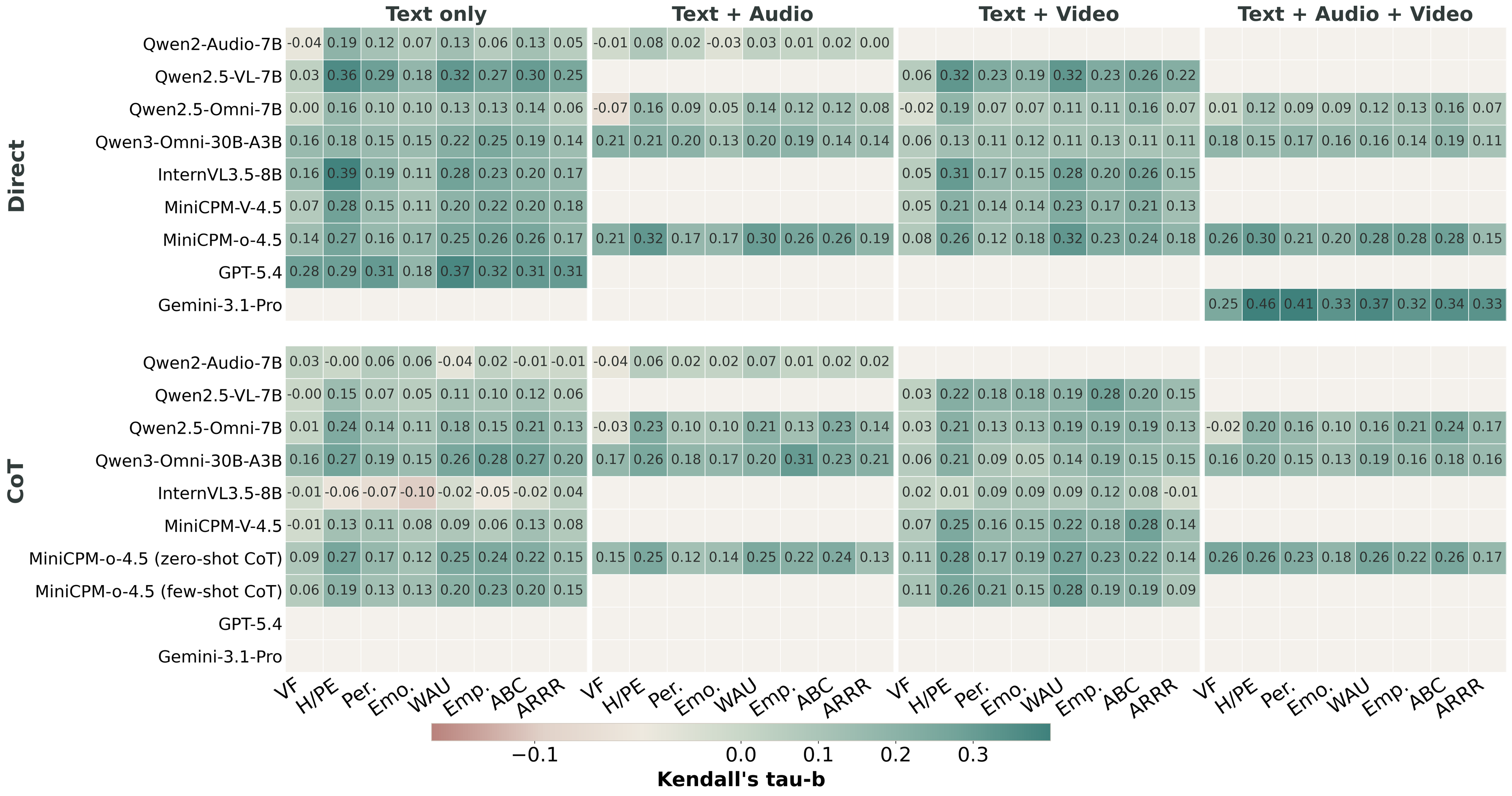}
\captionof{figure}{\textbf{Dimension-level Kendall's tau-b heatmaps on MIS-Bench.}}
\label{fig:appendix-kendall-heatmap}

\medskip

\noindent
\includegraphics[width=\textwidth]{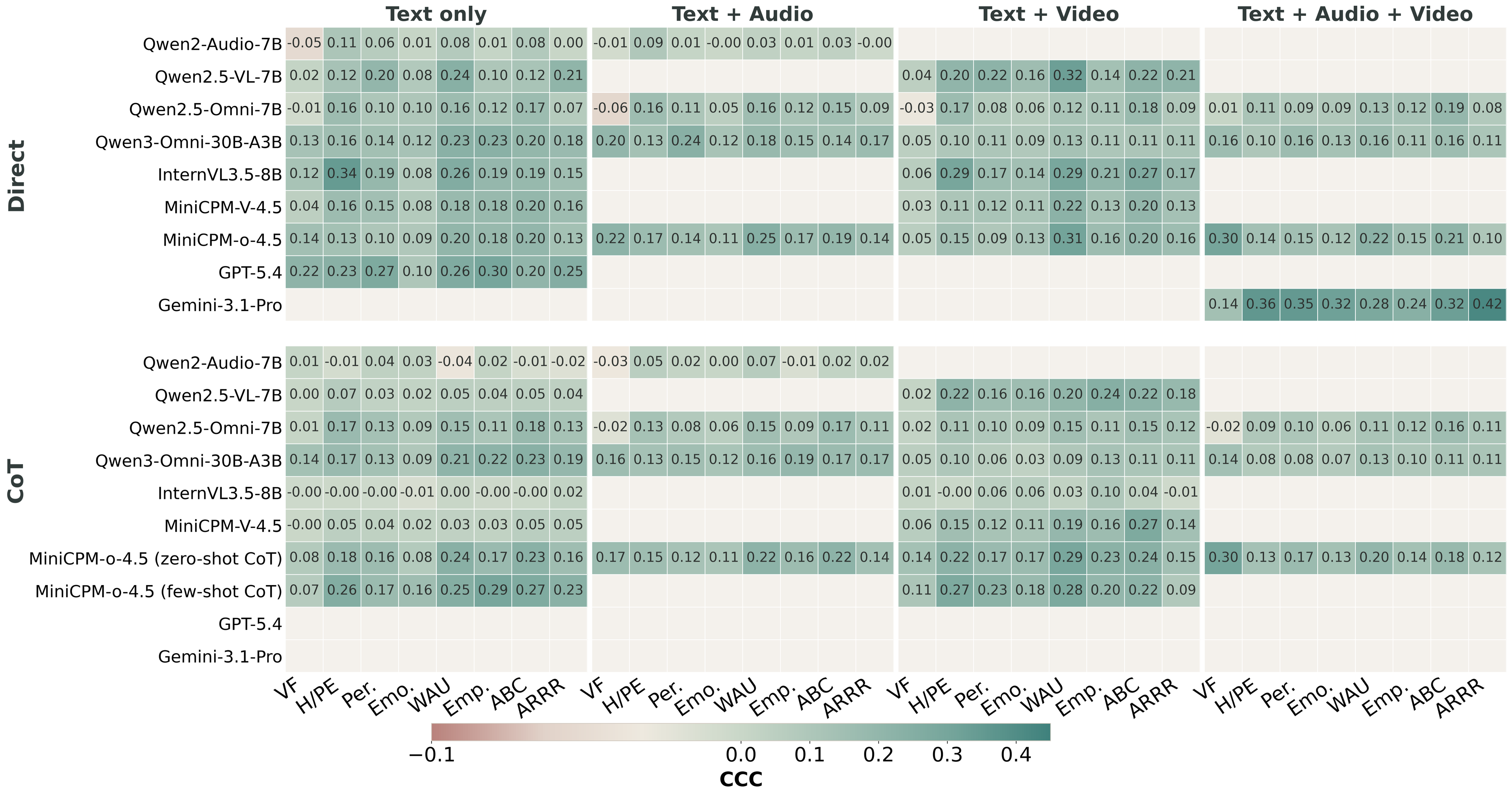}
\captionof{figure}{\textbf{Dimension-level CCC heatmaps on MIS-Bench.}}
\label{fig:appendix-ccc-heatmap}

\medskip

\noindent
\includegraphics[width=\textwidth]{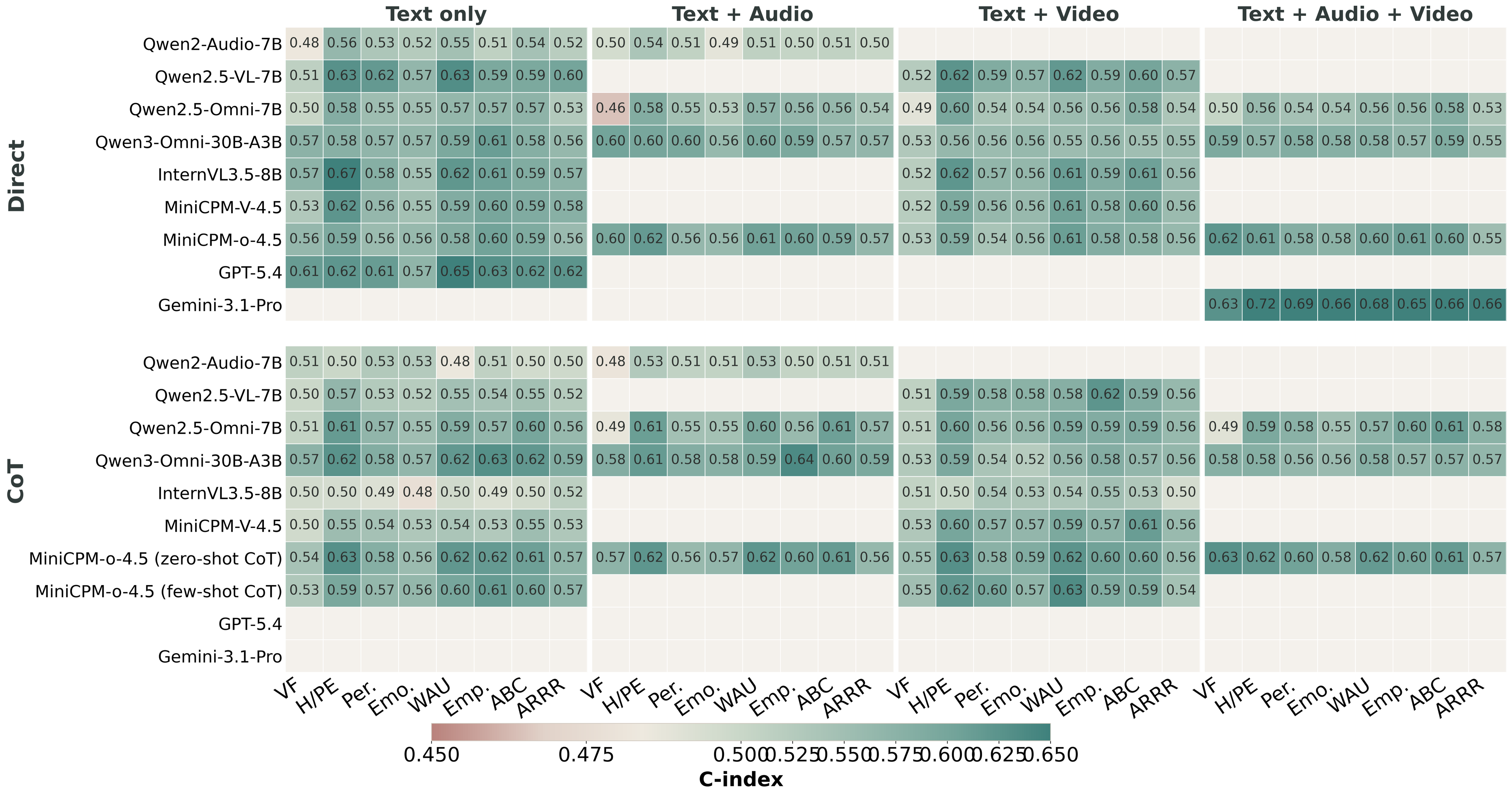}
\captionof{figure}{\textbf{Dimension-level C-index heatmaps on MIS-Bench.}}
\label{fig:appendix-cindex-heatmap}

\newcommand{\modelpanelhalf}[2]{
\begin{minipage}[t]{0.5\textwidth}
    \centering
    \includegraphics[width=\linewidth]{#1}
    
    {\small #2}
\end{minipage}
}

\newcommand{\histfull}[2]{
\noindent
\begin{minipage}[t]{\textwidth}
    \centering
    \includegraphics[width=\linewidth]{#1}
    
    {\small #2}
\end{minipage}

}

\subsection{Model-wise Radar Plots}
\label{sec:appendix-radar-plots}

\subsubsection{ICC Radar Plots}

\noindent
\modelpanelhalf{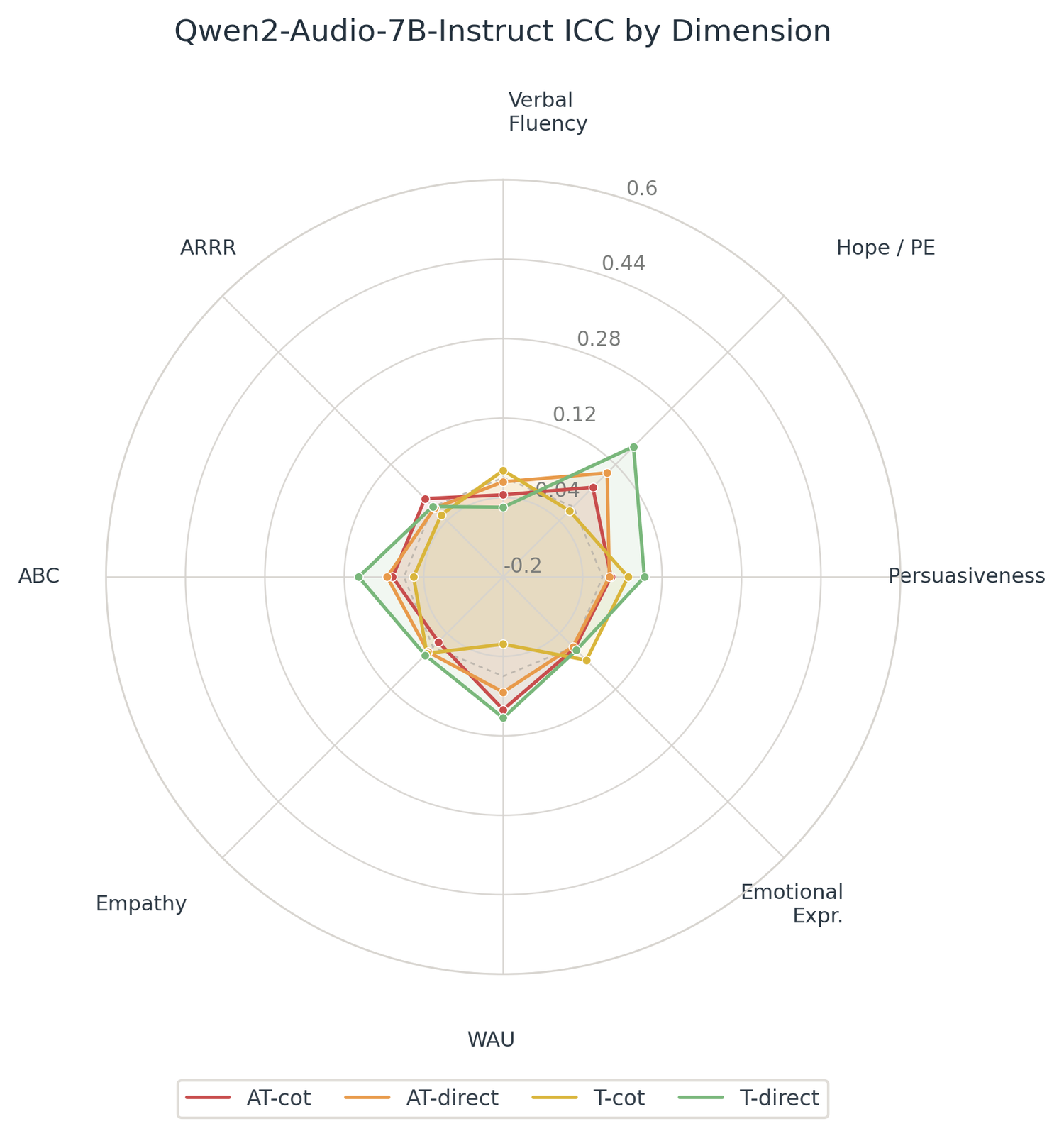}{Qwen2-Audio-7B-Instruct}%
\modelpanelhalf{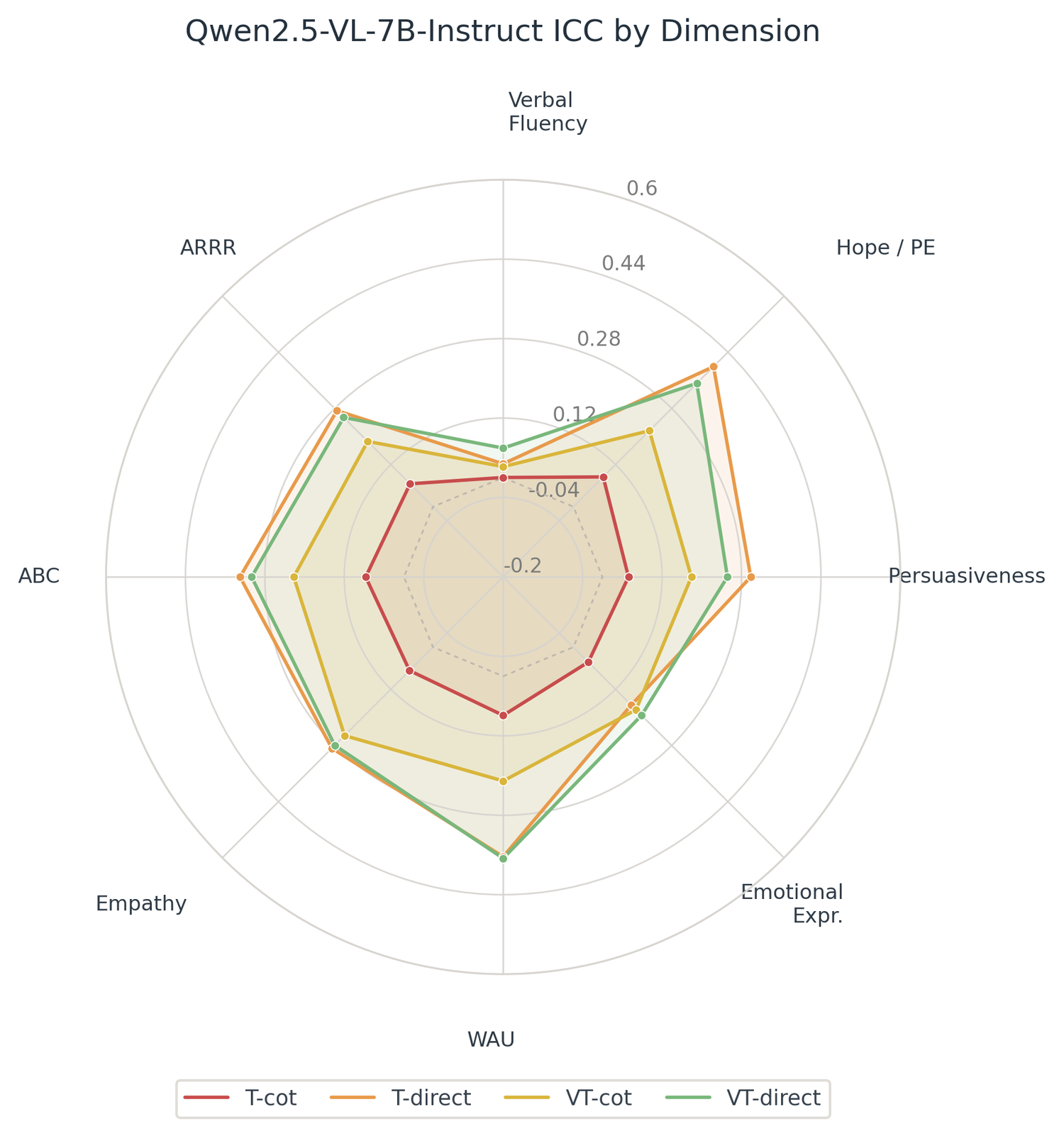}{Qwen2.5-VL-7B-Instruct}

\medskip
\noindent
\modelpanelhalf{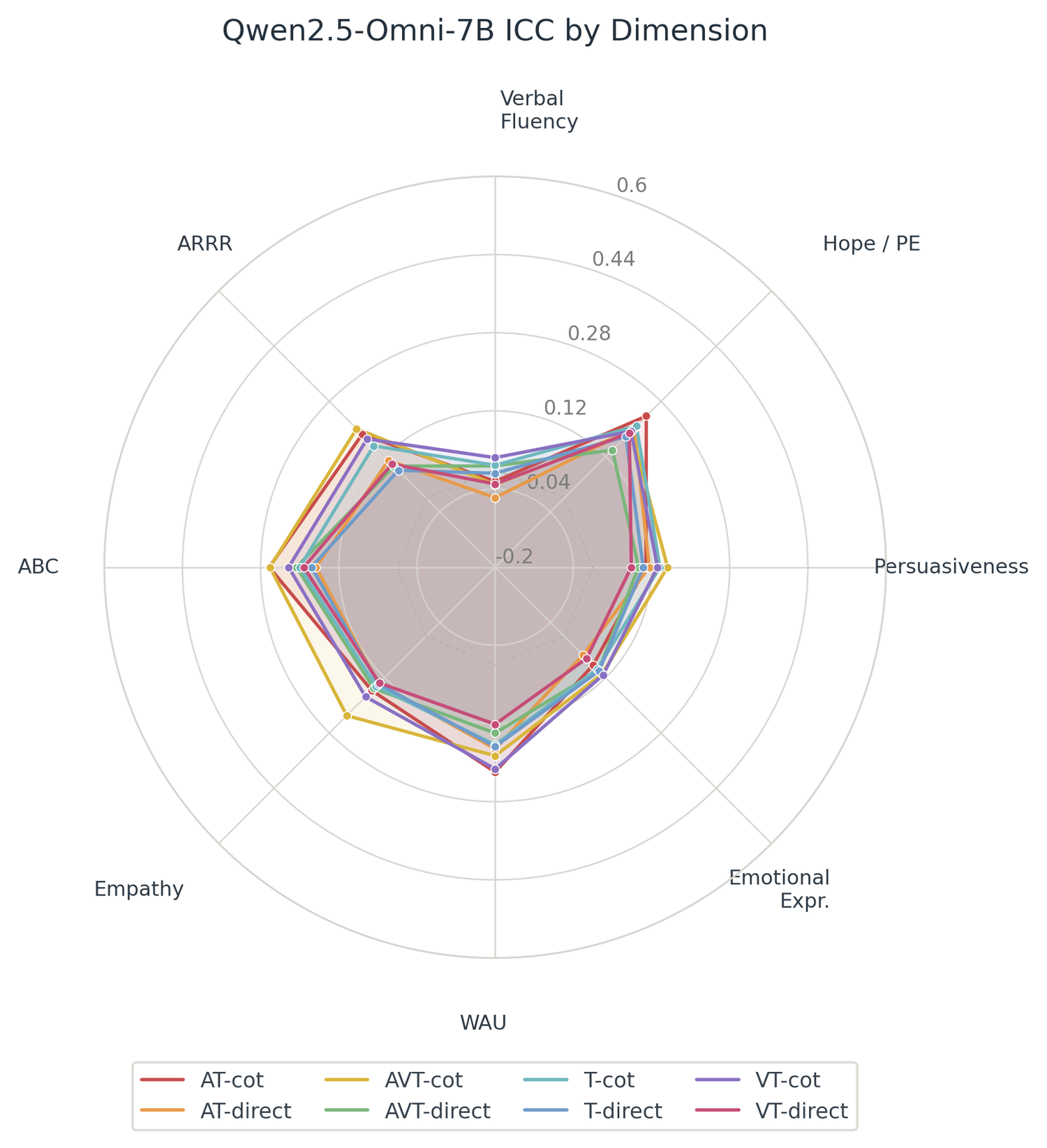}{Qwen2.5-Omni-7B}%
\modelpanelhalf{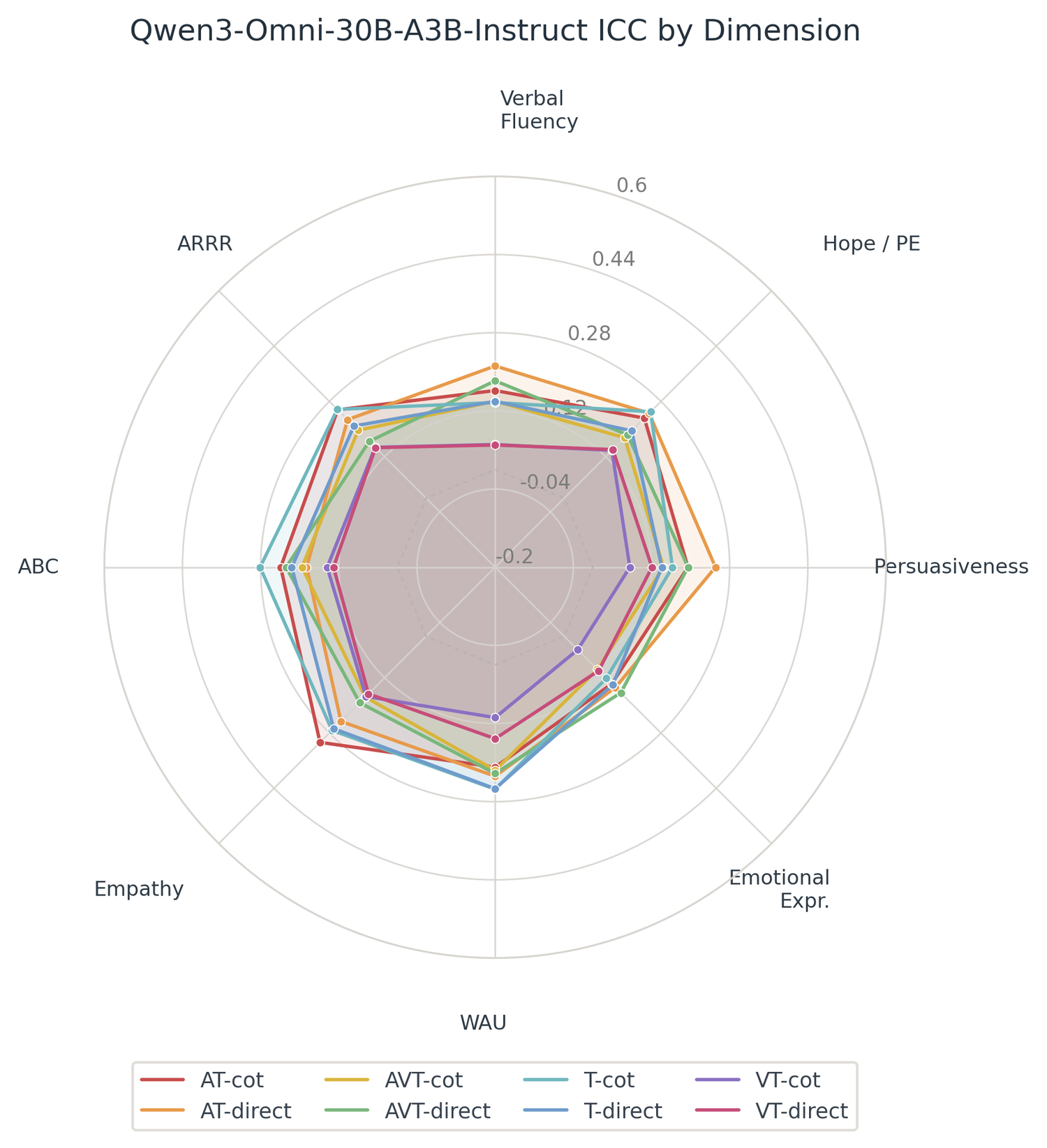}{Qwen3-Omni-30B-A3B-Instruct}

\medskip
\noindent
\modelpanelhalf{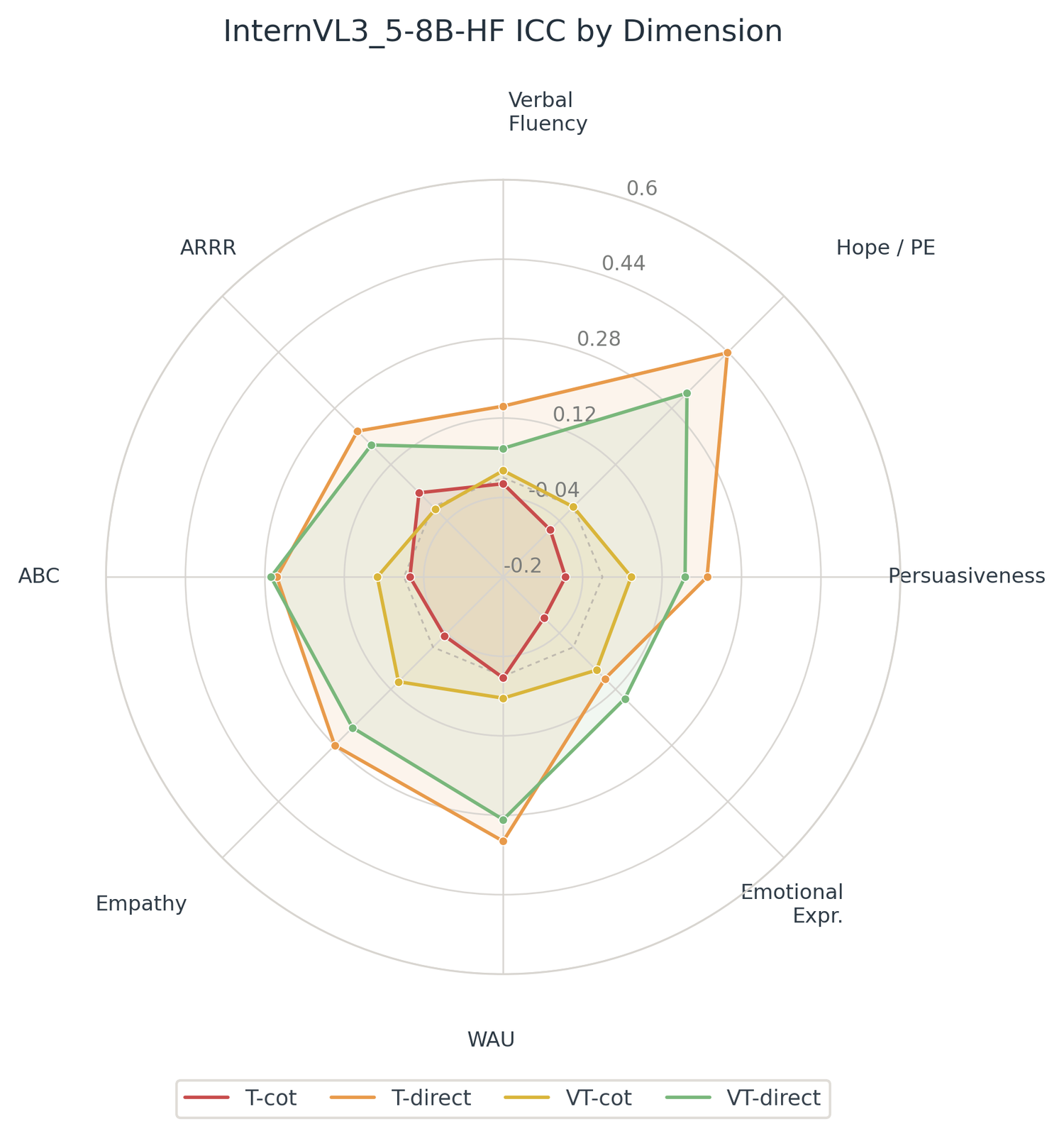}{InternVL3.5-8B-HF}%
\modelpanelhalf{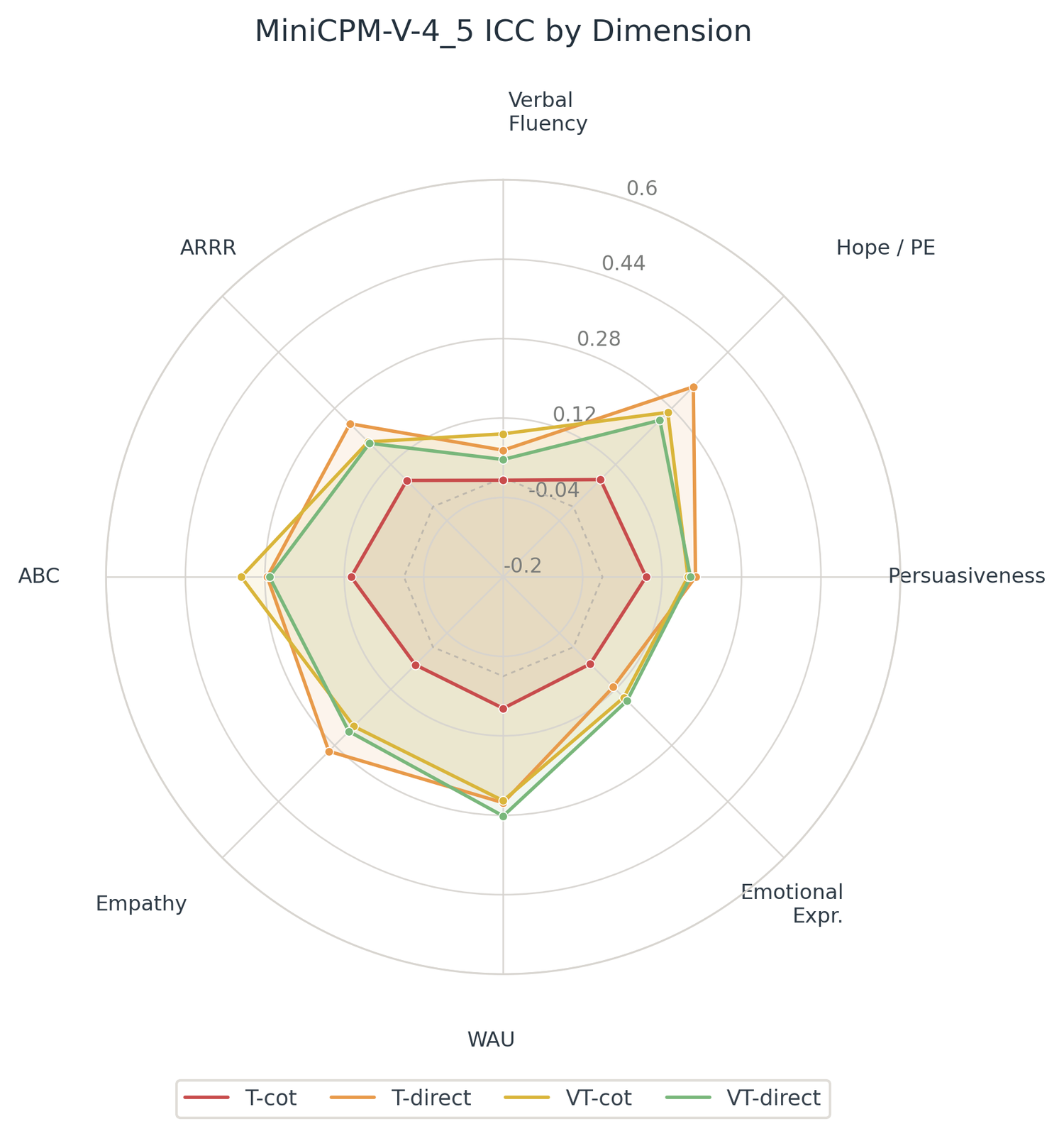}{MiniCPM-V-4.5}

\medskip
\noindent
\modelpanelhalf{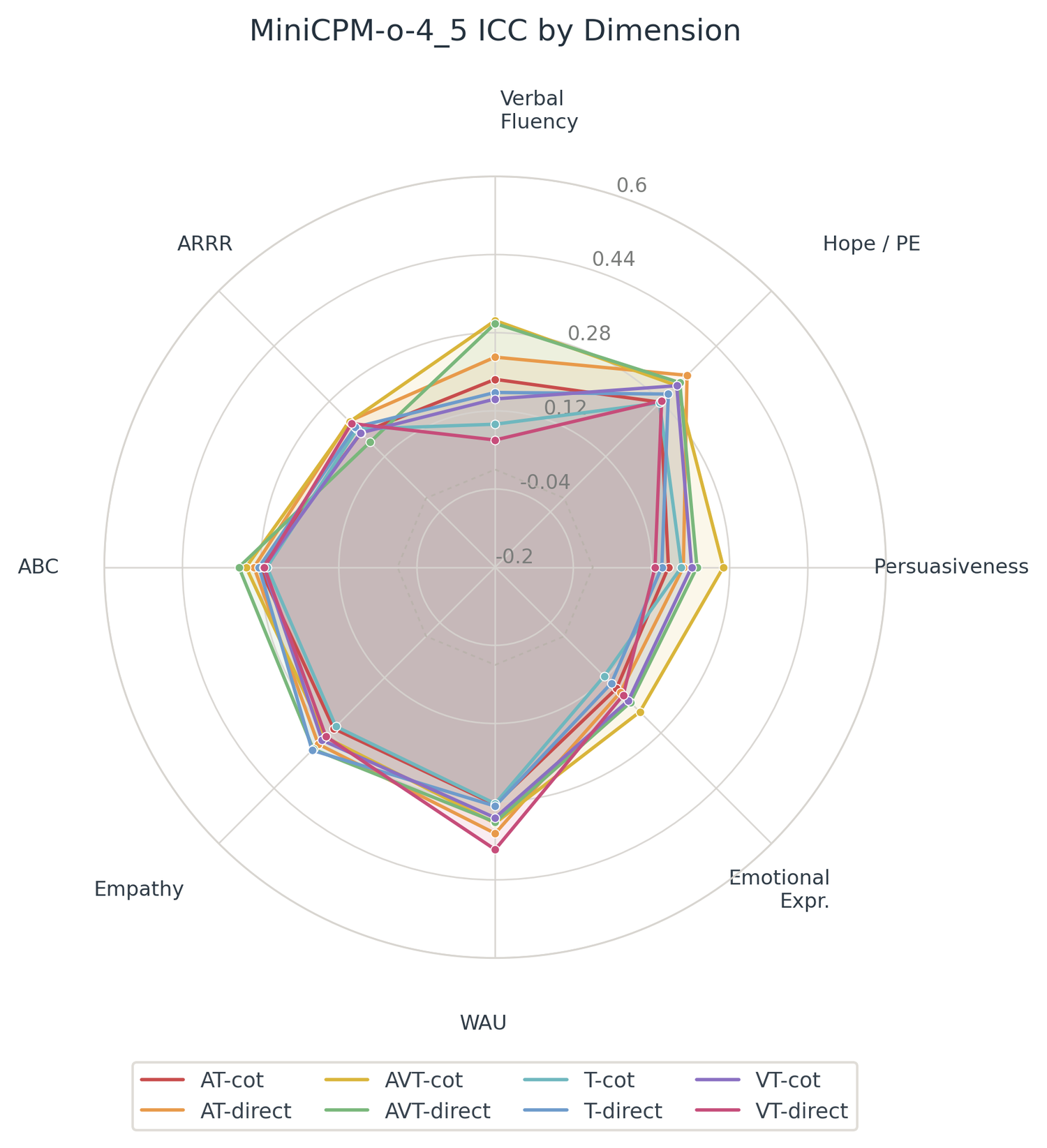}{MiniCPM-o-4.5}%
\modelpanelhalf{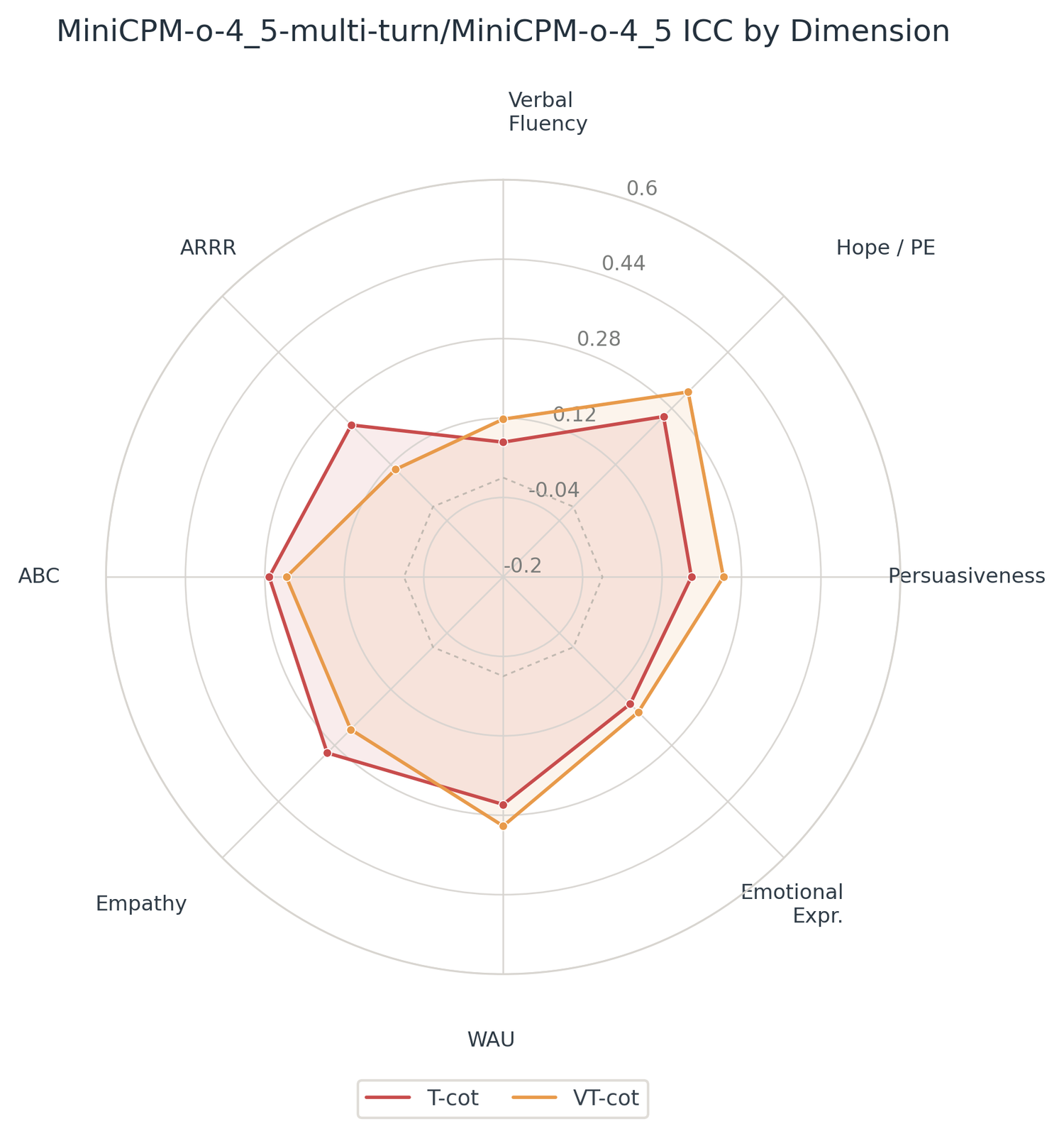}{MiniCPM-o-4.5 multi-turn}

\medskip
\noindent
\modelpanelhalf{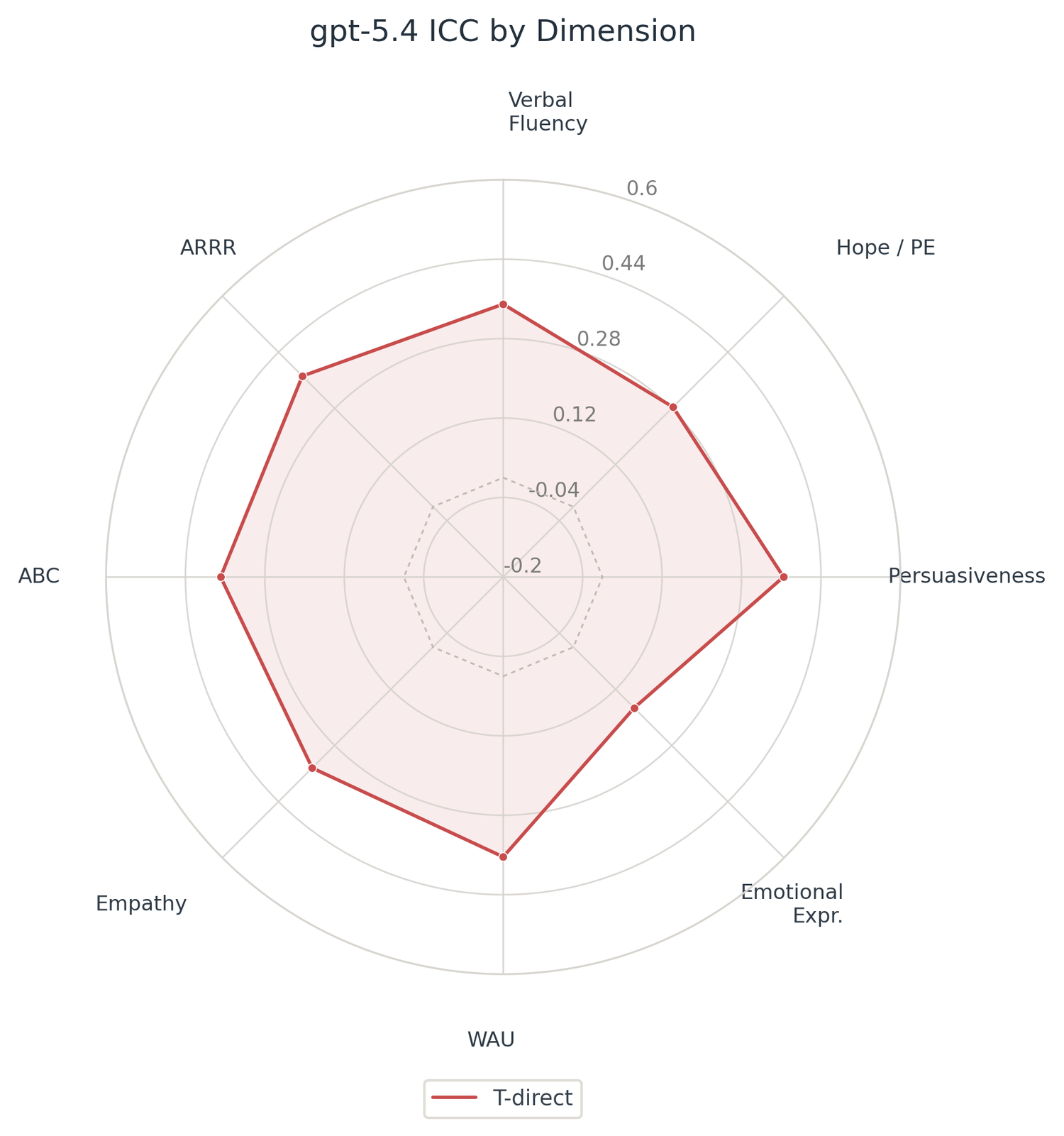}{GPT-5.4}%
\modelpanelhalf{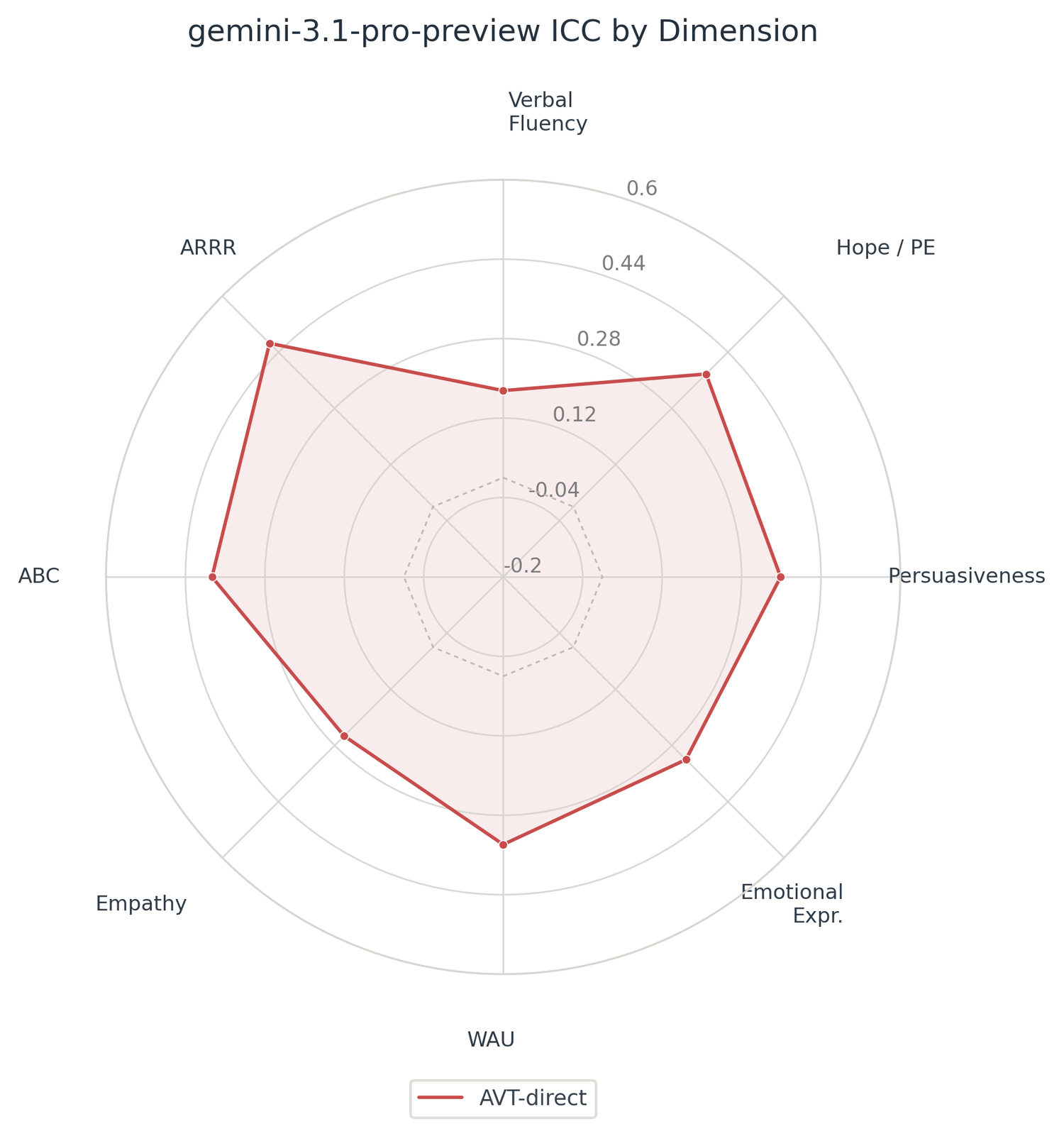}{Gemini-3.1-Pro-Preview}

\captionof{figure}{\textbf{Model-wise ICC radar plots.}}
\label{fig:appendix-icc-radars}

\Needspace{0.4\textheight}
\subsubsection{MSE Radar Plots}

\noindent
\modelpanelhalf{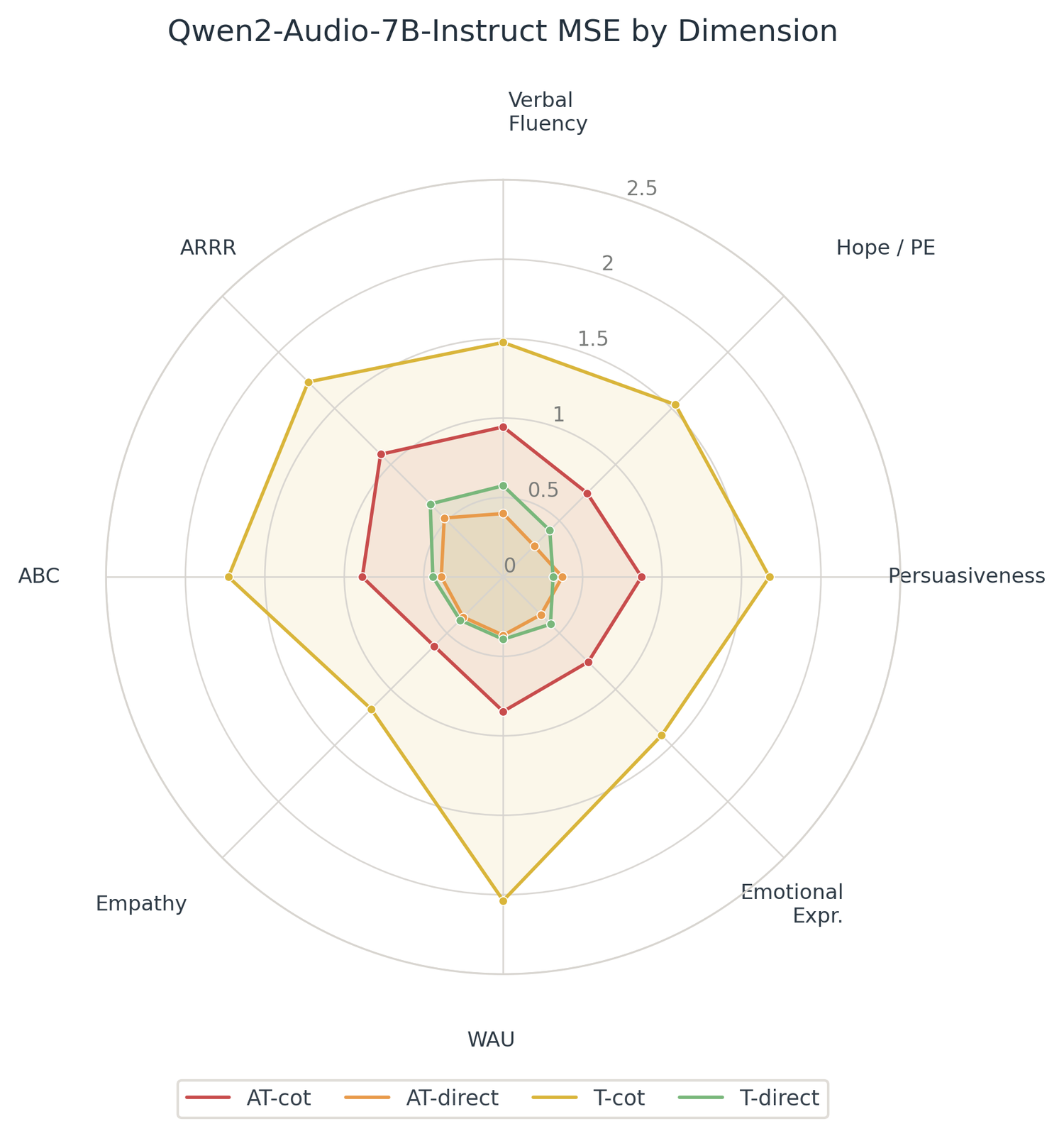}{Qwen2-Audio-7B-Instruct}%
\modelpanelhalf{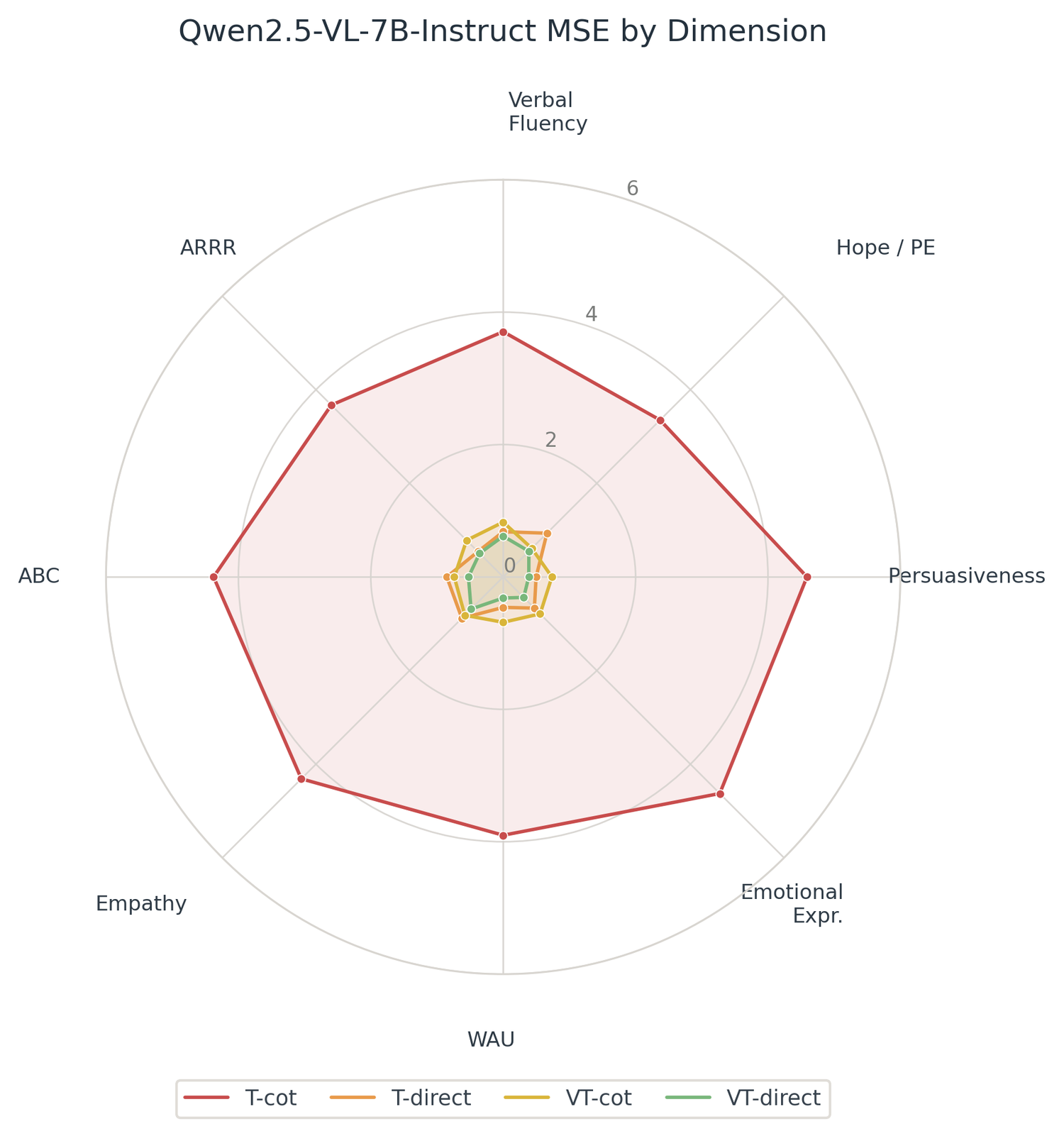}{Qwen2.5-VL-7B-Instruct}

\medskip
\noindent
\modelpanelhalf{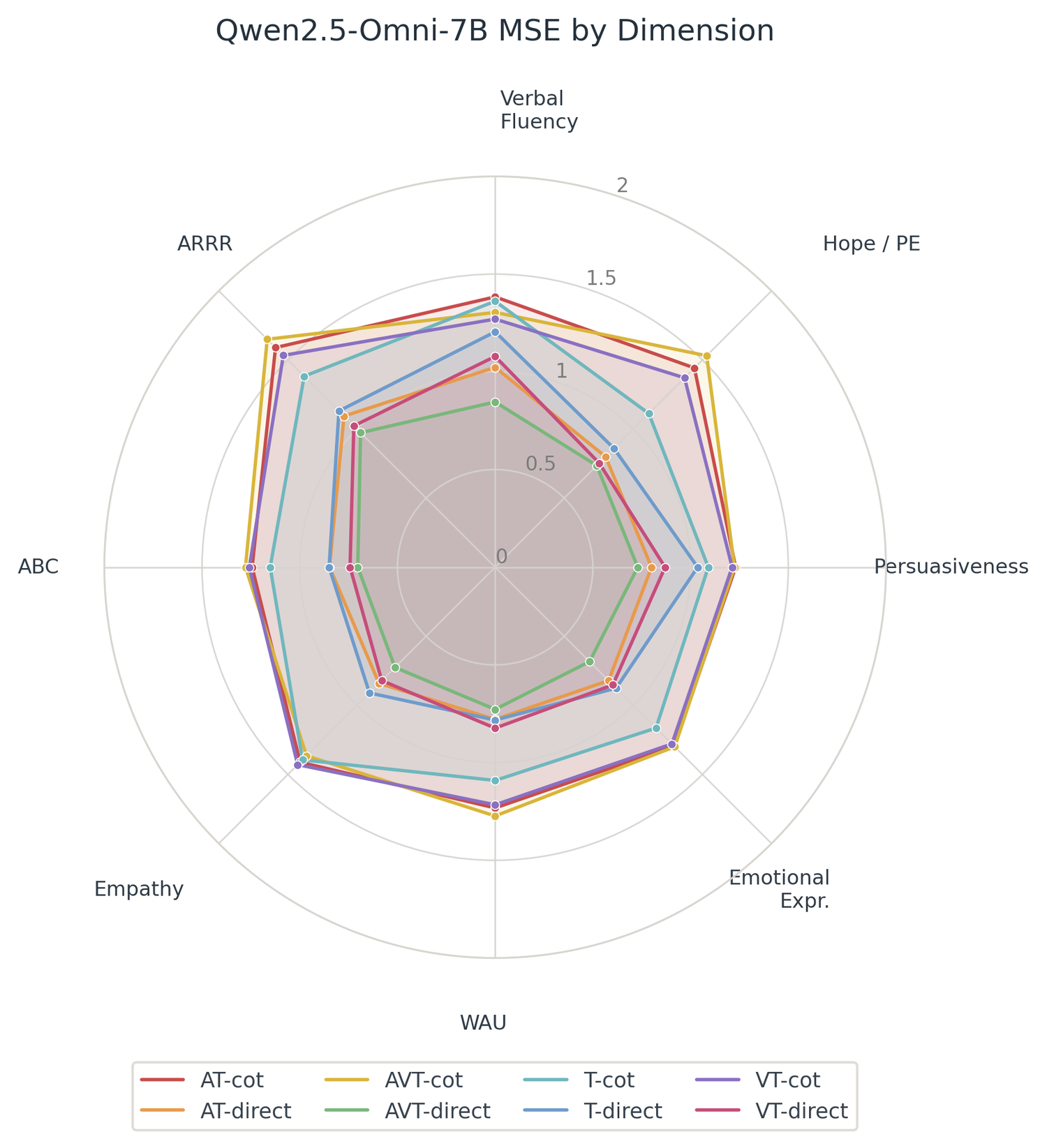}{Qwen2.5-Omni-7B}%
\modelpanelhalf{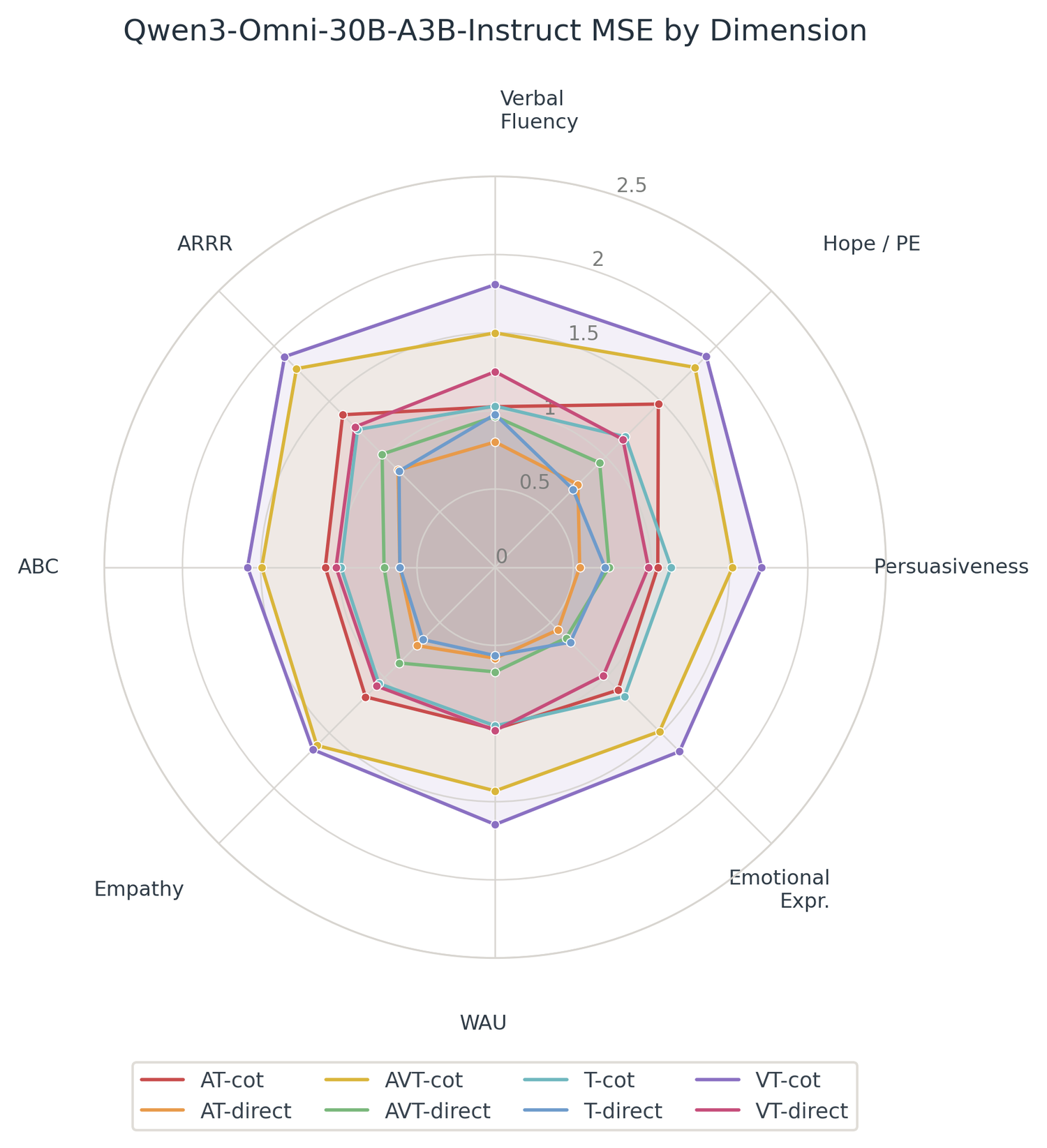}{Qwen3-Omni-30B-A3B-Instruct}

\medskip
\noindent
\modelpanelhalf{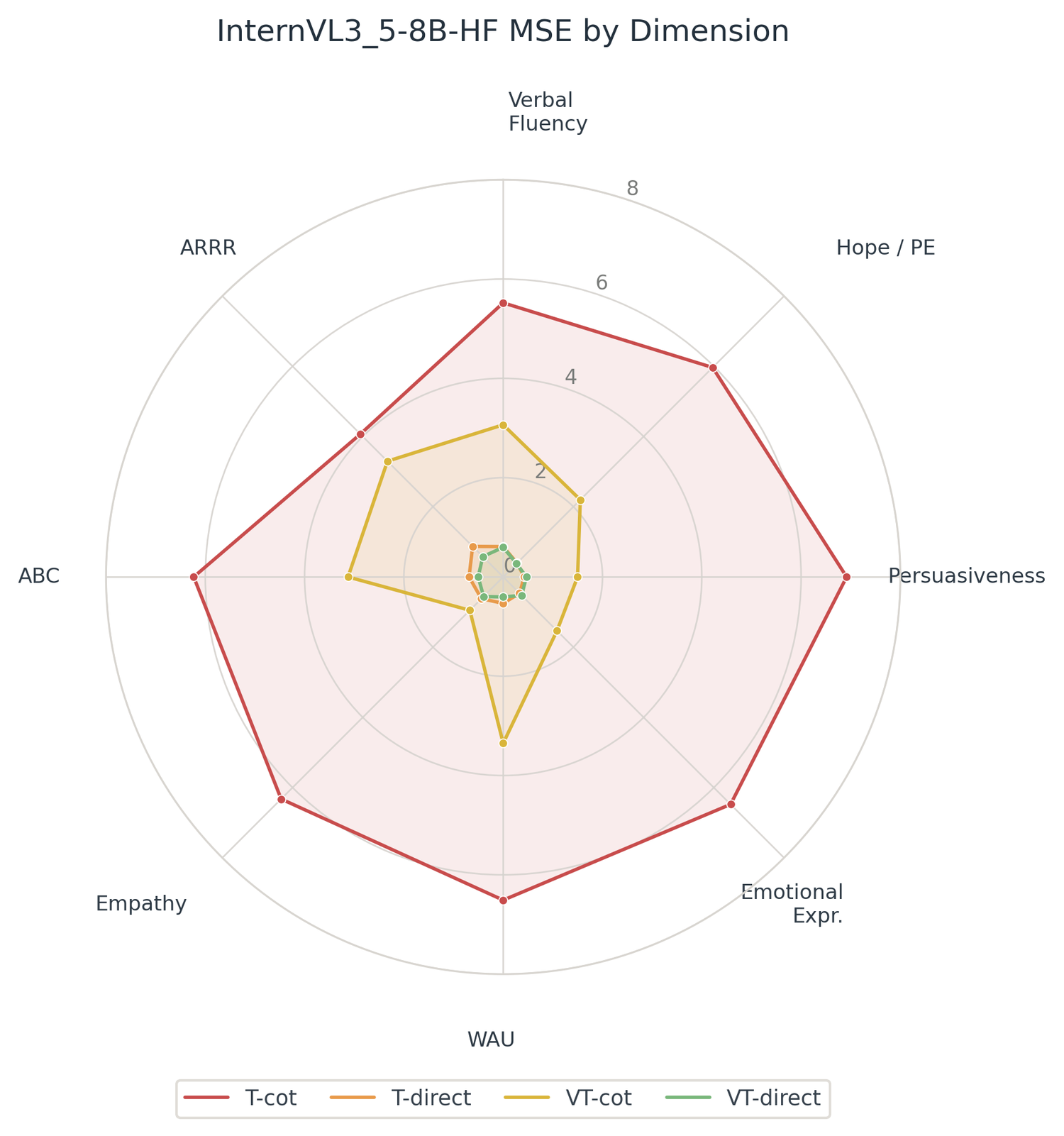}{InternVL3.5-8B-HF}%
\modelpanelhalf{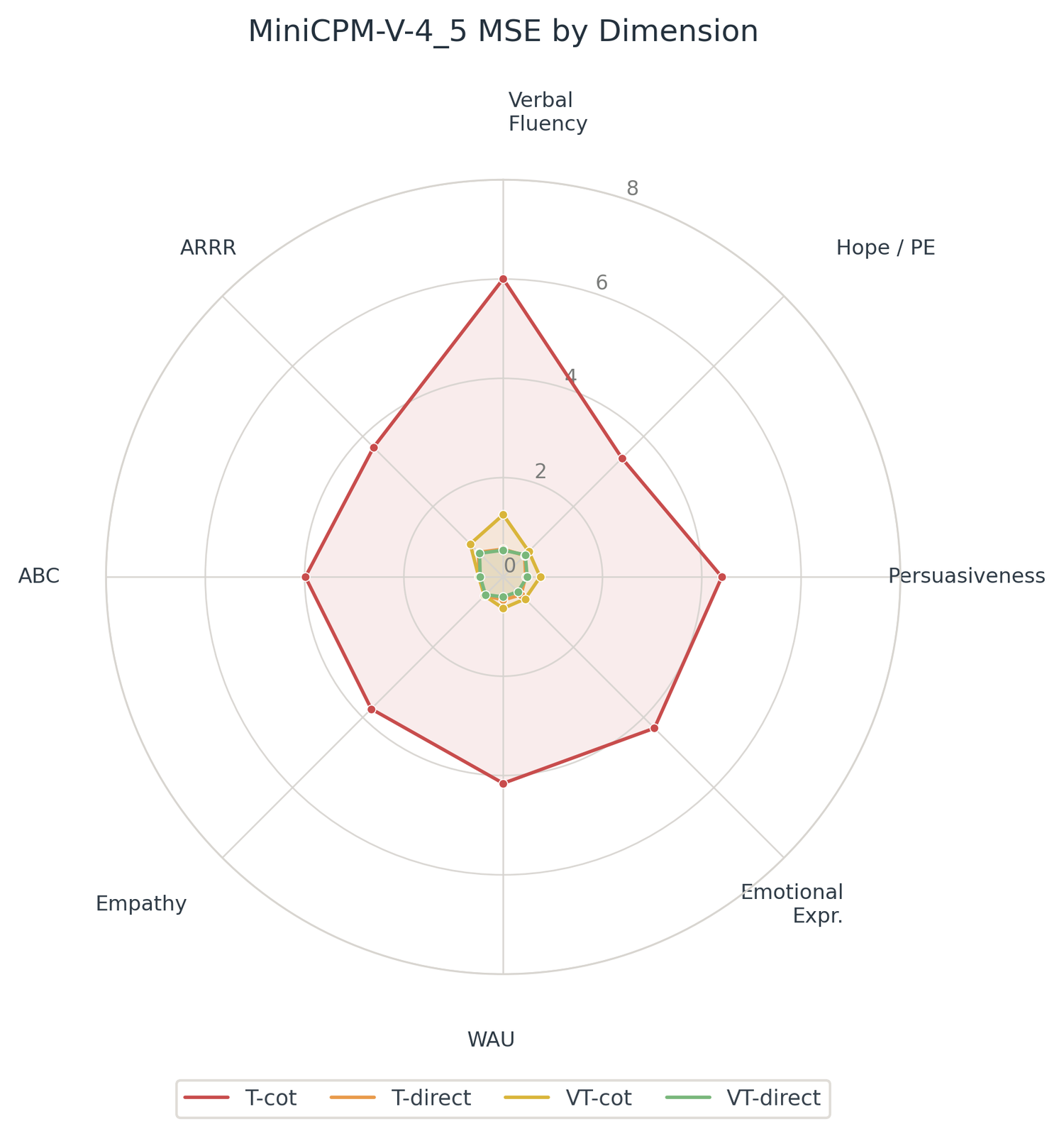}{MiniCPM-V-4.5}

\medskip
\noindent
\modelpanelhalf{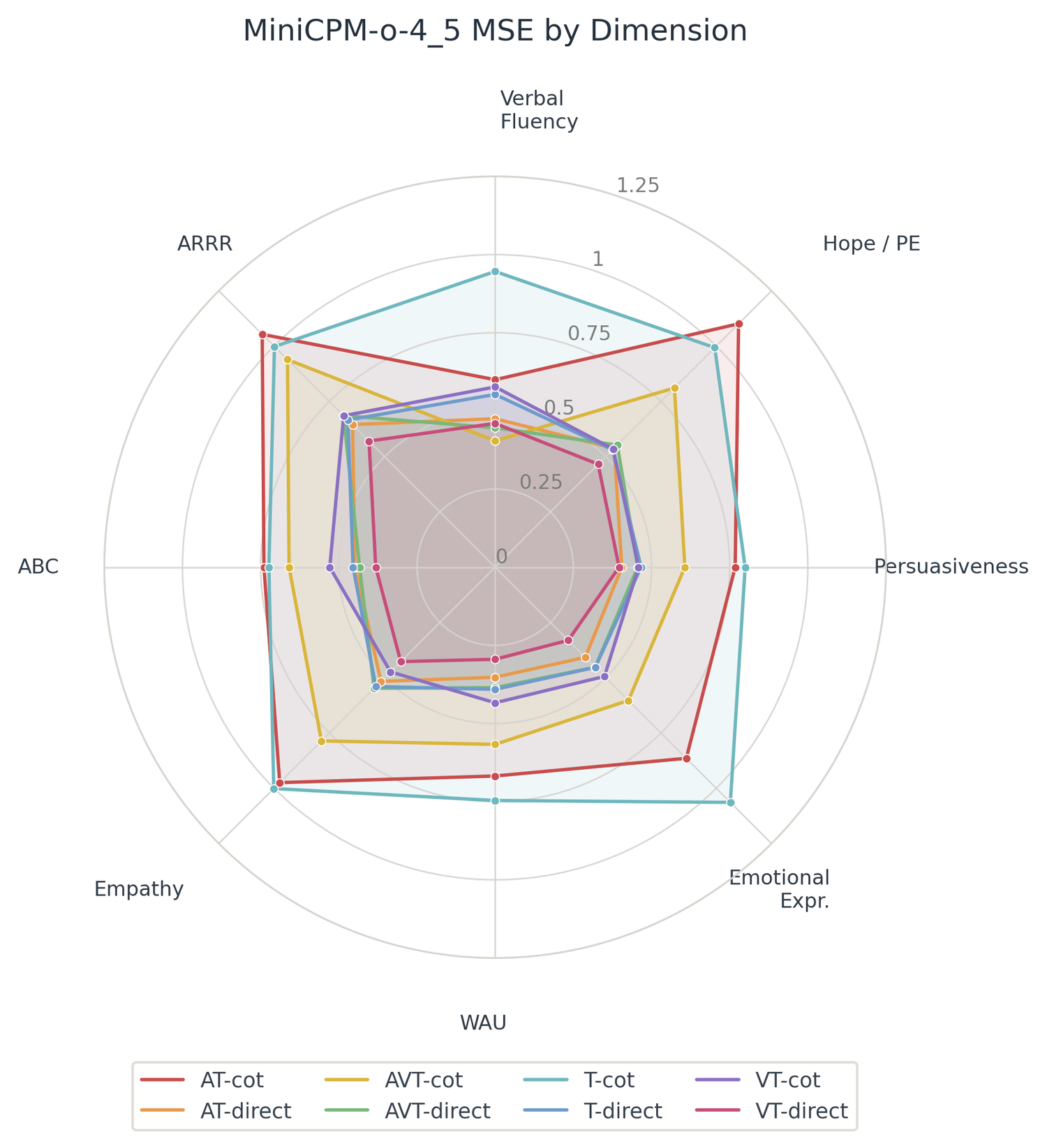}{MiniCPM-o-4.5}%
\modelpanelhalf{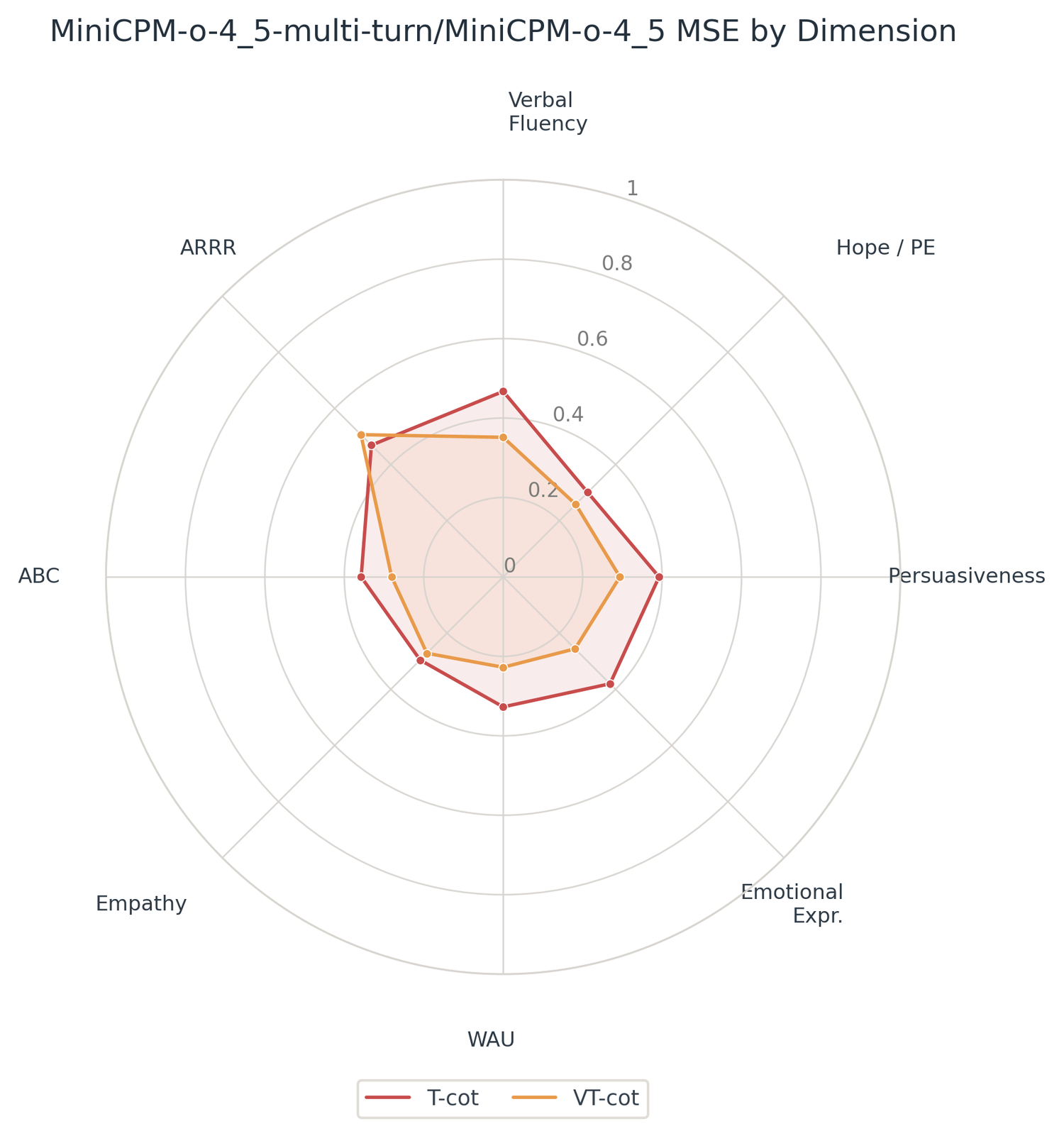}{MiniCPM-o-4.5 multi-turn}

\medskip
\noindent
\modelpanelhalf{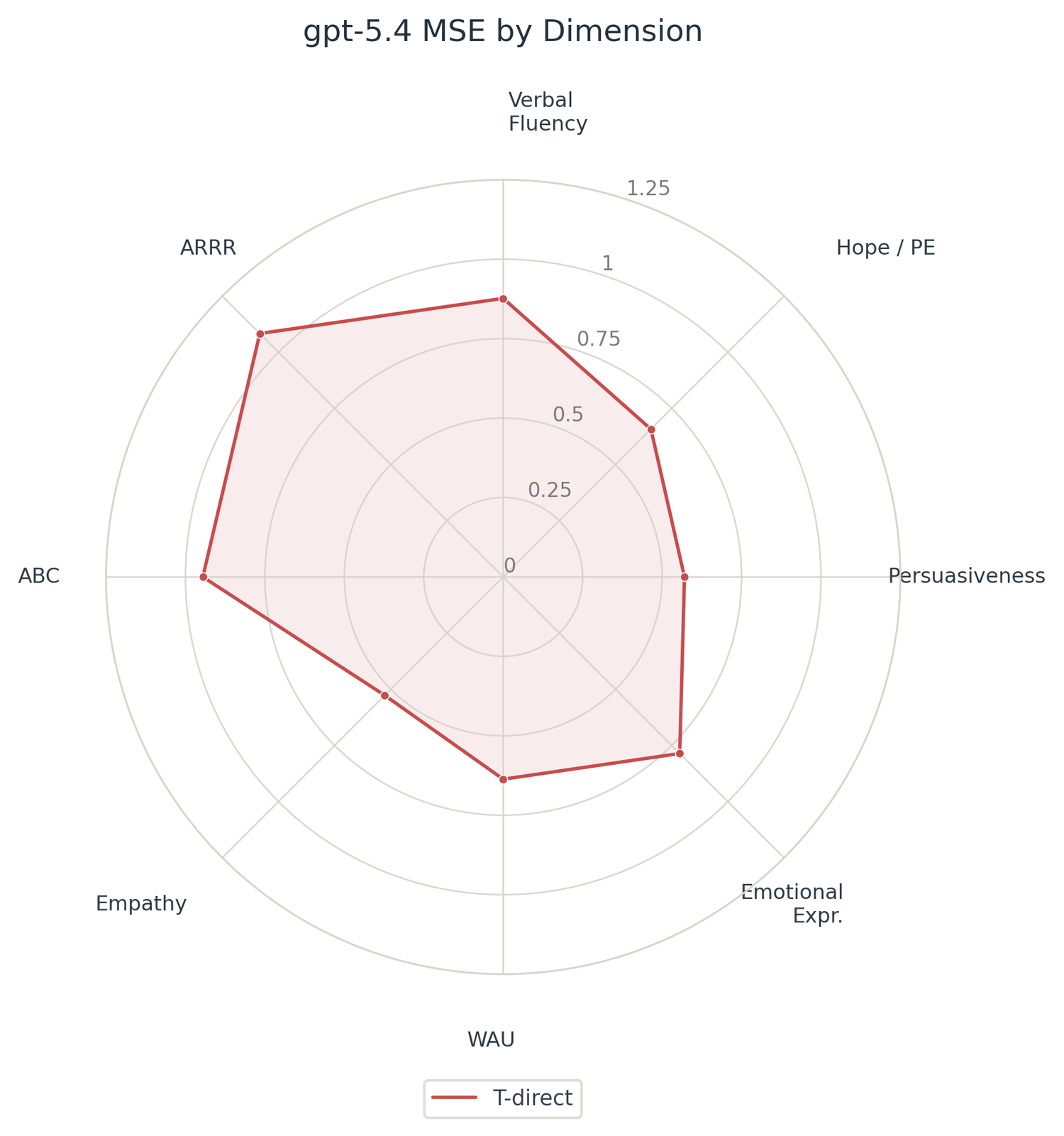}{GPT-5.4}%
\modelpanelhalf{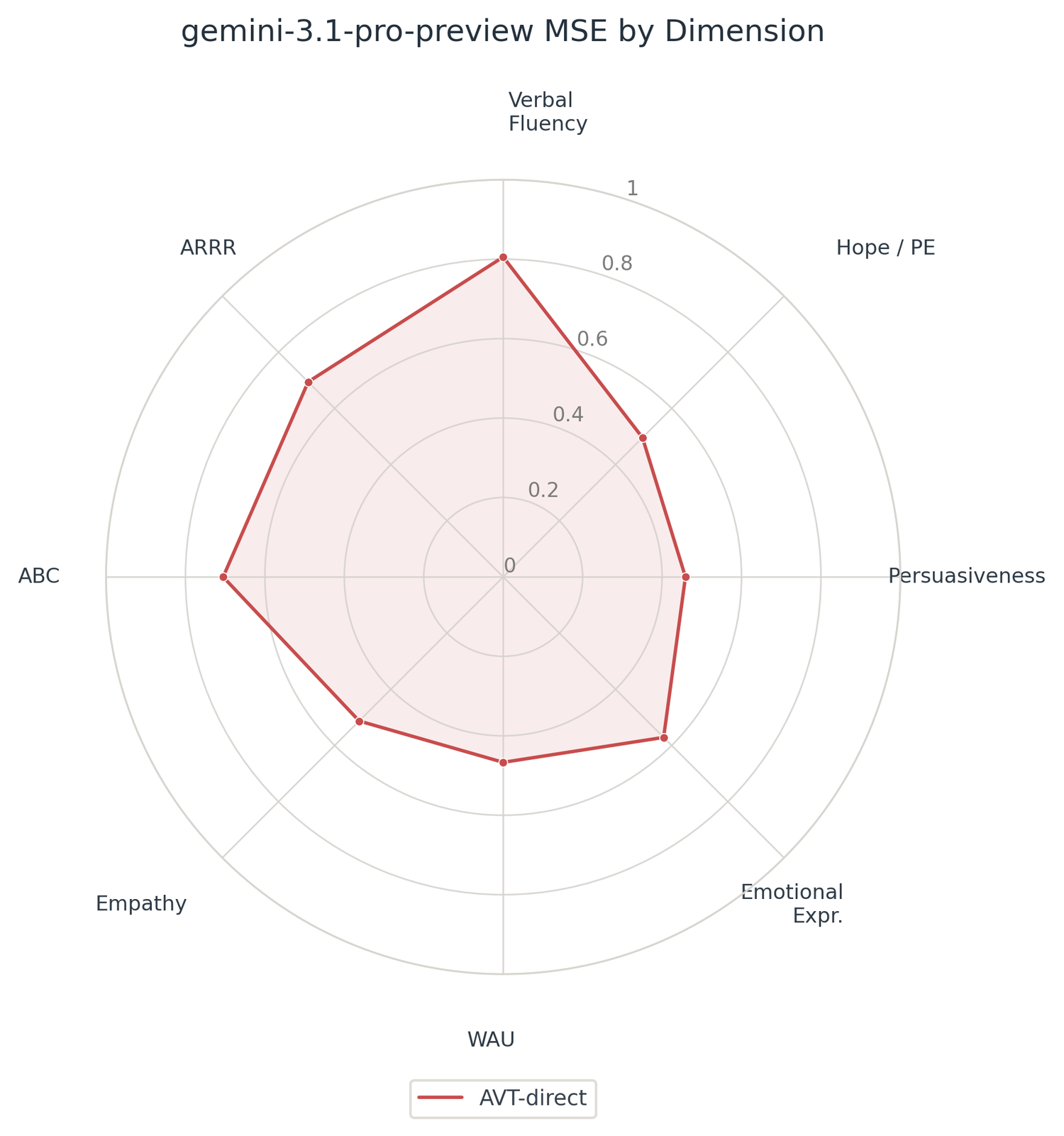}{Gemini-3.1-Pro-Preview}

\captionof{figure}{\textbf{Model-wise MSE radar plots.}}
\label{fig:appendix-mse-radars}

\Needspace{0.4\textheight}
\subsection{Prediction Histograms}
\label{sec:appendix-histograms}

\histfull{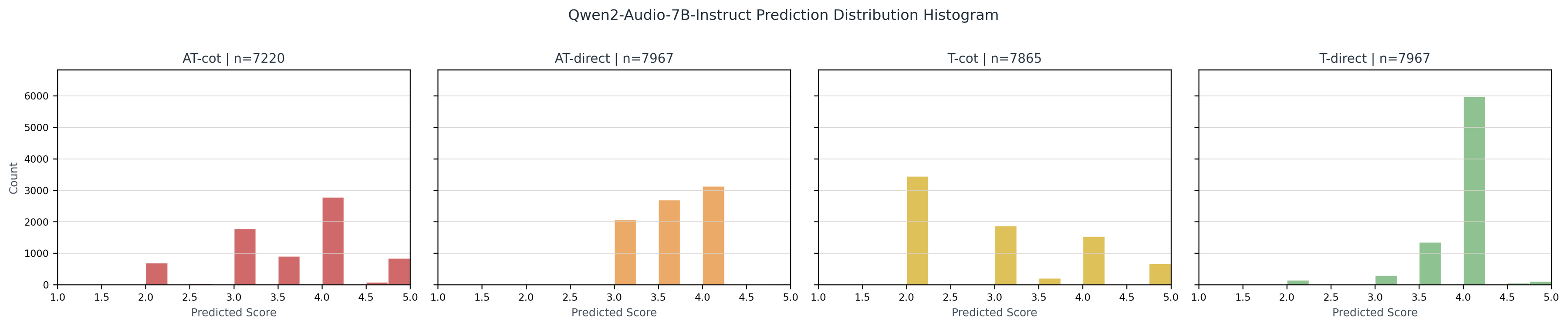}{Qwen2-Audio-7B-Instruct}

\histfull{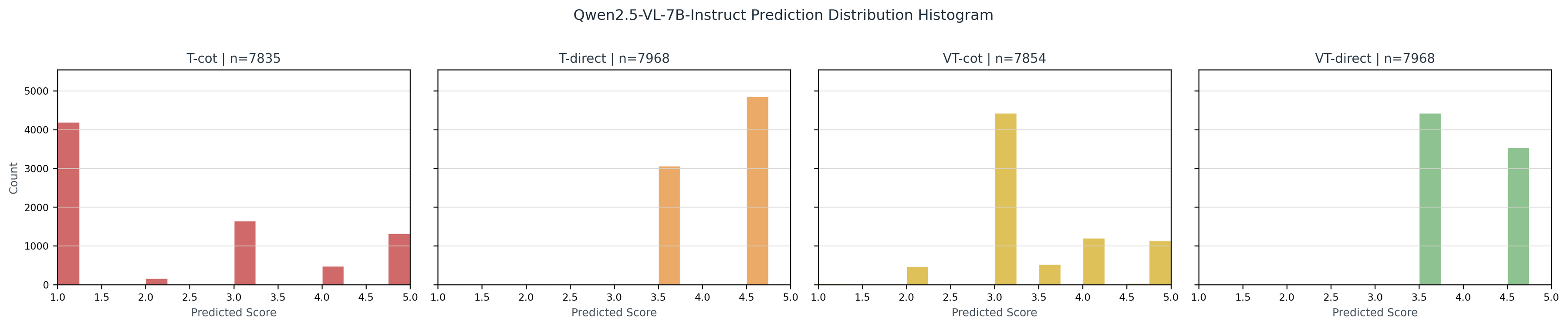}{Qwen2.5-VL-7B-Instruct}

\histfull{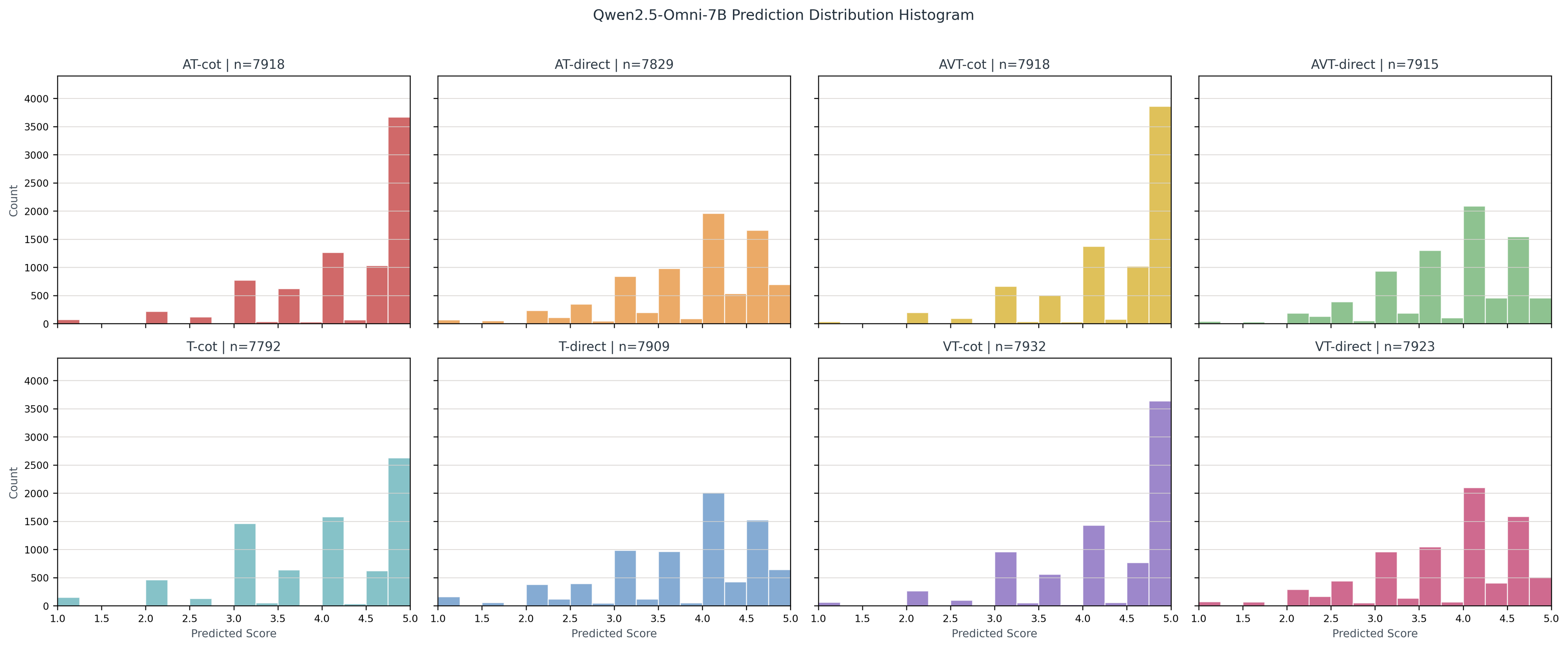}{Qwen2.5-Omni-7B}

\histfull{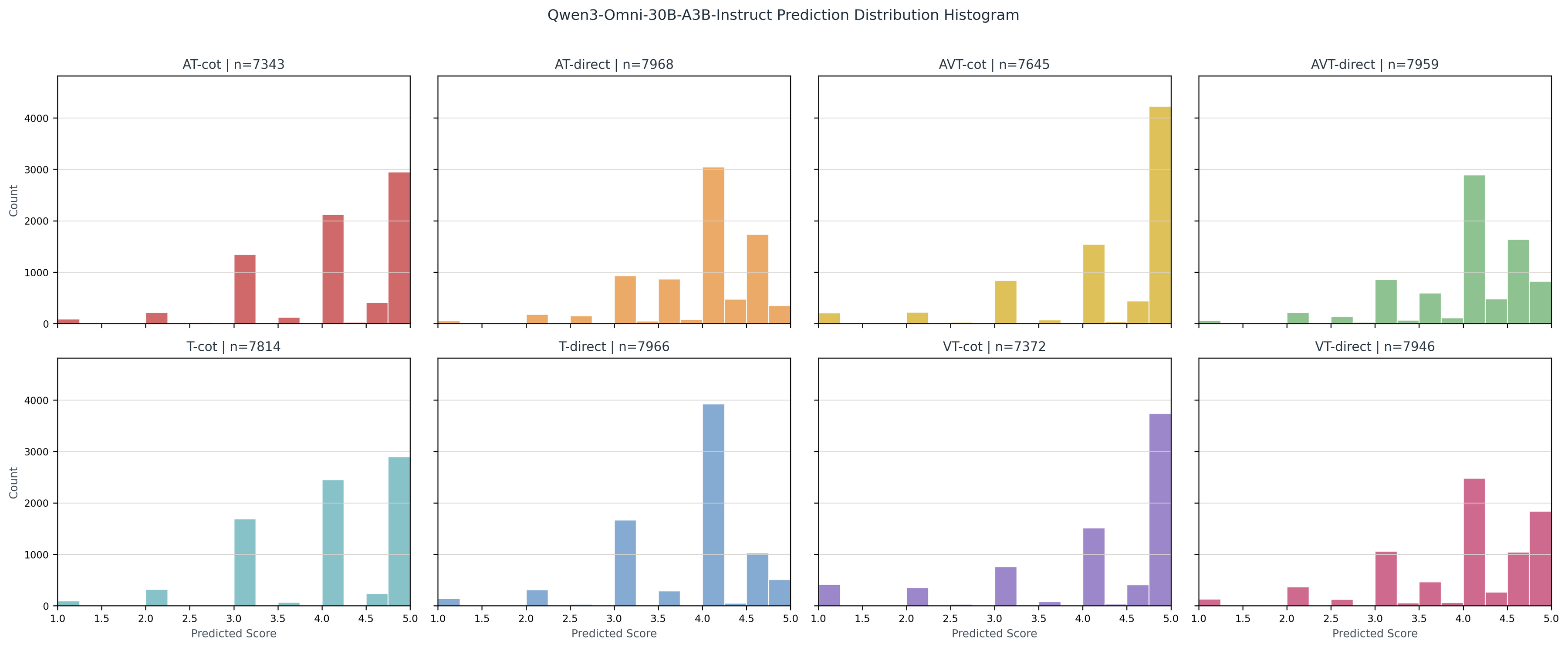}{Qwen3-Omni-30B-A3B-Instruct}

\histfull{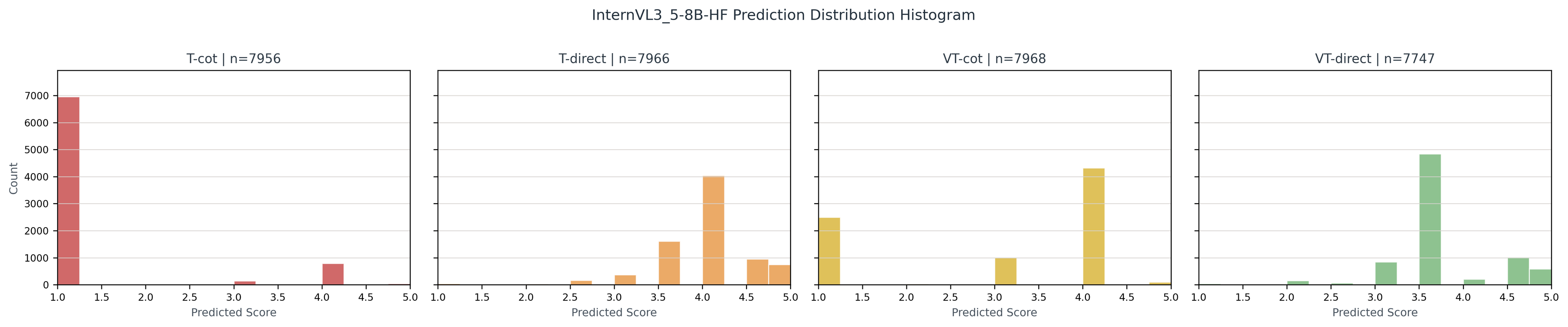}{InternVL3.5-8B-HF}

\histfull{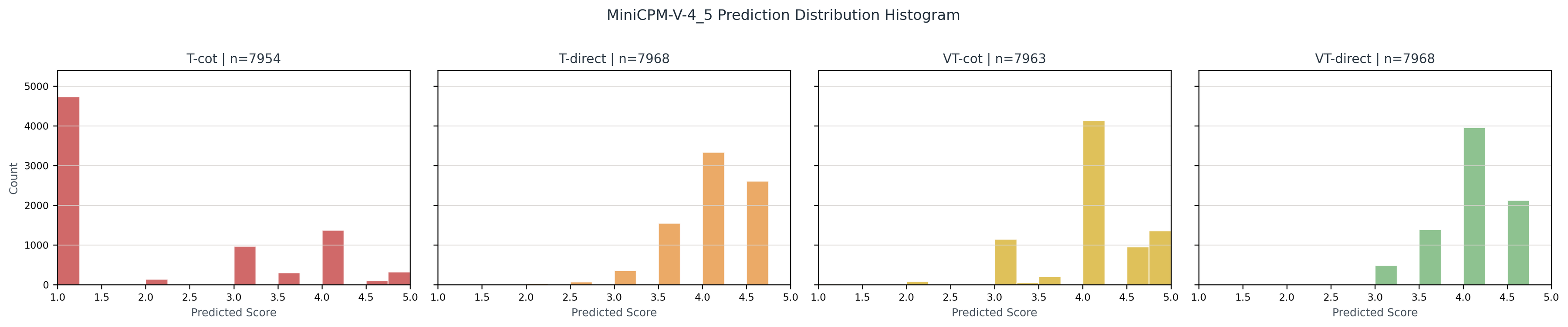}{MiniCPM-V-4.5}

\histfull{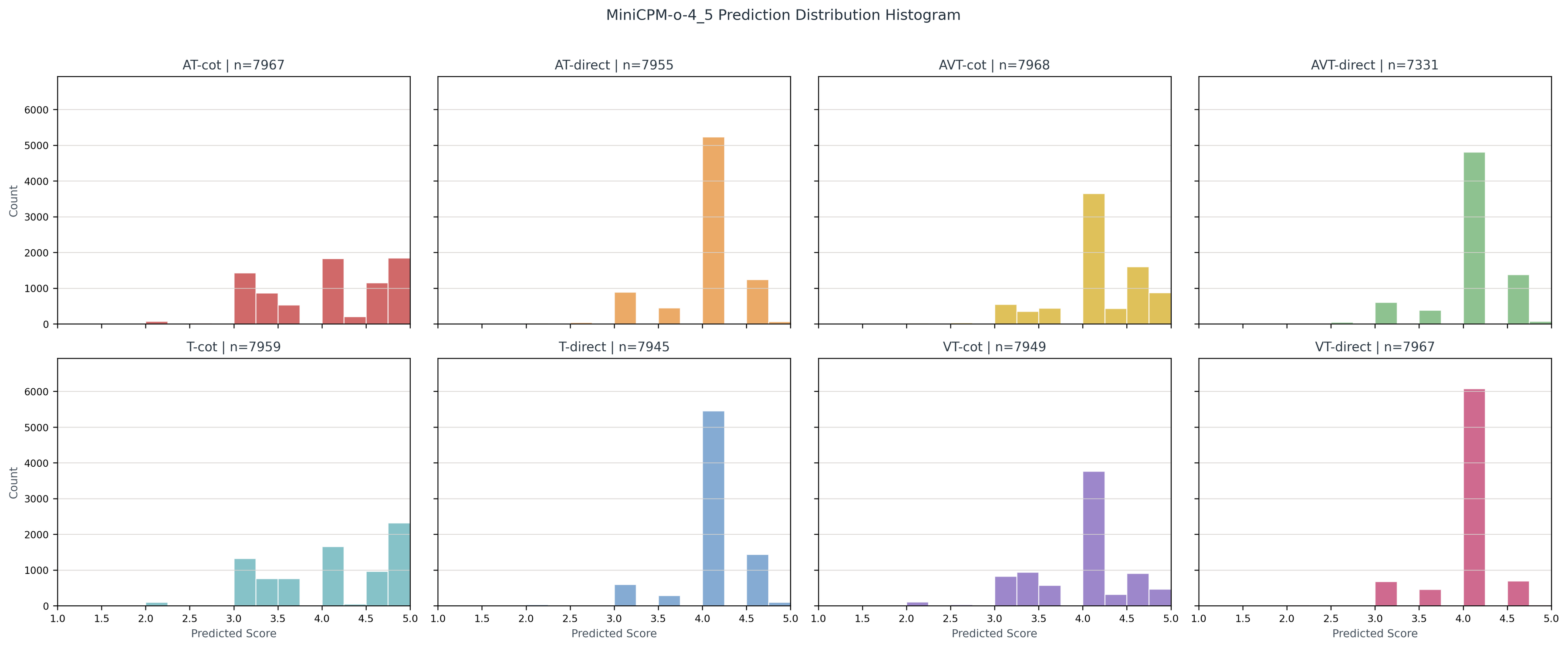}{MiniCPM-o-4.5}

\noindent
\begin{minipage}[t]{0.5\textwidth}
    \centering
    \includegraphics[width=\linewidth]{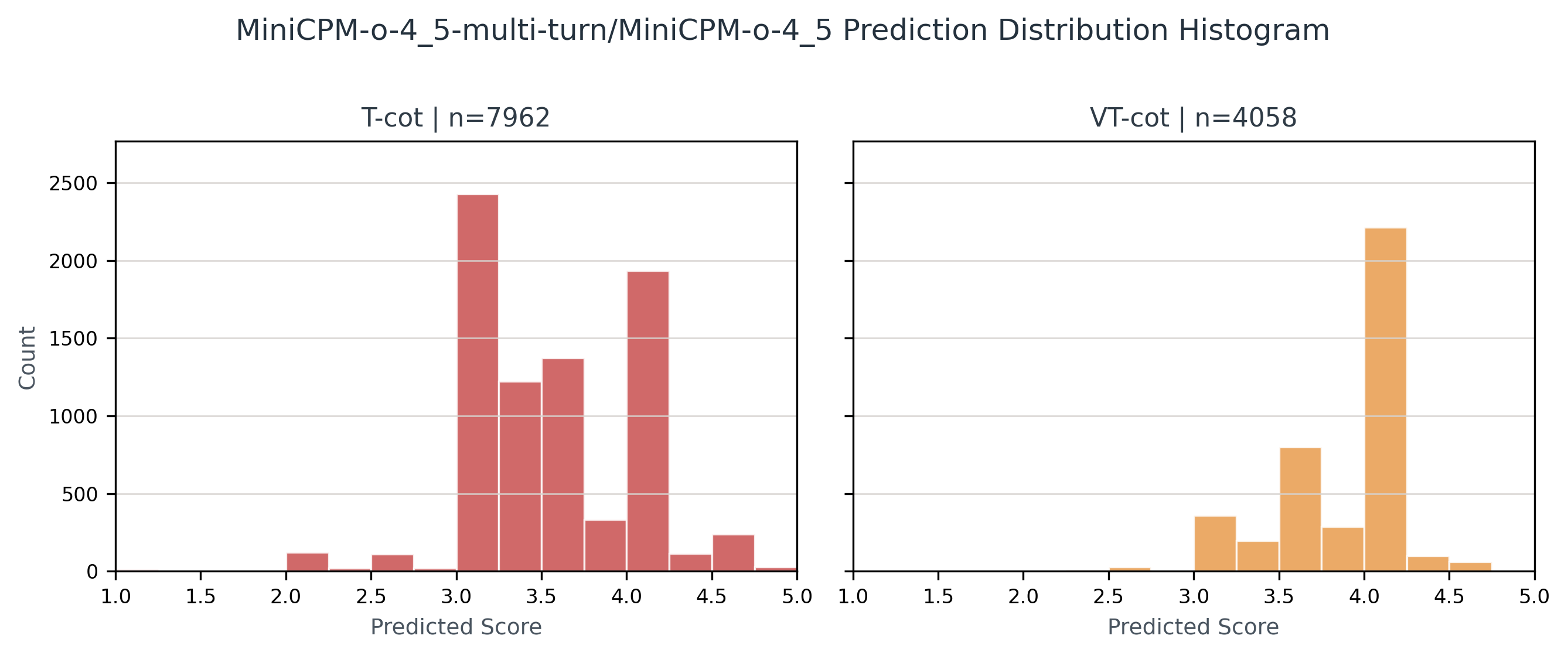}
    
    {\small MiniCPM-o-4.5 multi-turn}
\end{minipage}%
\begin{minipage}[t]{0.25\textwidth}
    \centering
    \includegraphics[width=\linewidth]{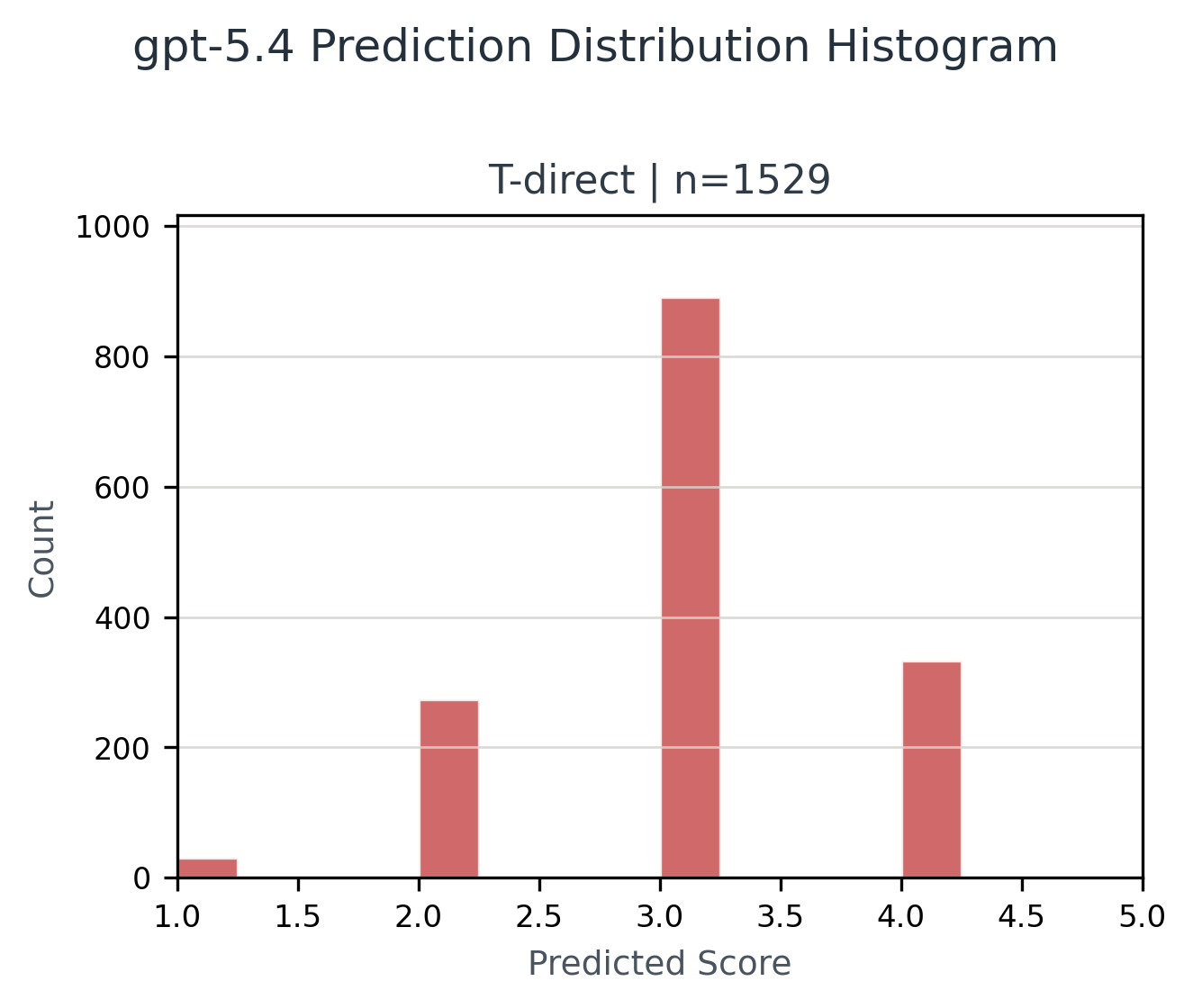}
    
    {\small GPT-5.4}
\end{minipage}%
\begin{minipage}[t]{0.25\textwidth}
    \centering
    \includegraphics[width=\linewidth]{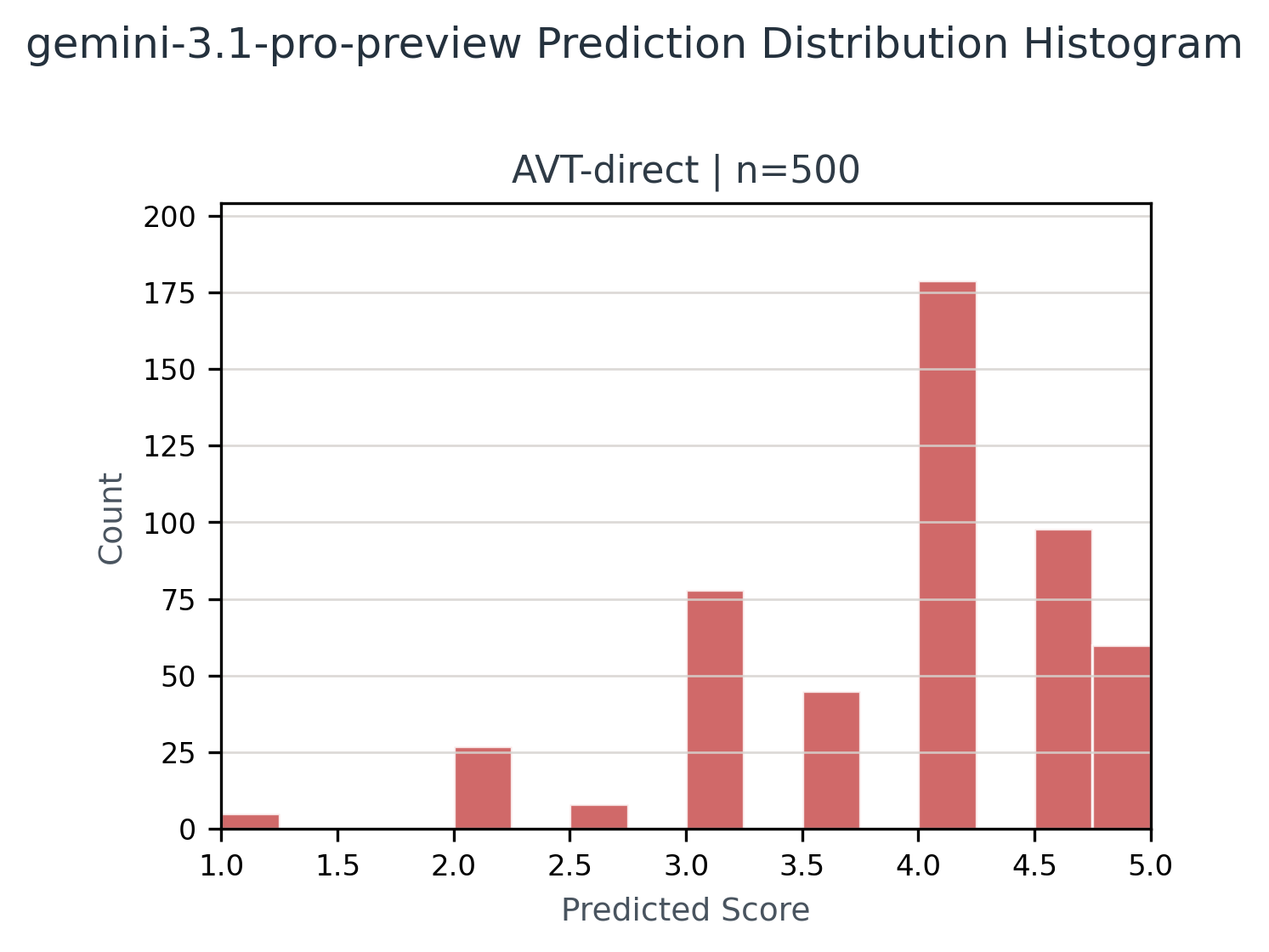}
    
    {\small Gemini-3.1-Pro-Preview}
\end{minipage}

\captionof{figure}{\textbf{Prediction histograms.}}
\label{fig:appendix-histograms-all}

\end{document}